\documentclass[9pt,twocolumn,twoside]{opticajnl}
\journal{opticajournal}

\setboolean{shortarticle}{false}

\usepackage{lineno}

\usepackage{hyperref}
\usepackage[abbreviations]{glossaries-extra}
\usepackage{xspace}
\PassOptionsToPackage{table}{xcolor}
\usepackage{colortbl}
\usepackage{array}
\usepackage{xcolor}
\usepackage{threeparttable}
\usepackage{subcaption}
\usepackage{afterpage}
\usepackage{lipsum}
\usepackage{booktabs}
\usepackage{multirow}
\usepackage{wrapfig}
\usepackage{caption}
\usepackage{amsmath}
\usepackage{sidecap}
\usepackage{enumitem}
\usepackage{placeins}
\usepackage{natbib}
\usepackage{pdflscape}
\newcommand{\shortcite}{\citep}
\definecolor{lightred}{rgb}{1,0.4,0.4}
\definecolor{lightgreen}{rgb}{0.6,1,0.3}
\definecolor{forestgreen}{rgb}{0.133, 0.545, 0.133}
\definecolor{lightyellow}{rgb}{1,1.0,0.6}

\newcommand{\etal}{~et al.\@\xspace}

\newcommand{\eg}{e.g.\@\xspace}

\newcommand{\ie}{i.e.\@\xspace}

\newcommand{\refSec}[1]{Sec.~\ref{sec:#1}}
\newcommand{\refSupSec}[1]{Sec.~S\ref{supplementary:#1}}
\newcommand{\refFig}[1]{Fig.~\ref{fig:#1}}
\newcommand{\refFigFull}[1]{Figure~\ref{fig:#1}}
\newcommand{\refEq}[1]{Eq.~(\ref{eq:#1})}
\newcommand{\refTbl}[1]{Tbl.~\ref{tbl:#1}}
\newcommand{\refObj}[1]{Objective~\ref{obj:#1}}

\newcommand{\novel}[1]{{\it\color{red}#1}}

\newabbreviation{HVS}{HVS}{Human Visual System}
\newabbreviation{AR}{AR}{Augmented Reality}
\newabbreviation{VR}{VR}{Virtual Reality}
\newabbreviation{SLM}{SLM}{Spatial Light Modulator}
\newabbreviation{FoV}{FoV}{Field Of View}
\newabbreviation{HOE}{HOE}{Holographic Optical Element}
\newabbreviation{3D}{3D}{Three-Dimensional}
\newabbreviation{CNN}{CNN}{Convolutional Neural Network}
\newabbreviation{PSF}{PSF}{Point-Spread Function}
\newabbreviation{MLP}{MLP}{Multilayer Perceptron}
\newabbreviation{CBAM}{CBAM}{Convolutional Block Attention Module}
\newabbreviation{FPN}{FPN}{Feature Pyramid Network}
\newabbreviation{MDE}{MDE}{Monocular Depth Estimation}
\newabbreviation{PSP}{PSP}{Pyramid Spatial Pooling}
\newabbreviation{BCP}{BCP}{Bin Center Predictor}
\newabbreviation{CGH}{CGH}{Computer-Generated Holography}
\newabbreviation{SA}{SA}{Segment Anything}
\newabbreviation{HDR}{HDR}{High Dynamic Range}
\newabbreviation{LR}{LR}{Learning Rate}
\newabbreviation{TV}{TV}{Total Variation}
\newabbreviation{SILog}{SILog}{Scale Invariant Log}
\newabbreviation{MTL}{MTL}{Multi-task Learning}
\newabbreviation{ASM}{ASM}{Angular Spectrum Method}
\newabbreviation{ViT}{ViT}{Vision Transformer}
\newabbreviation{2D}{2D}{Two-Dimensional}
\newabbreviation{FPS}{FPS}{Frames Per Second}
\newabbreviation{1D}{1D}{One-Dimensional}
\newabbreviation{DP}{DP}{Double Phase}
\newabbreviation{GM}{GM}{Gradient Matching}
\newabbreviation{KD}{KD}{Knowledge Distillation}
\newabbreviation{SAM}{SAM}{Skip Attention Module}
\newabbreviation{SOTA}{SOTA}{state-of-the-art}
\newabbreviation{fp32}{fp32}{32-bit precision}
\newabbreviation{BN}{BN}{batch normalization}
\newabbreviation{RQ}{RQ}{Research Question}
\newabbreviation{ANOVA}{ANOVA}{Analysis of Variance}

\global\long\def\RQ{\gls{RQ}\xspace}

\global\long\def\HVS{\gls{HVS}\xspace}

\global\long\def\SLM{\gls{SLM}\xspace}

\global\long\def\3D{\gls{3D}\xspace}
\global\long\def\CNN{\gls{CNN}\xspace}
\global\long\def\PSF{\gls{PSF}\xspace}
\global\long\def\MLP{\gls{MLP}\xspace}
\global\long\def\CBAM{\gls{CBAM}\xspace}
\global\long\def\FPN{\gls{FPN}\xspace}
\global\long\def\MDE{\gls{MDE}\xspace}
\global\long\def\PSP{\gls{PSP}\xspace}
\global\long\def\BCP{\gls{BCP}\xspace}
\global\long\def\CGH{\gls{CGH}\xspace}
\global\long\def\SA{\gls{SA}\xspace}

\global\long\def\LR{\gls{LR}\xspace}
\global\long\def\KD{\gls{KD}\xspace}
\global\long\def\TV{\gls{TV}\xspace}
\global\long\def\SILog{\gls{SILog}\xspace}
\global\long\def\MTL{\gls{MTL}\xspace}
\global\long\def\ASM{\gls{ASM}\xspace}

\global\long\def\1D{\gls{1D}\xspace}
\global\long\def\2D{\gls{2D}\xspace}
\global\long\def\3D{\gls{3D}\xspace}

\global\long\def\GM{\gls{GM}\xspace}

\global\long\def\KD{\gls{KD}\xspace}
\global\long\def\SOTA{\gls{SOTA}\xspace}
\global\long\def\fp32{\gls{fp32}\xspace}
\global\long\def\ANOVA{\gls{ANOVA}\xspace}

\newcommand{\inputRGBOnly}{I_{input}}
\newcommand{\conditionVars}{Param_{cond}}
\newcommand{\pixelPitch}{d_x}
\newcommand{\pIndex}{p}
\newcommand{\zIndex}{VD}
\newcommand{\phs}{u}

\newcommand{\propKernel}{h_\pIndex}
\newcommand{\scale}{s}
\newcommand{\tgtIntensity}{I_{(\pIndex, z)}}
\newcommand{\numSubFrames}{T}
\newcommand{\subFrameIndex}{t}
\newcommand{\laserIntensity}{l}
\newcommand{\optmLaserIntensity}{\hat{\laserIntensity}}
\newcommand{\slmPhaseSubFrame}{\phs_\subFrameIndex}
\newcommand{\optmSlmPhaseSubFrame}{\hat{\phs}_\subFrameIndex}
\newcommand{\wavelength}{\lambda}
\newcommand{\pAnchor}{\pIndex_{\text{anchor}}}

\newcommand{\lossFunc}{\mathcal{L}}
\newcommand{\lossTrain}{\mathcal{L}_{train}}
\newcommand{\lossRecon}{\mathcal{L}_{recon}}
\newcommand{\lossLight}{\mathcal{L}_{light}}
\newcommand{\lossScaleInvariant}{\mathcal{L}_{silog}}
\newcommand{\lossGradingMatching}{\mathcal{L}_{gm}}
\newcommand{\lossTV}{\mathcal{L}_{tv}}
\newcommand{\lossDepth}{\mathcal{L}_{depth}}

\title{Configurable Holography: Towards Display and Scene Adaptation}

\author[1]{Yicheng Zhan}
\author[2]{Liang Shi}
\author[2]{Wojciech Matusik}
\author[3]{Qi Sun}
\author[1,*]{Kaan Akşit}

\affil[1]{University College London, Gower Street, London, UK, WC1E 6BT}
\affil[2]{Massachusetts Institute of Technology, MA 02139, Massachusetts, USA}
\affil[3]{New York University, New York, USA}

\affil[*]{kaanaksit@kaanaksit.com}

\begin{abstract}
Rendering holograms for holographic displays is often an iterative and computationally costly process.
Emerging learned holography methods have alleviated this bottleneck by enabling fast hologram rendering with
improved reconstruction quality. However, existing methods still depend on fixed display hardware and scene parameters,
requiring retraining for each new configuration. This limits rapid adaptation to different visual needs,
including scene brightness, user focus preference, and hardware compatibility.

We introduce \emph{Configurable Holography}, a learned CGH framework in which a single model adapts to diverse display-scene parameters through explicit conditioning, eliminating the need for retraining.
As a prototype, we present a configurable structure and derive a family of models that continuously adapt to propagation distance, volume depth, peak brightness, pixel pitch, and wavelength.
To further improve efficiency, we incorporate auxiliary monocular depth estimation for depth-aware 3D hologram synthesis from RGB-only inputs and apply knowledge distillation for interactive inference.
Our extensive simulation and hardware experiments on three holographic display prototypes with different combinations of configurations show on-par reconstruction quality with existing methods, offering up to 2$\times$ speed-up in fp32.
Our work represents an initial step toward flexible, general-purpose learned holography systems that can seamlessly adapt across diverse hardware and user-specific visual requirements.
\end{abstract}

\setboolean{displaycopyright}{false}

\begin{document}

\maketitle

\begin{figure*}[ht!]
  \centering
  \includegraphics[width=1\linewidth]{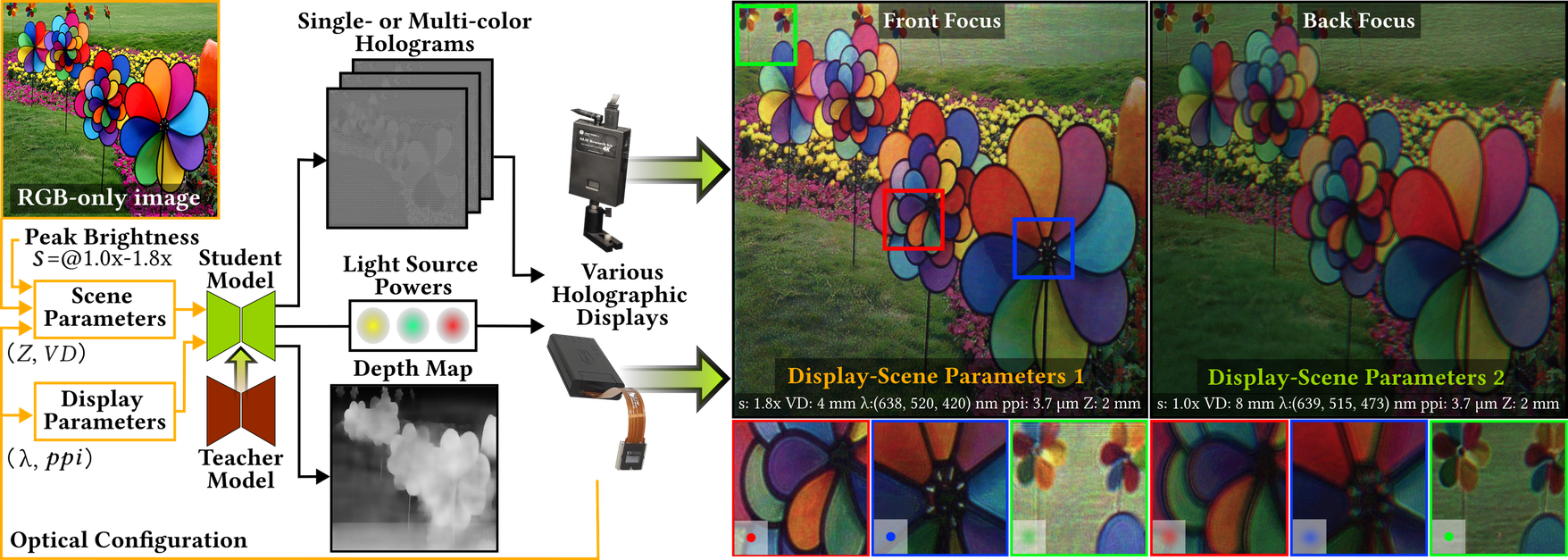}
    \caption{
            Our configurable holography model supports a range of display–scene parameters without retraining, including peak brightness ($s$),
            propagation distance ($Z$), volume depth ($VD$), working wavelength ($\lambda$), and pixel pitch ($ppi$).
             We distill our model into a student model, achieving 2x speed up w.r.t. literature~\cite{shi2022end}.
             Our model can synthesize single or multi-color 3D holograms from RGB-only 2D input images  by internally predicting depth, simply eliminating the need for depth input
             (Source Image: \cite{Windmills2009}).
            }
  \label{fig:teaser}
  \vspace{-5mm}
 \end{figure*}

\section{Introduction}
\label{sec:intro}

Holographic displays~\cite{kim2024holographic} support optical focus cues and multiple perspectives, promising authentic immersive \3D visual experiences as a potential future display technology.
However, rendering visuals for holographic displays remains computationally demanding and slow \cite{blinder2019signal}.
Emerging learned holography methods \cite{shi2022end} promise to accelerate and improve visual quality in the holographic displays.
However, existing learned holography models are inflexible as they require training a dedicated model for each set of display-scene parameters.
This constraint becomes a bottleneck in practice.
For instance, users demand instant, continuous control over focus range and brightness levels;
those users with prescriptions also require focus adjustments for clearer images;
and developers building displays need full tunability at interactive-rate across system parameters
to accelerate the development processes of display prototypes.
\textit{Meeting these diverse requirements with separate models would cause substantial overhead in model management, deployment speed, initialization time, and training, compromising
seamless visual experiences. This makes model configurability an important capability for future holographic displays.}
We also acknowledge that for a deployed holographic display with fixed focal length, per-configuration models remain practical.

We advocate \emph{Configurability} as an important objective for learned \CGH.
We define configurability as the ability of a single learned model to \emph{adapt} its hologram synthesis behavior according to the requested display-scene parameters at inference time, without retraining for each configuration.
In this view, learned \CGH is not merely a fast alternative to optimization under one setting, but a programmable hologram generator that can accommodate parameter changes that naturally arise across users, scenes, and devices.
Achieving this goal is non-trivial because display-scene parameters influence diffraction physics and perceived \3D appearance; a configurable model must therefore balance generality, reconstruction fidelity, and efficiency.

Building on this concept, our work depicted in~\refFig{teaser} focuses on interactive \3D hologram synthesis across a continuous range of various display-scene parameters using a single learned model.
We introduce a highly configurable network structure that synthesizes conventional single-color~\cite{kavakli2023realistic} and multi-color~\cite{kavakli2023multicolor, Eric2023MultiSourceHolo, zhan2024autocolor}
holograms while being conditioned on scene parameters (propagation distance, volume depth, peak brightness), and display parameters (pixel pitch and wavelength).
Alongside configurability, we also explore two strategies to maximize the efficiency of learned \CGH methods.
First, we treat \MDE as an auxiliary training task to improve the \3D accuracy of hologram synthesis.
Importantly, our goal is not to compete with advanced \MDE methods, but to demonstrate that depth information can serve as an effective
additional supervision signal for accurate \3D hologram synthesis from RGB-only inputs with no depth, the most common form of media.
Second, we apply \KD~\cite{wang2021knowledge} to distill a compact and configurable student model that is substantially accelerated for interactive inference and preserves reconstruction quality.
While our models demonstrate that configurable \CGH is feasible, we emphasize that this work is a prototype and configurability itself warrants more rigorous and scaled evaluation.
Accordingly, beyond reporting performance, we propose several \RQ: why is achieving configurability important and non-trivial?
Why are \MDE and \KD beneficial to \CGH? What limitations does our method exhibit, and how might future work address them?
We hope these observations will help readers reason about configurable \CGH and inspire future research.
Our contributions are as follows:
\begin{itemize}[leftmargin=*, nosep]
  \item \textbf{Configurable Holography.}
  We introduce configurability as a target capability for the learned \CGH method and present a configurable model structure as a prototype.
  From which we derive a family of models supporting both single- and multi-color \3D holograms across a range of display-scene parameters,
  including propagation distance, volume depth, peak brightness, wavelength, and pixel pitch.
  \item \textbf{Advancing efficient and accurate \3D Hologram Synthesis from RGB-only inputs.}
  We unearth the correlation between depth estimation and \3D hologram synthesis tasks in learned methods.
  Our model leverages this correlation and adopts multitask learning with hard-parameter sharing \cite{caruana1993multitask} to convert RGB-only \2D images to accurate \3D holograms.
  We also apply \KD to train a compact student model that achieves up to 2$\times$ faster hologram synthesis than prior learned approaches~\cite{shi2022end} under \fp32 while preserving reconstruction quality.
  \item \textbf{Evaluation and empirical validation.}
  We conduct extensive quantitative and qualitative experiments comparing against existing learned \CGH methods and validate our findings on three physical holographic display prototypes.
\end{itemize}
Our design targets conventional holographic displays and therefore inherits their typical FoV and eyebox constraints~\cite{Shi2017};
extending configurability to emerging display architectures~\cite{Changwon2023Waveguide, Grace2023MultiSource, Chae23Sig}
and novel hologram representations~\cite{kim2024holographic, choi2022time} remains an important direction for future work.
Our code is available at [REVIEW].

\section{Related Work}
\paragraph{Learned Computer-Generated Holography. }
Early learned approaches to single-color hologram synthesis primarily adopt U-Net–based \CNN architectures~\cite{siddique2021u}
or resolution-preserving residual stacks to accelerate inference~\cite{shi2022end}.
These methods require RGB-D inputs and are restricted to a fixed set of display-scene parameters.
To remove the dependency on depth input, Liu \etal~\cite{liu2023dge} introduce a multi-stage pipeline that separately performs depth estimation and RGB-D hologram optimization, resulting in increased computational cost.
Ishii \etal~\cite{Yoshi2023Optics} further extend this design with an additional hologram refinement stage, leading to even slower synthesis compared to prior methods~\cite{aksit2023holobeam, liu2023dge}.
Akşit and Itoh~\cite{aksit2023holobeam} collapse these stages into a single-stage \CNN for improved efficiency;
however, the resulting 3d hologram exhibits inaccurate depth reconstruction.
Concurrent works also explore feature distillation~\cite{wang2025cafdn}, diffractive decoding~\cite{isil2025snapshot},
propagation-adaptive \CGH~\cite{liu2025sfo}, and hologram synthesis from 2D-only inputs~\cite{chang2023picture, chang2025photonix, sonker2025netholo}.
\textit{However, they either focus on 2D holograms or assume fixed optical configurations, our work addresses these limitations by
introducing a depth-input-free, single-stage approach that jointly supports display-scene parameters
configuring for both single- and multi-color 3D holograms at interactive rates within single model.}

\section{Method: Configurable Holography}
\label{sec:method}

Our method aims to synthesize \3D holograms for various holographic displays at inference time using a single learned model.
Our proposed method takes RGB-only images and display-scene parameters as inputs and outputs depths and holograms.
\textit{We leverage depth estimation as a beneficial parallel task to help generate accurate 3D holograms from 2D images.}
\textit{Thus, our model does not aim to compete the \MDE models,
rather our model could benefit from any potential advancements in their accuracy in the future.}

\subsection{RQ{}1: Why is configurable \CGH important and non-trivial?}
Configurability is important for both users and researchers: practical holographic displays and viewing preferences
require instant and frequent changes in propagation distance, volume depth, and brightness, while display development involves exploring wavelength
and pixel pitch variations across hardware prototypes. However, achieving configurability is non-trivial because it requires a network to
continuously adapt the \emph{diffraction-driven} image formation process across parameters, rather than memorizing a
fixed setting. This is particularly challenging for long-range light propagation in 3D hologram, where the optical
field oscillates rapidly as light travels through space, and where different parameters exhibit uneven conditioning
difficulty—an observation we analyze empirically later in ~\refSec{configurability}.

\subsubsection{Problem Definition: Synthesizing 3D Holograms}

Holographic displays rapidly play successive holograms to generate full-color images, which the \HVS fuses into a perceived color reconstruction.
Single-color holograms are computed for one wavelength; multi-color holograms are computed jointly for multiple primaries.
Hologram synthesis can be modeled by the optimization
\begin{equation}
\begin{split}
Z_{0} = Z - \frac{\zIndex}{2},
Z_{n} = Z + \frac{\zIndex}{2}, \\
I_r(p,t,z) = \left\lvert \laserIntensity_{(\pIndex, \subFrameIndex)} e^{i\frac{\wavelength_{\pIndex}}{\wavelength_{\pAnchor}}\slmPhaseSubFrame} * \propKernel(\wavelength_p, z, d_x) \right\rvert^2, \\
\optmSlmPhaseSubFrame, \optmLaserIntensity_{(\pIndex, \subFrameIndex)}
\leftarrow
\operatorname*{argmin}_{\slmPhaseSubFrame, \laserIntensity_{(\pIndex, \subFrameIndex)}} \lossFunc_{\text{img}} \left( \sum_{z = Z_{0}}^{Z_{n}} \sum_{\pIndex=1}^{P} \sum_{\subFrameIndex=1}^{\numSubFrames} I_r(p, t, z), \scale \tgtIntensity\right),
\end{split}
\label{eq:hologram_synthesis}
\end{equation}
where $Z \in \mathbb{R}$ denotes the light propagation distance, $\zIndex \in \mathbb{R}$ the volume depth, $P \in \mathbb{Z}$ the number of color primaries (\ie typically three),
and $\pIndex \in \mathbb{Z}$ the primary index. $T \in \mathbb{Z}$ denotes the number of subframes required to reproduce a full-color image (\ie typically three),
and $\subFrameIndex \in \mathbb{Z}$ denotes the subframe index. $\laserIntensity_{(\pIndex, \subFrameIndex)} \in \mathbb{R}^{P \times T}$ represents the light source power
of the $\pIndex$-th primary at the $\subFrameIndex$-th subframe. $\wavelength_\pIndex \in {400\text{–}700 \mathrm{nm}}$ denotes the active primary wavelength,
while $\wavelength_{\pAnchor}$ denotes the anchor wavelength used for calibration. $\slmPhaseSubFrame \in \mathbb{C}^{H \times W}$ represents the phase-only hologram,
and $d_x$ denotes the pixel pitch.
Here, $\tgtIntensity \in \mathbb{R}^{H \times W}$ denotes the target intensity image, $\scale \in \mathbb{R}$ controls brightness scaling,
and $\propKernel \in \mathbb{C}^{H \times W}$ denotes the wavelength- and distance-dependent propagation kernel~\cite{matsushima2009band, kavakli2022learned, Choi2021neural}.
$\lossFunc_{\text{img}}$ measures the visual discrepancy between the target and reconstructed images. For single-color holography,
$\laserIntensity$ reduces to identity selection, whereas multi-color holography requires joint optimization across wavelengths.
Existing optimization pipelines~\cite{kavakli2023multicolor} typically require minutes per requested $\scale$ and rely on RGB-D inputs.
\textit{Configurability} refers to replacing this parameterized optimization with a single model capable of adapting across a continuous range of $(\scale, Z, VD, \wavelength, d_x)$.

\begin{figure*}[t]
  \centering
  \includegraphics[width=0.98\textwidth]{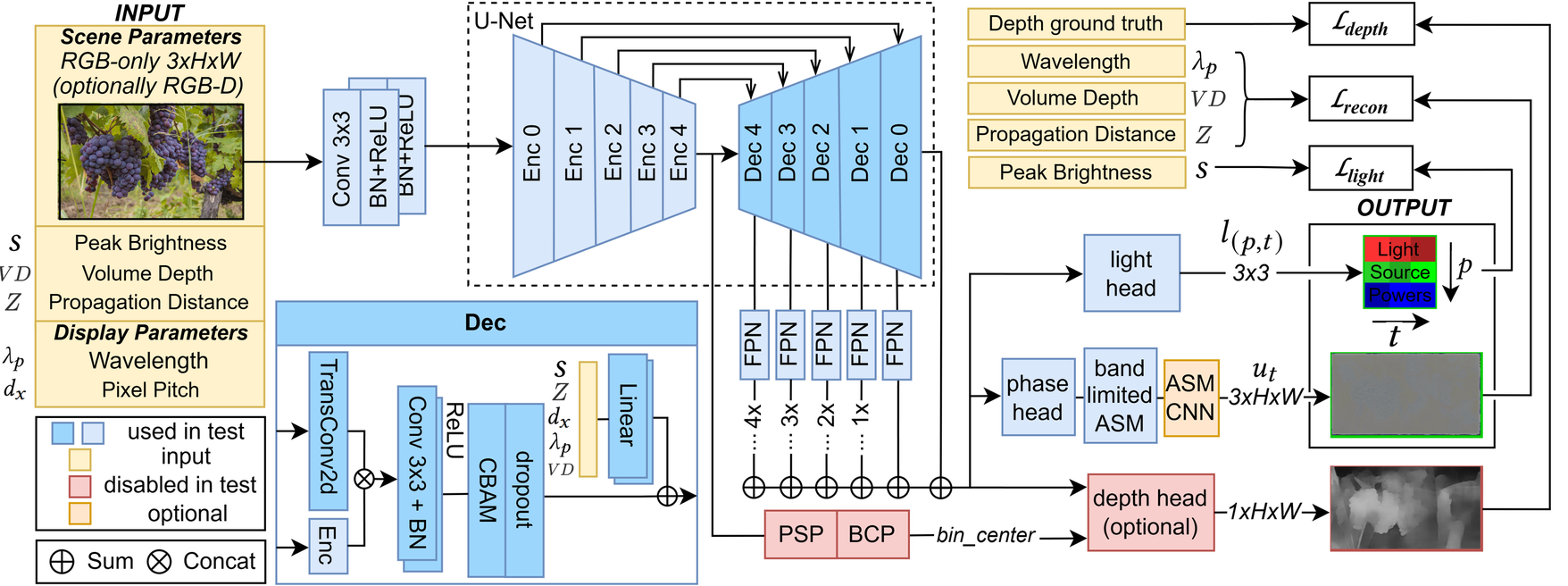}
  \caption{Our teacher model.
A \FPN~\cite{lin2017feature} connects to every decoder stage of our U-Net to leverage multi-scale spatial features from RGB-only or RGB-D inputs.
\CBAM~\cite{woo2018cbam} introduces channel and spatial attention, while each decoder stage is conditioned on peak brightness, wavelength, volume depth, propagation distance, and pixel pitch.
A \PSP layer~\cite{he2015spatial} aggregates global context for depth estimation, and a dedicated light head predicts the required light source powers.
Here, $BCP$ denotes the \BCP~\cite{agarwal2022attention} (RGB-only source: \cite{grapes2012}).
 }
 \vspace{-5mm}
  \label{fig:teacher_network}
 \end{figure*}

Diffraction exhibits scaling properties that relate wavelength, pixel pitch, and propagation distance: a hologram computed at one wavelength
can approximate another by rescaling $Z$ as $Z_2 = \frac{\lambda_2}{\lambda_1} \times Z_1$, and a similar relation connects pixel pitch and distance
as $Z_2 = (\frac{p_1}{p_2})^2 \times Z_1$ (see Supplementary \refSupSec{diffraction_scalability} for details and \PSF visualizations).
While these properties motivate shared structure across configurations, they hold only for single-color settings; multi-color holograms require jointly optimizing across wavelengths, where simple rescaling is insufficient.
This partially explains why some parameters are harder to condition than others and motivates our explicit conditioning strategy.
Additionally, iterative random-phase methods like GS and SGD~\cite{GSMethod, chakravarthula2020wirtinger} are inherently non-configurable,
as each run solves for a hologram under a fixed propagation kernel parameterized by $(\wavelength, \pixelPitch, Z)$.
To verify this, we conduct an experiment in Supplementary~\refSupSec{SGD_configurable}, where we jointly optimize a hologram using SGD across four propagation distances.
Our findings confirm that changing any parameter invalidates the accumulated phase updates, and joint optimization produces non-ideal phases due to conflicting gradients.

\subsection{Answer to RQ{}1: A Configurable Model Prototype}
We convert the parameterized hologram synthesis process in \refEq{hologram_synthesis} into a single-stage learned model that is explicitly conditioned on display-scene parameters.
Our final deployed model is a compact student model distilled from a stronger teacher using \KD.
This design targets configurability as \emph{continuous generalization} over supported parameter ranges.
Our model unifies three stages:
(a)~\textit{wavefront synthesis}, mapping RGB inputs to a complex-valued field via a parameter-conditioned U-Net;
(b)~\textit{wave propagation}, propagating the field via kernel; and
(c)~\textit{phase extraction}, using imaginary part of the propagated field as hologram.
We first introduce teacher, then the student model.

\subsubsection{Teacher Model}
We provide a complete layout of our teacher model in \refFig{teacher_network}, the details of the modules can be found in Supplementary \refSupSec{Model_structure}.
The teacher takes $\inputRGBOnly$, $\wavelength_{\pIndex}$, $\pixelPitch$, $\scale$, $VD$, and $Z$ as inputs, denoted as $\conditionVars$.
During training, we vary these inputs so that the model can adapt the hologram synthesis process to preferred display--scene parameters at test time.
Specifically, our training draws the input variables from
\begin{equation}
\begin{split}
\wavelength_{\pIndex} \subseteq \{ (640,515,470)\} ~nm, \scale \subseteq \{1.0, 1.4, 1.8 \},\\
 VD \subseteq \{4.0, 8.0\} ~mm, Z \subseteq \{2, 4, 7, 10\} ~mm, \pixelPitch \subseteq \{3.74\} ~\mu m.
\label{eq:variable_set}
\end{split}
\end{equation}
Our choice of $Z$ follows recent learned holography literature~\cite{shi2022end, shi2021towards, aksit2023holobeam}, and our $VD$ choices cover common VR focal ranges (40--75~$mm$), roughly corresponding to placing virtual images from our $VD$ between 5 Diopter to infinity.
Given the scalability property in \refSupSec{diffraction_scalability}, we deliberately include a single pixel pitch in \refEq{variable_set} to control training permutations and computational cost.
To study broader conditioning over $\pixelPitch$ (and other display parameters) with wider permutations, we also introduce an RGB-D condition variant (depth provided as input) and report its configurations and results in \refTbl{scene_params} and Supplementary \refSupSec{RGBD_condition}.
Overall, we generate permutations of \refEq{variable_set}, resulting in 24 training cases, and train the teacher on the full set in one session.

Our teacher processes $\inputRGBOnly$ using a U-Net structure~\cite{ronneberger2015u} with encoder EfficientNet 1B~\shortcite{tan2019efficientnet} (code from ~\cite{smp2019}).
We condition the decoder on the display-scene parameters and aggregate multi-scale decoder features with an \FPN to form a shared latent code.
We feed this latent code into three task-specific heads for predicting \textit{phase-only holograms}, \textit{light source powers}, and \textit{depth} from RGB-only input.
Their outputs are regularized by
\begin{equation}
\lossTrain = \alpha_0 \lossRecon + \alpha_1 \lossLight + \alpha_2 \lossDepth,
\label{eq:total_training_loss}
\end{equation}
where $\alpha_0=1$, $\alpha_1=1$, $\alpha_2=30$ balance reconstruction, light power, and depth terms.
$\lossRecon$ is a multiplane reconstruction loss~\shortcite{kavakli2023realistic} that measures the discrepancy between the optically simulated reconstruction and the target image across depth planes.
Given a ground-truth depth map, we generate target focal-stack images by applying per-pixel depth-dependent defocus following Kavakl{\i} \etal~\cite{kavakli2023multicolor}.
$\lossRecon$ combines three $L_2$ terms: a global reconstruction term, a masked term emphasizing in-focus regions, and a self-weighting term prioritizing high-intensity targets:
\begin{equation}
\begin{aligned}
\lossRecon &= m_0 L_2(rec_k, target_k) + m_1 L_2(rec_k \cdot M_k, target_k \cdot M_k) \\
           &\quad + m_2 L_2(rec_k \cdot target_k, target_k \cdot target_k) + L_{smooth},
\end{aligned}
\label{eq:recon_loss}
\end{equation}
where $rec_k$ and $target_k$ denote the reconstructed and target images at depth plane $k$, $M_k$ is the binary in-focus mask derived from the depth map, and $L_{smooth}$ is a multi-scale total variation regularizer on the predicted phase~\cite{kavakli2023multicolor, shi2022end}.
$\lossLight$ constrains per-subframe laser powers to match the target color balance, following~\shortcite{kavakli2023multicolor}.
We also incorporate FLIP loss~\shortcite{andersson2020flip} to improve color consistency.
Full mathematical definitions of all loss terms are provided in Supplementary \refSupSec{Loss}.

\paragraph{Parameter Embedding.}
We propose to encode $\conditionVars$ using novel scalar embeddings and a 1D \PSF that captures the underlying physical conditions.
The scalar display–scene parameters are embedded via sinusoidal encoding. A complex-valued 2D PSF parameterized by $(\wavelength,\pixelPitch,Z)$ is computed,
\begingroup
\setlength{\intextsep}{0.5pt}
\setlength{\columnsep}{10pt}
\begin{wrapfigure}{t!}{0.68\columnwidth}
\centering
\includegraphics[width=0.99\linewidth]{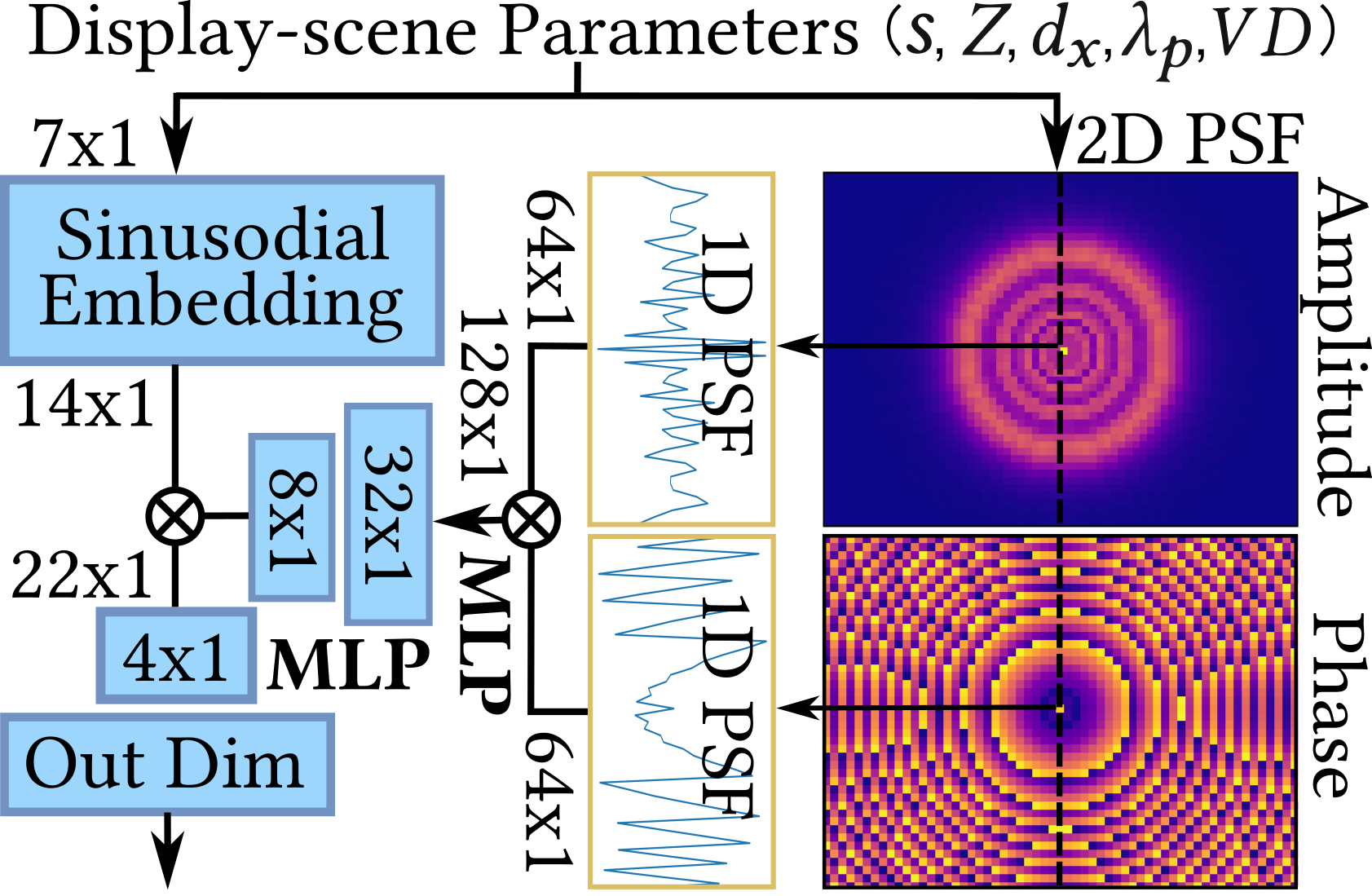}
\caption{Parameters Embedding layer.}
\label{fig:1D_PSF}
\end{wrapfigure}
from which the central x-axis is extracted as a 1D PSF. This 1D PSF is processed by two linear layers and concatenated with the scalars;
the fused features are then passed through two additional linear layers to produce a conditioning vector injected into each decoder stage.
Compared to prior 2D PSF conditioning~\cite{Asano2024SigPoster, liu2025sfo},
our design is more efficient and uniquely supports multi-parameter conditioning.
We include an ablation study in Supplementary~\refSupSec{additional_analyses} to validate the effectiveness of both branches.

\paragraph{Multi-task Learning.}
We adopt hard-parameter sharing~\cite{caruana1993multitask, ruder2019latent, sarwar2019incremental}, where a common U-Net backbone is shared across hologram, light power, and depth estimation tasks while maintaining task-specific heads.
This design is motivated by the observation that prior single-stage RGB-only methods~\cite{aksit2023holobeam} produce highly inaccurate depth reconstruction; jointly learning \MDE forces the shared representation to encode geometry relevant to \3D focus.

\paragraph{Phase-only Hologram Synthesis Task.}
Our phase head maps the latent code to a complex-valued field, which is propagated using a band-limited \ASM with the input $Z$ and $\wavelength$ across all subframes.
We extract the imaginary part as the phase-only hologram.
To support long propagation distances (e.g., $Z=10~mm$), we incorporate an ASM CNN block and skip-connect its output with the propagated phase.
We find this block empirically necessary for long propagation; see Supplementary~\refSupSec{Phase_Prediction_Layer_Structure} for details.

\paragraph{Light Power Estimation Task.}
Our light head predicts a $\subFrameIndex \times \pIndex$ (e.g. $3\times3$) matrix of light source powers in $[0,1]$.
It aggregates spatial information from the latent code and regresses intensities used to match the color of target reconstruction.

\paragraph{Depth Estimation Task.}
Besides phase and light heads, the teacher includes a depth head as an auxiliary task that improves hologram prediction when only RGB input is available.
The head follows a bin-based formulation and combines encoder features with the latent code to predict dense depth.
We define $\lossDepth = \lossScaleInvariant + \lossGradingMatching + \lossTV$, with \SILog~\shortcite{eigen2014depth}, \GM, and \TV terms detailed in Supplementary \refSupSec{Loss}.

\subsubsection{Student Model}
\begin{figure}[t!]
  \centering
  \includegraphics[width=0.48\textwidth]{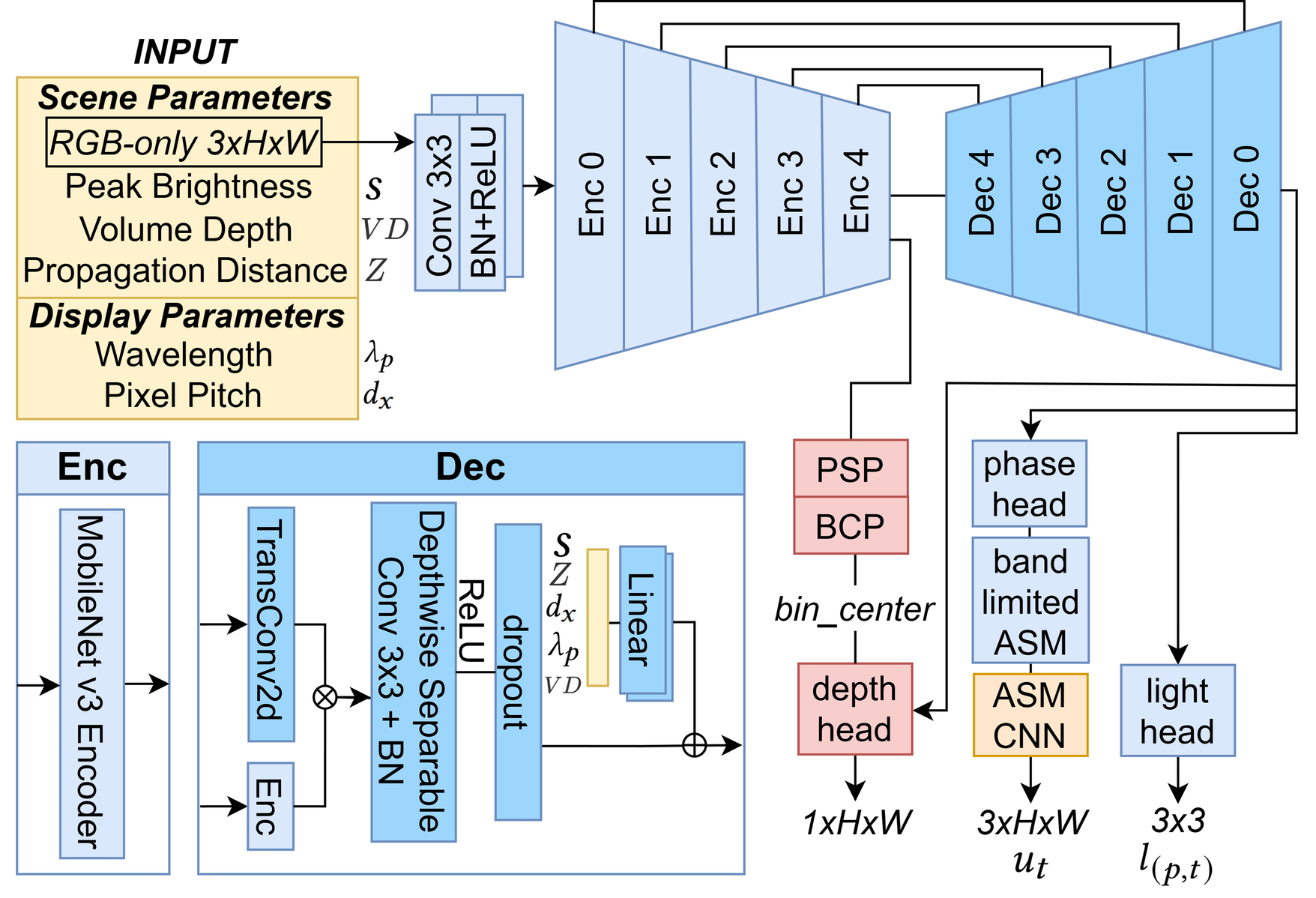}
  \caption{Overview of our student model.}
  \label{fig:student_network}
 \end{figure}
While the teacher model achieves strong reconstruction quality,
it requires 651~ms per frame on an NVIDIA A100 GPU at $1920 \times 1080$ under \fp32 (10.74M parameters), being inefficient.
This motivates \KD: rather than deploying the teacher directly or training a small model from scratch (which fails; see \refSec{KD}),
we distill the teacher's knowledge into a compact student (2.19M parameters) that achieves 39~ms inference time---a 16.7$\times$ speed-up over the teacher and 1.9$\times$ over Tensor V2~\cite{shi2022end}---while preserving reconstruction quality.
As shown in \refFig{student_network}, we accelerate the teacher by converting it to a smaller student model using \KD~\cite{hinton2015distilling, wang2021knowledge, gou2021knowledge}.
\KD transfers knowledge from large teacher models to students, typically via response-based~\cite{micaelli2019zero} or feature-based~\cite{chen2020learning} distillation,
and has been combined with \MTL in other domains~\cite{li2020knowledge, ignatov2021fast}.
\textit{Without \KD, smaller models trained independently fail to reliably solve our hologram synthesis task (see \refSec{KD})}.
The student takes the same inputs as the teacher and preserves the same outputs (phase, light, and depth), but replaces the encoder with MobileNetV3~\cite{howard2019mobileNetv3}
(code from ~\cite{smp2019}) and uses a lightweight decoder (e.g., depthwise separable convolutions).
Conditioning embeddings and prediction heads are shared with the teacher; module-level descriptions are provided in Supplementary \refSupSec{RGBD_condition}.
The last stage of our student decoder conditions the U-Net with $\wavelength$, $\scale$, $\pixelPitch$, $VD$, and $Z$.
We apply logit-based \KD~\cite{hinton2015distilling} for our training
\begin{equation}
 \mathcal{L}_{KD} = T^2 \cdot D_{KL}(S(y_{student} / T), S(y_{teacher} / T)),
\end{equation}
where $D_{KL}$ is the Kullback-Leibler divergence~\cite{kullback1951kullback}, $S$ is softmax, and $T=5$.
We distill both phase and depth: $\lossDepth$ targets the teacher-predicted depth, and we add a Charbonnier loss~\cite{Jonathan2017CBloss} to encourage smooth phase profiles.
Our distillation loss is
\begin{equation}
\mathcal{L}_{\text{distill}} = \mathcal{L}_{\text{KD}_{\text{phase}}} + \mathcal{L}_{\text{Charbonnier}} + \mathcal{L}_{\text{KD}_{\text{depth}}} + \mathcal{L}_{\text{depth}}.
\end{equation}
Final loss uses equal weighting: $\mathcal{L} = \mathcal{L}_{\text{distill}} + \mathcal{L}_{\text{train}}$.
\begin{figure*}[htbp]
  \centering
  \includegraphics[width=1\textwidth]{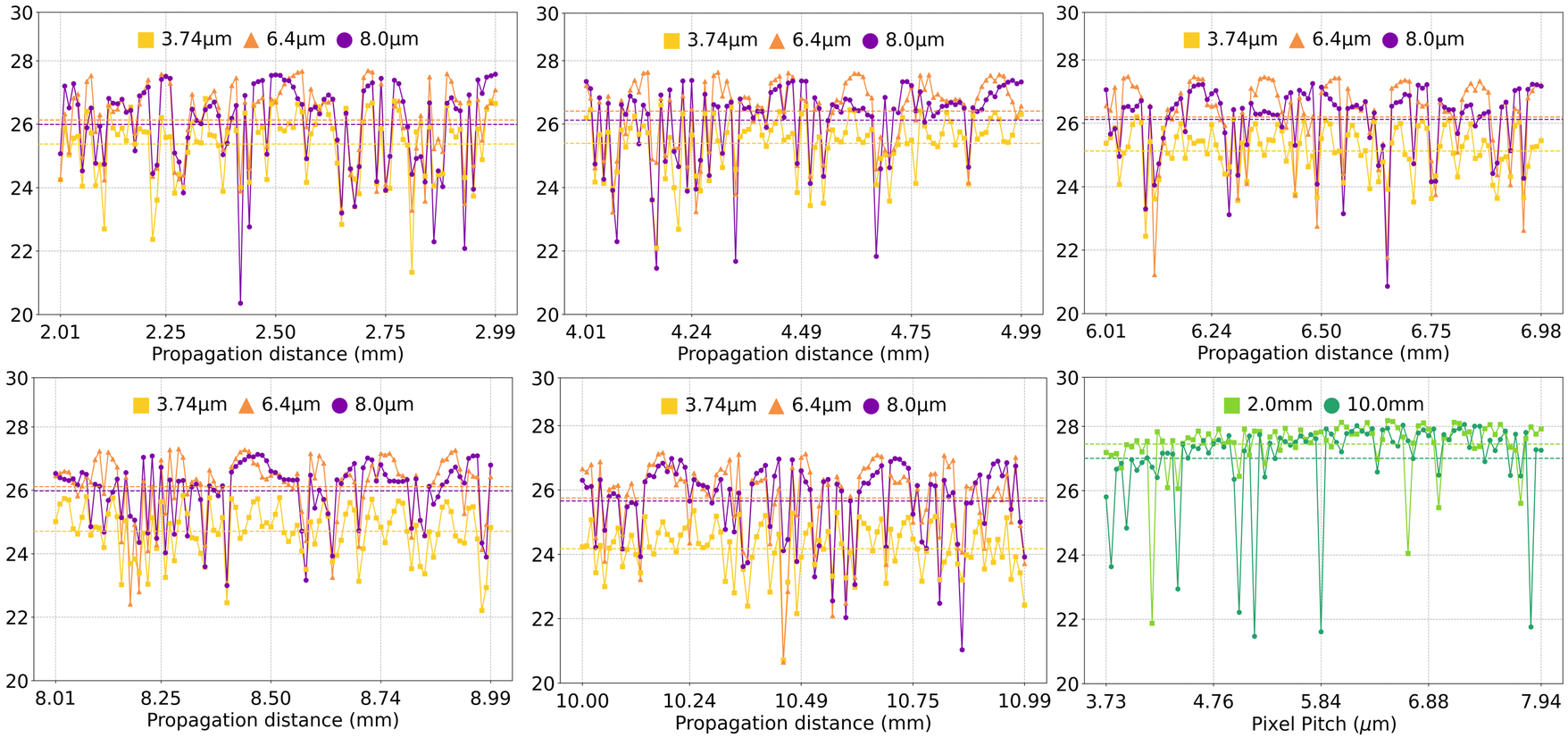}
\caption{PSNR (dB) of the RGB-D conditioned model on 100 DIV2K test images under novel (unseen) $Z$ and $d_x$ settings.
Each point shows the mean PSNR at a randomly sampled $(Z, VD)$ or $(d_x, VD)$ pair.
The top row and bottom-left panels evaluate a continuous 1~mm $Z$ range at pixel pitches $3.74$, $6.4$, and $8.0~\mu$m,
while the bottom-right panel evaluates a continuous $d_x$ range (3.7–8~$\mu$m) at $Z=2.0$ and $10.0~mm$.
The model achieves an average PSNR of ${\approx}26$~dB with ${\approx}1.1$~dB standard deviation, with isolated drops (10–20\%)
due to limited model capacity across the wide conditioning range.}
  \label{fig:PSNR5mm_continue_Z_main_paper}
\end{figure*}

\section{Implementation}
We summarize key training and evaluation settings here, with full implementation details provided in the Supplementary material.
Our training data are sampled from the \SA dataset SA-1B~\cite{kirillov2023segment}, with ground-truth depth generated using MiDaS~\cite{MiDaS}. We select 44,000 images (0.4\% of SA-1B) and synthesize target reconstructions following Kavaklı \etal~\cite{kavakli2023multicolor}. All models are implemented in PyTorch~\cite{pytorch} using Odak~\cite{odak}, and optimized with Adam~\cite{kingma2014adam} ($\beta_{1}=0.9$, $\beta_{2}=0.99$) and a CosineAnnealingLR~\cite{CosineAnnealingLR} scheduler. We use a batch size of 16 and train both models for 20 epochs with an initial \LR of 0.001. Owing to memory constraints, the teacher is trained at $800 \times 800$, while the student is trained at $1024 \times 1024$. All experiments are conducted on eight NVIDIA RTX 3090 GPUs, with a total training and distillation time of four days.

Training a single network to \textit{continuously} adapt across multiple display–scene parameters is computationally intensive,
and auxiliary depth supervision further increases training cost.
To make this trade-off explicit, we (i) train a teacher–student pair to study configurability with RGB-only \3D hologram synthesis, and
(ii) introduce an RGB-D condition variant (depth as input, no depth head) to enable broader conditioning over $\wavelength$, $\pixelPitch$, and long-range $Z$ with reduced training complexity.
\section{Evaluation}
\label{sec:evaluation}

In this section, we aim to provide a comprehensive evaluation of our method and directly provide evidence to answer
\textbf{RQ1} and \textbf{RQ2: Why are MDE and KD Beneficial to CGH?}

\subsection{Evidence to RQ{}1: Quantitative and Qualitative results}
\label{sec:configurability}

\paragraph{Quantitative Analysis.}
\begin{table}[!htbp]
  \centering
  \footnotesize
  \setlength{\tabcolsep}{2pt}
  \renewcommand{\arraystretch}{1.15}
  \begin{tabular}{l c c c c c c}
    \toprule
    \textbf{Method}
    & \textbf{Input}
    & \textbf{PSNR↑}
    & \textbf{SSIM↑}
    & \textbf{LPIPS↓}
    & \textbf{FVVDP↑}
    & \textbf{Conf} \\
    \midrule

    Our Method (teacher)
    & \cellcolor{lightgreen}RGB-only
    & 27.4 & 0.91 & 0.42 & 8.0 & \cellcolor{lightgreen}Yes \\

    Our Method (student)
    & \cellcolor{lightgreen}RGB-only
    & 27.4 & 0.91 & 0.41 & 7.9 & \cellcolor{lightgreen}Yes \\

    Our Method (RGB-D)
    & \cellcolor{lightred}RGB-D
    & 28.2 & 0.93 & \cellcolor{lightyellow!50}\textbf{0.37} & \cellcolor{lightyellow!50}\textbf{8.5} & \cellcolor{lightgreen}Yes \\

    HoloBeam~\cite{aksit2023holobeam}
    & \cellcolor{lightred}RGB-D
    & 25.6 & 0.88 & 0.44 & 8.2 & \cellcolor{lightred}No \\

    Tensor V2~\cite{shi2022end}
    & \cellcolor{lightred}RGB-D
    & 26.5 & \cellcolor{lightyellow!50}\textbf{0.94} & 0.38 & 8.3 & \cellcolor{lightred}No \\

    modified 3D NH~\cite{peng2020neural}
    & \cellcolor{lightred}RGB-D
    & \cellcolor{lightyellow!50}\textbf{28.9} & 0.91 & 0.42 & 8.2 & \cellcolor{lightred}No \\

    Two-stage DepthAny V2
    & \cellcolor{lightred}RGB-D
    & 28.3 & 0.93 & 0.39 & 8.2 & \cellcolor{lightred}No \\

    \bottomrule
  \end{tabular}
  \caption{Averaged quantitative comparison across methods.}
  \label{tbl:avg_holography_evaluation}
\end{table}
As shown in \refTbl{avg_holography_evaluation}, our method achieves on par performance across all metrics.
Although modified 3D NH attains the highest PSNR, \refFig{Tensorv2_NH_ours} shows pronounced blurriness in its reconstructions, resulting in worse perceptual quality compared to our method.
Our approach uniquely supports continuous configurability over display--scene parameters, whereas all baselines require costly retraining per configuration.
Due to the large number of methods and parameter settings, full quantitative results (PSNR, SSIM, LPIPS~\shortcite{zhang2019LPIPS}, FLIP~\shortcite{andersson2020flip}, and
FVVDP~\shortcite{Mantiuk2021FVVDP}) and extended analyses are deferred to Supplementary \refSupSec{Extended_Evaluation_Analysis}.
In summary, \KD preserves fidelity: the student remains comparable to the teacher and existing learned \CGH methods ($\triangle$PSNR<$0.1\%$) while retaining configurability.
Notably, this marginal quality gap enables instant and continuous parameter adjustment at inference time, replacing costly per-configuration retraining (hours) and weight swapping (minutes).
Consistent with iterative CGH pipelines~\cite{kavakli2023multicolor, kavakli2023realistic}, increasing $s$ and $Z$ degrades image quality across methods;
we quantify these effects and report statistical tests in Supplementary \refSupSec{Pairwise_ANOVA}.
\begin{table}[!htbp]
  \centering
  \footnotesize
  \setlength{\tabcolsep}{2.5pt}
  \begin{tabular}{c c c c c c}
  \toprule
  \textbf{Ours} &
  \textbf{$s$} &
  \textbf{$Z$} &
  \textbf{$VD$} &
  \textbf{$\lambda$} &
  \textbf{$d_x$} \\
  \midrule
  RGB-only &
  \cellcolor{lightgreen}\textit{continuous} &
  \cellcolor{lightyellow}\textit{discrete} &
  \cellcolor{lightgreen}\textit{continuous} &
  \cellcolor{lightred}\textit{fixed} &
  \cellcolor{lightred}\textit{fixed} \\
  RGB-D &
  \cellcolor{lightred}\textit{fixed} &
  \cellcolor{lightgreen}\textit{continuous} &
  \cellcolor{lightgreen}\textit{continuous} &
  \cellcolor{lightgreen}\textit{continuous} &
  \cellcolor{lightgreen}\textit{continuous} \\
  \bottomrule
  \end{tabular}
\caption{Configurability of our methods.
\textit{Continuous} denotes generalization to arbitrary values within the range at inference;
\textit{discrete} denotes a fixed set of trained values;
\textit{fixed} denotes a constant parameter.
The RGB-only model prioritizes scene parameters ($s$, $VD$) with discrete $Z$,
while the RGB-D model supports broader display parameters ($\lambda$, $d_x$, $Z$).}
  \label{tbl:scene_params}
\end{table}
\paragraph{Configurability and Qualitative Evaluation.}
\begin{figure*}[h]
  \centering
  \includegraphics[width=1\textwidth]{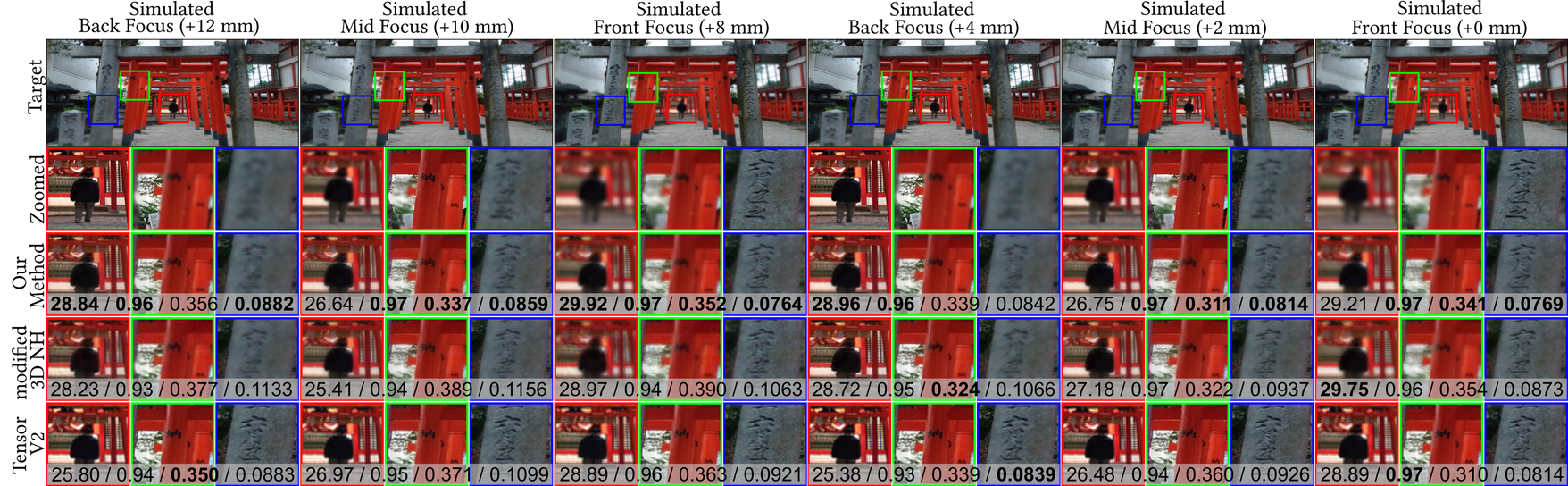}
  \caption{Simulated reconstructions comparing our method, Tensor V2, and modified 3D NH for short and long propagation distances.
  Numbers report PSNR, SSIM, LPIPS, and FLIP, respectively (Source Image: \cite{Temple2022}).}
  \label{fig:Tensorv2_NH_ours}
\end{figure*}
\refTbl{scene_params} summarizes the flexibility of each model variant.
Our RGB-only models continuously support $s$ in $[1.0,1.8]\times$ and $VD$ in $[4,8]~mm$, and support a discrete set of $Z \subseteq \{2,4,7,10\}~mm$, while keeping $d_x$ and $\lambda$ fixed.
This reflects the practical trade-off: \textit{simultaneously covering more parameters continuously will increase training permutations and resource requirements significantly.}
To probe broader display-parameter conditioning with reduced complexity, we use the RGB-D condition model and conduct three examples that support continuous ranges for $Z$, $d_x$, and $\lambda$:
\begin{itemize}[leftmargin=*, nosep]
  \item \textbf{$Z$:} continuous ranges of 2--3, 4--5, 6--7, 8--9, and 10--11~$mm$ (total 5~$mm$) with three $d_x$ values (3.74, 6.4, 8.0~$\mu m$).
  \item \textbf{$d_x$:} continuous range of 3.7--8.0~$\mu m$ with two $Z$ (2.0, 10.0~$mm$).
  \item \textbf{$\lambda$:} continuous ranges of 425--480, 510--565, 625--680~$nm$ with two $d_x$ (3.74, 8.0~$\mu m$) and two $Z$ values (2.0, 10.0~$mm$).
\end{itemize}
All RGB-D training supports continuous $VD$ in $[4,8]~mm$ with fixed $s$.
\refFig{PSNR5mm_continue_Z_main_paper} reports the PSNR distribution for long-range $Z$ and $d_x$ conditioning, with each point evaluated on the same 100 DIV2K images~\cite{Agustsson_2017_CVPR_Workshops_DIV2K}.
The full configuration experiments are included in Supplementary \refSupSec{ParametersRange},
where total of 3,364 novel (unseen) configurations is evaluated across $Z$, $d_x$, and $\lambda$.
Across these studies, we observe that conditioning difficulty is not uniform: $s$ and $VD$ can be learned more smoothly, while long-range $Z$ and display parameters
($d_x$, $\lambda$) require broader permutations and stronger inductive bias (e.g., our ASM CNN block for long $Z$).
Overall, our model maintains consistent quality (average PSNR $\approx 26$) with low variance (average std $\approx 1.1$) across randomly generated parameters.
\refFig{different_display_main} shows captured reconstructions from two of our three holographic display prototypes with different configurations,
verifying our method under hardware setups; extra captured results are provided in Supplementary \refSupSec{Hardware_Results}.
Notably, our three prototypes span pixel pitches of 3.74, 6.4, and 8.0~$\mu$m, directly validating our configurability claim.
\begin{figure*}[htbp!]
  \centering
  \includegraphics[width=1\textwidth]{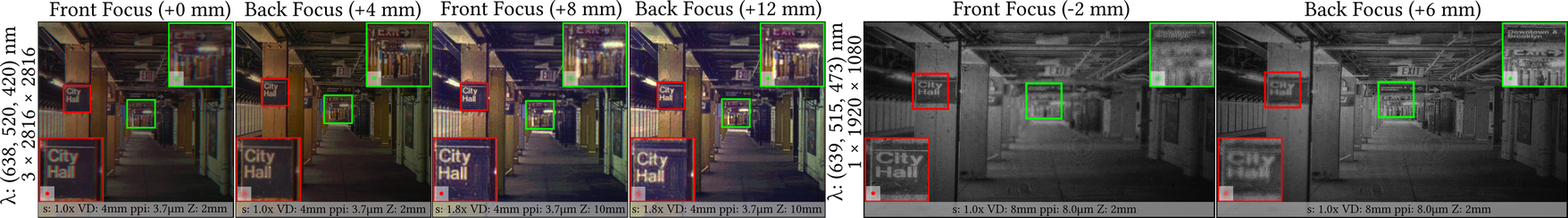}
  \caption{Captured results from Jasper and Holoeye SLMs (2.1× pixel pitch difference) under varied parameters (source image: \cite{subway2023}).}
  \label{fig:different_display_main}
\end{figure*}
Additionally, \refFig{Tensorv2_NH_ours} compares reconstructions of our method, Tensor V2, and modified 3D NH across short and long $Z$.
Compared with the other methods that requires retraining, our configurable model achieves competitive image quality while preserving correct focus/defocus cues across depth planes.
To construct the modified 3D NH, we reimplement NH~\cite{peng2020neural} and extend it to support RGB-D inputs.

\subsection{RQ2: Why are MDE and KD Beneficial to CGH?}
\label{sec:depth_ness}
\subsubsection{Depth Estimation}
For RGB-only \3D hologram synthesis, depth ambiguity directly leads to incorrect focus/defocus cues across planes.
Such errors are perceptually obvious even when image metrics like PSNR change only slightly.
In particular, single-stage RGB-only learned \CGH methods that do not estimate depth during training (e.g., Holobeam~\cite{aksit2023holobeam}) may produce plausible \emph{2D} reconstructions but fail to generate \3D holograms with correct focus (see \refFig{student_fail}).
\textit{Our key insight is that jointly learning \MDE and hologram synthesis enables depth-aware hologram generation from RGB-only inputs.}
Specifically, the predicted monocular depth is normalized to the reconstruction range $Z \pm \tfrac{VD}{2}$, compressing the focal stack into the target volume.
As in prior methods~\cite{shi2021towards, shi2022end, aksit2023holobeam}, we do not model eyepiece-to-diopter mapping, as it is orthogonal to SLM-plane synthesis.
This makes our method applicable to photos, videos, and live streaming, where depth is unavailable.

Our depth estimation is intended as an auxiliary task for hologram synthesis, not as a replacement for dedicated \MDE methods~\cite{yang2024depth}.
Our model is trained on 44K images, whereas state-of-the-art \MDE methods use tens of millions.
Despite this large gap, the depth head serves as a proof of concept that joint learning improves \3D hologram quality from RGB-only inputs.
Therefore, we do not report standalone depth estimation metrics.
\paragraph{Training Data Strategy.}
Because \CGH requires significantly less data than \MDE to converge~\cite{shi2022end}, our joint training must account for this imbalance.
We use SA-1B~\cite{kirillov2023segment} for its scale and resolution, generating depth pseudo-labels with MiDaS~\cite{MiDaS}.
Our framework is agnostic to the specific depth estimator and can naturally benefit from future improvements;
further discussion is provided in Supplementary~\refSupSec{training_data_strategy}.
\begin{figure}[t!]
  \centering
  \includegraphics[width=0.99\columnwidth]{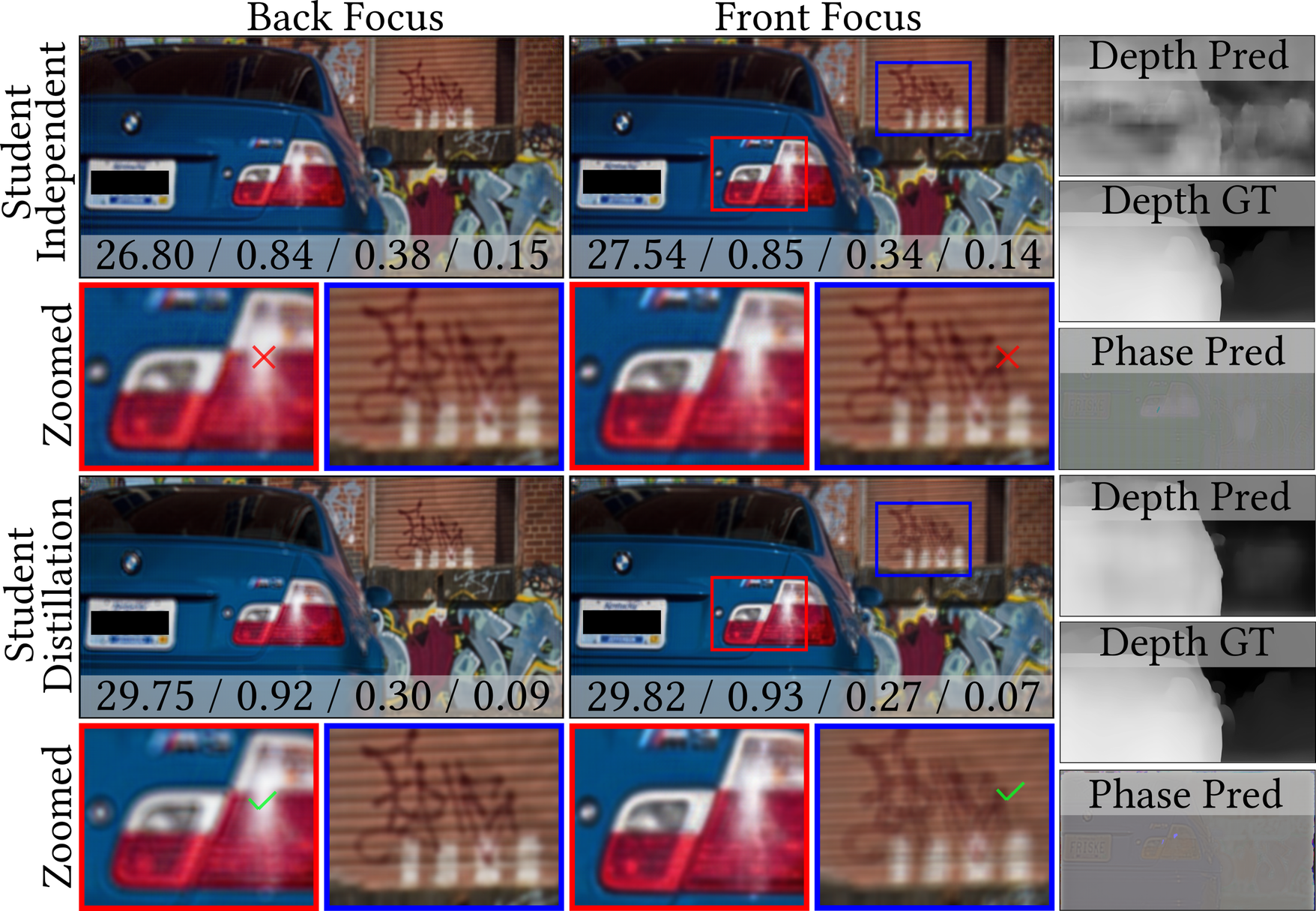}
  \caption{Comparison of reconstructions, depth, and phase predictions between the independently trained student mode
  and the distilled student model. Numbers report PSNR, SSIM, LPIPS, and FLIP, respectively. (Source Image: \cite{Bluecar2008})}
  \label{fig:student_fail}
\end{figure}
\subsubsection{Knowledge Distillation}
\label{sec:KD}
KD helps us to create smaller and faster models for hologram synthesis that could not otherwise be trained effectively from scratch.
\refFig{student_fail} compares an independently trained student with our distilled student.
When trained from scratch with the same setting as distillation, the independent student struggles to produce accurate depth maps, which leads to incorrect focus/defocus behavior and degraded reconstruction quality.
In contrast, the distilled student demonstrates improved depth estimation, better color preservation, and higher image quality.
Additional failure cases are provided in Supplementary \refSupSec{KD_analysis}.

\paragraph{Inference Time.}
Distilling the teacher into a student model results in a great improvement in speed.
\refFig{infer_time} compares inference time across models and resolutions.
Our student model is consistently faster than all baselines, achieving 2$\times$ speed-up compared to Tensor V2~\cite{shi2022end}.
Specifically, our student model is 41\%, 46\%, and 44\% faster than \textit{Tensor V2}; and 17\%, 14\%, and 17\% faster than \textit{modified 3D NH}, respectively.
The only method faster than our student is Holobeam~\cite{aksit2023holobeam};
however, as demonstrated in \refFig{student_fail},
excluding \MDE as an auxiliary task leads to failures in reproducing correct focus cues in \3D hologram.
Our student model can be further accelerated by removing the ASM CNN block; however, this component is essential for predicting holograms with $Z$ larger than 4~$mm$.
All timings are measured on an NVIDIA A100 40G GPU under \fp32 with PyTorch.
\begin{figure}[htbp]
  \centering
  \includegraphics[width=0.99\columnwidth]{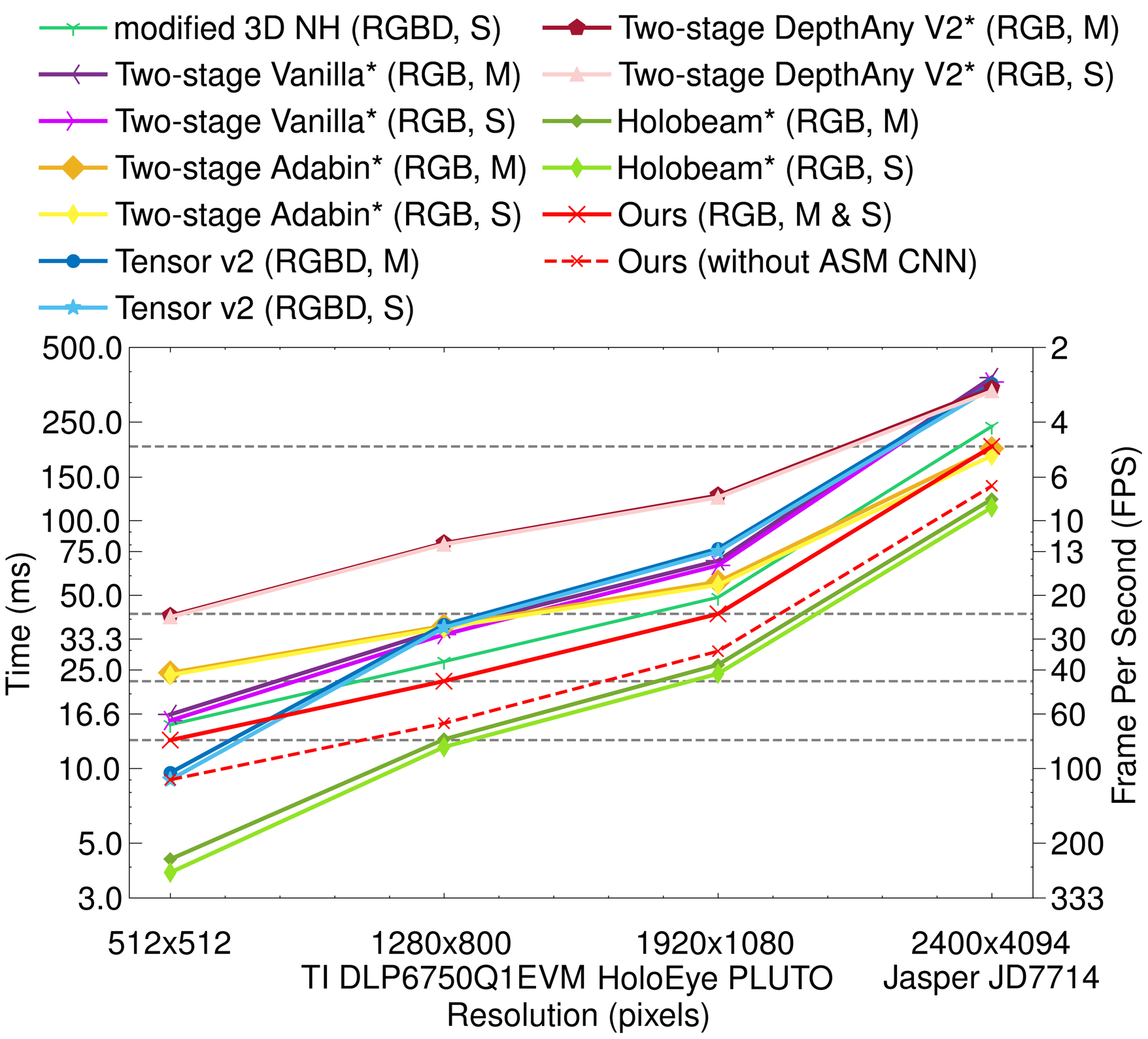}
  \caption{Inference time comparison across models. \textit{S} denotes single-color hologram and \textit{M}
  denotes multi-color hologram synthesis using our light head module.
  \textit{Two-stage} methods first estimate depth from RGB inputs and then generate \3D holograms.
  \textit{Vanilla} uses U-Net* for both stages; \textit{Adabin} employs Adabin~\cite{SFBhats2020AdaBin}; and \textit{DepthAny V2}
  uses DepthAnything V2~\cite{yang2024depth} for depth estimation. \textsuperscript{*}Same architecture used in
  ~\cite{aksit2023holobeam, eybposh2020deepcgh, Goi2020Optica, Wu2021Optica, Horisaki2021Optica}.}
  \label{fig:infer_time}
\end{figure}

\section{RQ3: Limitation and Future Work}
\label{sec:discussion}

\paragraph{Out-of-distribution Behavior.}
Extrapolation beyond the trained parameter ranges can degrade performance sharply (e.g., training over $Z$ in 2--11~$mm$ and evaluating at 12~$mm$).
This limitation is relevant to training cost: expanding parameter range coverage by brute-force sampling becomes exponentially expensive.
A core open problem for configurable holography is achieving robust generalization to novel configurations \emph{without} densely training over them, especially for propagation distance under arbitrary volume depth in 3D hologram synthesis.

\paragraph{Defocus Accuracy.}
The defocus accuracy of reconstructions in our method remains strongly correlated with depth estimation accuracy; we provide error examples and their analysis in Supplementary \refSupSec{additional_analyses}.
We emphasize that our depth estimation is introduced solely as an auxiliary signal to improve hologram synthesis, rather than to compete with dedicated \MDE models such as DepthAnything V2~\cite{yang2024depth}; Accordingly, we do not include standalone depth estimation metrics.
While the \3D accuracy in our method are greatly improved compared with Holobeam, the image quality remains slightly lower to RGB-D condition model.
This gap is primarily attributable to training scale: our depth head is trained on only 44k images, whereas state-of-the-art \MDE models typically rely on supervision from tens of millions of images.
Future variants can therefore directly benefit from advances in \MDE, enabling better geometric regularization and defocus in RGB-only \3D hologram synthesis.

\paragraph{Fluctuating Quality Across Conditioned Ranges.}
As shown in \refFig{PSNR5mm_continue_Z_main_paper}, we observe that reconstruction quality for \3D holograms
can fluctuate as conditioning variables vary within a supported range (see Supplementary \refSupSec{ParametersRange} for detailed analysis).
The underlying cause is not yet fully understood.
Compared with iterative methods~\cite{kavakli2023multicolor}, which optimize holograms under a fixed configuration with stable image quality, configurable holography must implicitly approximate a \textit{global optimal solution} across a wide range of optical configurations.
\textit{Finding such a solution is inherently several orders of magnitude more complex than optimizing for a single configuration}, which partially explains the observed quality fluctuations.
Narrowing the conditioning range reduces these fluctuations, as shown in Supplementary~\refSupSec{ParametersRange}.
Improving stability in feed-forward configurable methods may benefit from information driven inductive biases~\cite{2024InformationDrivenNeurips} or more carefully designed loss functions than naive per-pixel supervision~\cite{Chu2025RealTime}.

\paragraph{Training Cost and Data Trade-offs.}
A single model that continuously adapts over multiple display-scene parameters is expensive to train,
and incorporating auxiliary depth estimation further greatly increases data and computational requirements.
As a result, we must trade off between (i) variety of configurability,
(ii) range of configurability, and (iii) depth supervision diversity.
In this paper, we trained a teacher-student pair as proof of concept to validate the feasibility of efficient and configurable RGB-only \3D hologram synthesis.
Additionally, we train an RGB-D condition model (without a depth head) to demonstrate how far and wide configurability
itself can extend under reduced computational complexity.
We acknowledge that due to the limited resources, using network variants to validate different aspects of our approach is not ideal.
The computational cost of dense training becomes prohibitive as parameter ranges expand: distilling a unified RGB-only student spanning 10 mm $Z$, 6 $\mu$m $d_x$, the visible spectrum, 10 mm $VD$, and 1–1.8× $s$ would require 16 A100 GPUs and ~20 training days due to exponentially growing parameter permutations.
Accordingly, while our network is more efficient than existing learned \CGH methods, it should be viewed as a prototype rather than the perfect solution for configurable holography.
Future work should pursue unified architectures that jointly support broad configurability and depth-free inference, while avoiding dense parameter permutations through improved training and sampling strategies.

\paragraph{Different Hologram Types and Hardware Non-idealities.}
Our work assumes an ideal \SLM with \textit{smooth phase} and does not model non-idealities such as pixel fill-factor,
crosstalk, higher-order diffraction, or SLM tilt. Although we provide extensive simulated results,
broad hardware generalization and validation remain challenging because of the large number of combinations in the display hardware design.
We therefore focus more on simulation and evaluate on three commonly used systems with similar optical architectures.
Practical systems span a much wider design space—including
wearable near-eye displays~\cite{kim2022Siggraph}, waveguide combiners~\cite{jang2024waveguide}, and beam splitters—each introducing additional aberrations,
stray light, and non-uniform degradation.
At present, our framework does not yet support this extended design space.
Further investigation is required for generalization across fundamentally different optical architectures,
hardware calibration~\cite{peng2020neural, choi2021neural},
and configurable random-phase generation~\cite{Chu2025RealTime, chao2025random} in future work.

\section{Conclusion}
We acknowledge that our research is based on well-established techniques, including networks, \MTL, \MDE, and \KD.
Our contribution lies not in them, but in (i) formulating \emph{configurability} as a concrete objective for learned \CGH,
(ii) identifying which display–scene parameter can be continuously conditioned within a single model and which remain challenging, and
(iii) providing, to our knowledge, the first empirical study quantifying the trade-offs between speed, parameter range, input requirements,
conditioning difficulty, and image quality across thousands of novel configurations,
(iv) revealing \MDE as a beneficial auxiliary task for \CGH.
We believe these insights provide a useful reference for future configurable \CGH research beyond any specific architecture.
Our final goal is to build a \CGH model that is continuously configurable and can adapt to novel configurations outside the training set without dense retraining.
In other words, we seek a physically accurate \CGH method that understands light propagation robustly across display and scene variations.

\begin{backmatter}


\bmsection{Disclosures}
The authors declare no conflicts of interest.


\end{backmatter}

\newpage

\renewcommand{\refSec}[1]{Sec.~S\ref{sec:#1}}
\renewcommand{\refSupSec}[1]{Sec.~S\ref{supplementary:#1}}
\renewcommand{\refFig}[1]{Fig.~S\ref{fig:#1}}
\renewcommand{\refFigFull}[1]{Figure~S\ref{fig:#1}}
\renewcommand{\refEq}[1]{Eq.~S(\ref{eq:#1})}
\renewcommand{\refTbl}[1]{Tbl.~S\ref{tbl:#1}}
\renewcommand{\refObj}[1]{Objective~S\ref{obj:#1}}

\section*{Supplementary 1}
\section{Phase Profile and Experimental Setup}
\label{supplementary:Other}

\subsection{Phase Image Example}

\refFig{phase_img} shows the example phase profile our model predicted.

\begin{figure}[t!]
  \centering
  \includegraphics[width=0.99\columnwidth]{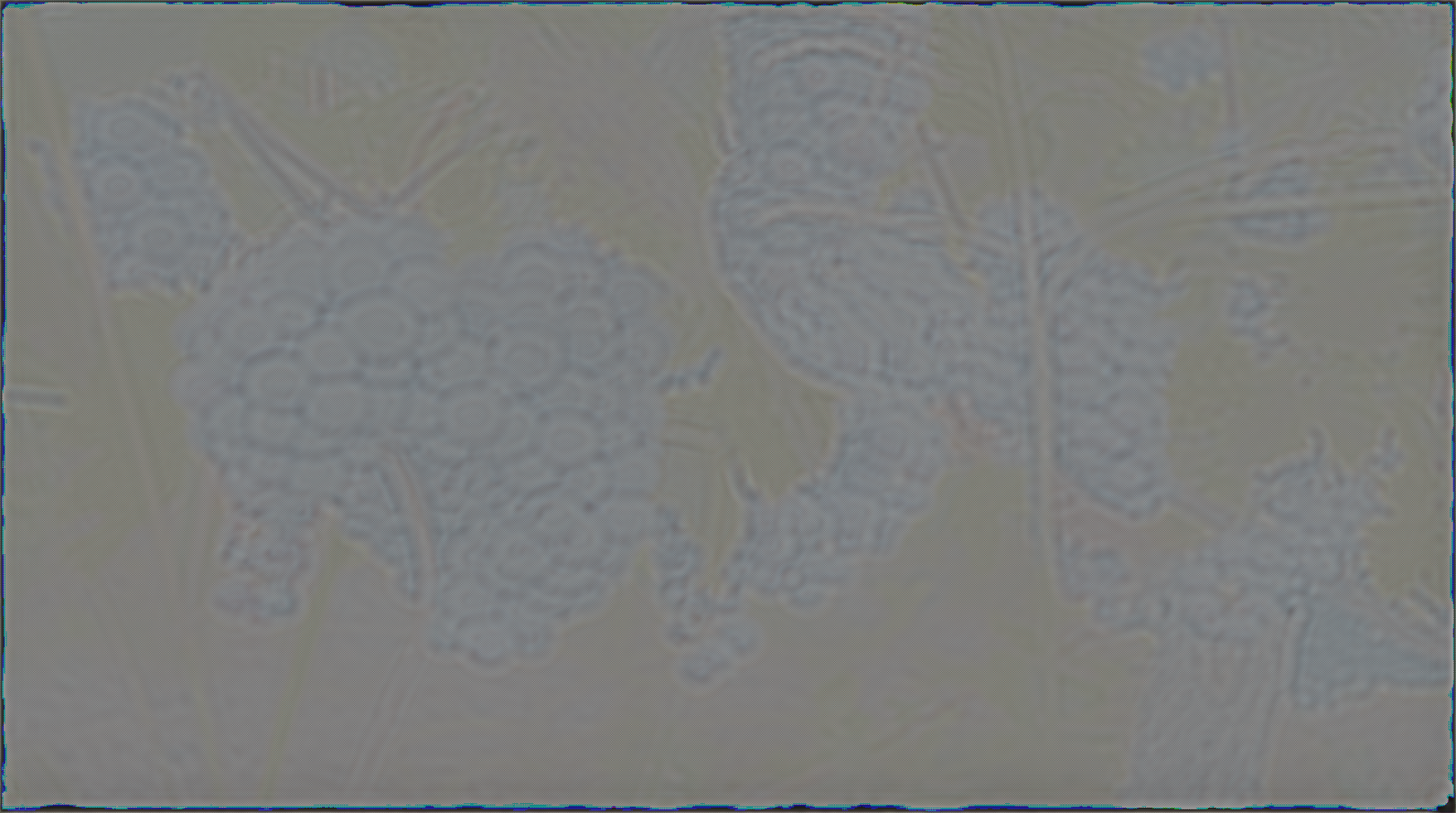}
  \caption{The example of our  phase-only hologram.}
  \label{fig:phase_img}
\end{figure}

\subsection{Hardware Image}

\refFig{hardware_jasper} and \refFig{hardware_holoeye} show the photographs of two of the three holographic display prototypes we used in this paper.
The optical path of our display prototype begins with a laser light source (LASOS MCS4), which integrates three individual laser lines.
The emitted light from a single-mode fibre is collimated using a Thorlabs LA1708-A plano-convex lens with a 200~mm focal length.
This linearly polarized, collimated beam is then directed by a beamsplitter (Thorlabs BP245B1) toward our phase-only \SLM,
the Holoeye Pluto-VIS (1920×1080~px, 8~$\mu$m), Holoeye LETO (1920×1080~px, 6.4~$\mu$m), or Jasper JD7714 (2400×4094, 3.74~$\mu$m).
The modulated beam subsequently passes through a lens system comprising Thorlabs LA1908-A and LB1056-A,
with focal lengths of 500~mm and 250~mm, respectively. Following this, a pinhole aperture (Thorlabs SM1D12)
is positioned at the focal plane of the lenses. Finally, we capture the holographic reconstructions using a lensless image sensor
(Point Grey GS3-U3-23S6M-C USB 3.0), which is mounted on an X-stage (Thorlabs PT1/M) with a travel range of 0 to 25~mm and a
positioning precision of 0.01~mm.

\begin{figure*}[t!]
  \centering
  \begin{subfigure}{0.45\textwidth}
    \centering
    \includegraphics[width=\textwidth]{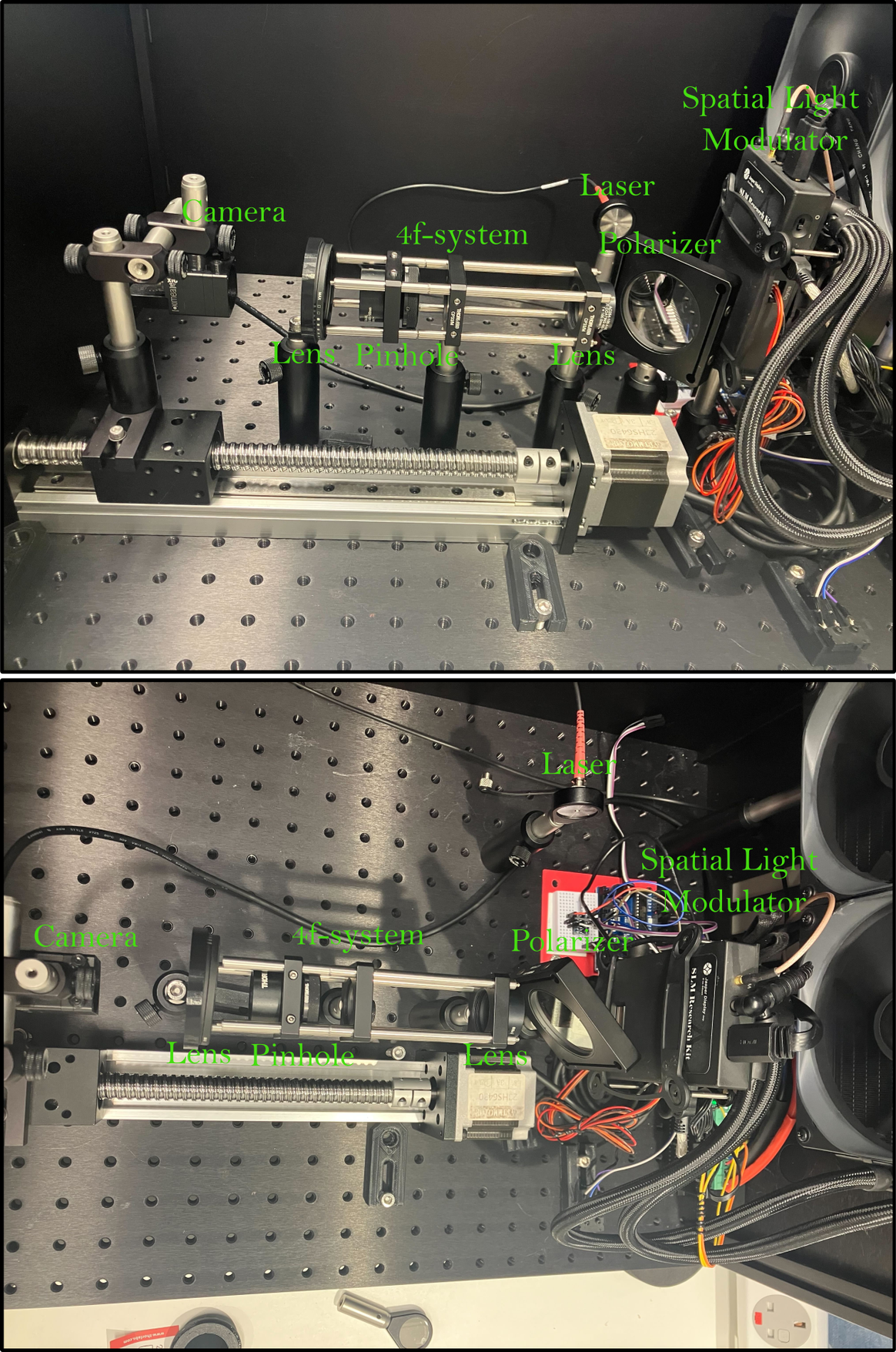}
    \caption{First holographic display prototype: Jasper JD7714.}
    \label{fig:hardware_jasper}
  \end{subfigure}
  \hfill
  \begin{subfigure}{0.45\textwidth}
    \centering
    \includegraphics[width=\textwidth]{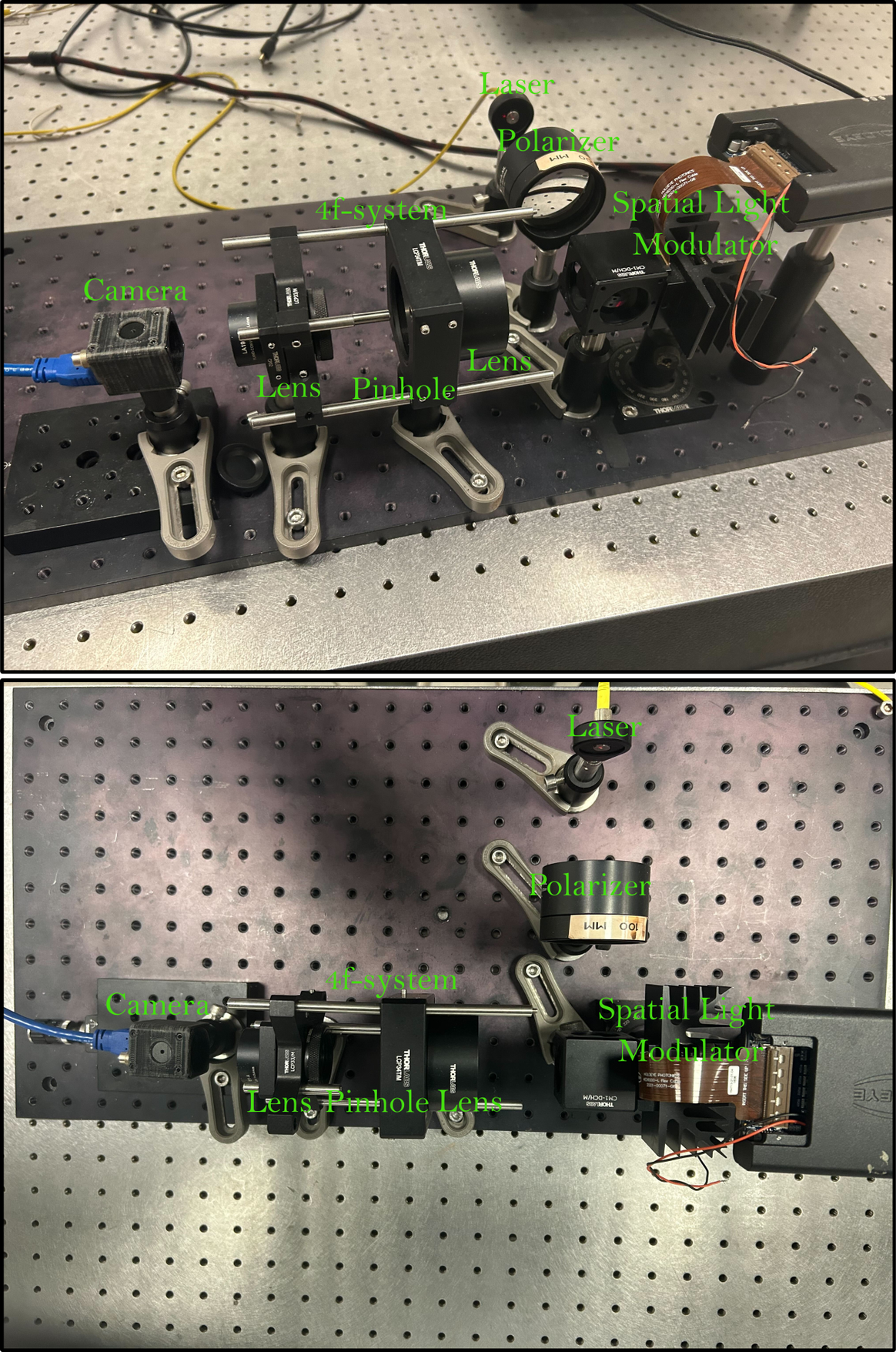}
    \caption{Second holographic display prototype: Holoeye Pluto-VIS.}
    \label{fig:hardware_holoeye}
  \end{subfigure}
  \caption{Hardware prototypes used in evaluation: (a) Jasper JD7714 and (b) Holoeye Pluto-VIS.}
  \label{fig:hardware_prototypes}
\end{figure*}

\section{Diffraction's Scalability Property}
\label{supplementary:diffraction_scalability}

A hologram computed for one wavelength can be reused for another by scaling the propagation distance as $Z_2 = \frac{\lambda_2}{\lambda_1} \times Z_1$.
\refFig{PSF_diff} illustrates this property: the two PSF patterns are visually similar despite different $(Z, \wavelength)$ pairs, because the Fresnel number is preserved under this scaling.
A related scaling connects pixel pitch and propagation distance as $Z_2 = (\frac{p_1}{p_2})^2 \times Z_1$.
These properties apply to \textit{single-color} settings where a fixed-wavelength model can be reused across wavelengths by adjusting $Z$.
However, they do not resolve multi-color cases (Eq.~2 in the main paper), where wavelengths interact through a joint optimization objective and a single rescaling factor is insufficient.
Given the scalability property, we deliberately include a single pixel pitch in our RGB-only training variable set to control training permutations and computational cost, while using the RGB-D condition variant to study broader pixel pitch conditioning.

\begin{figure}[t!]
  \centering
  \includegraphics[width=0.99\columnwidth]{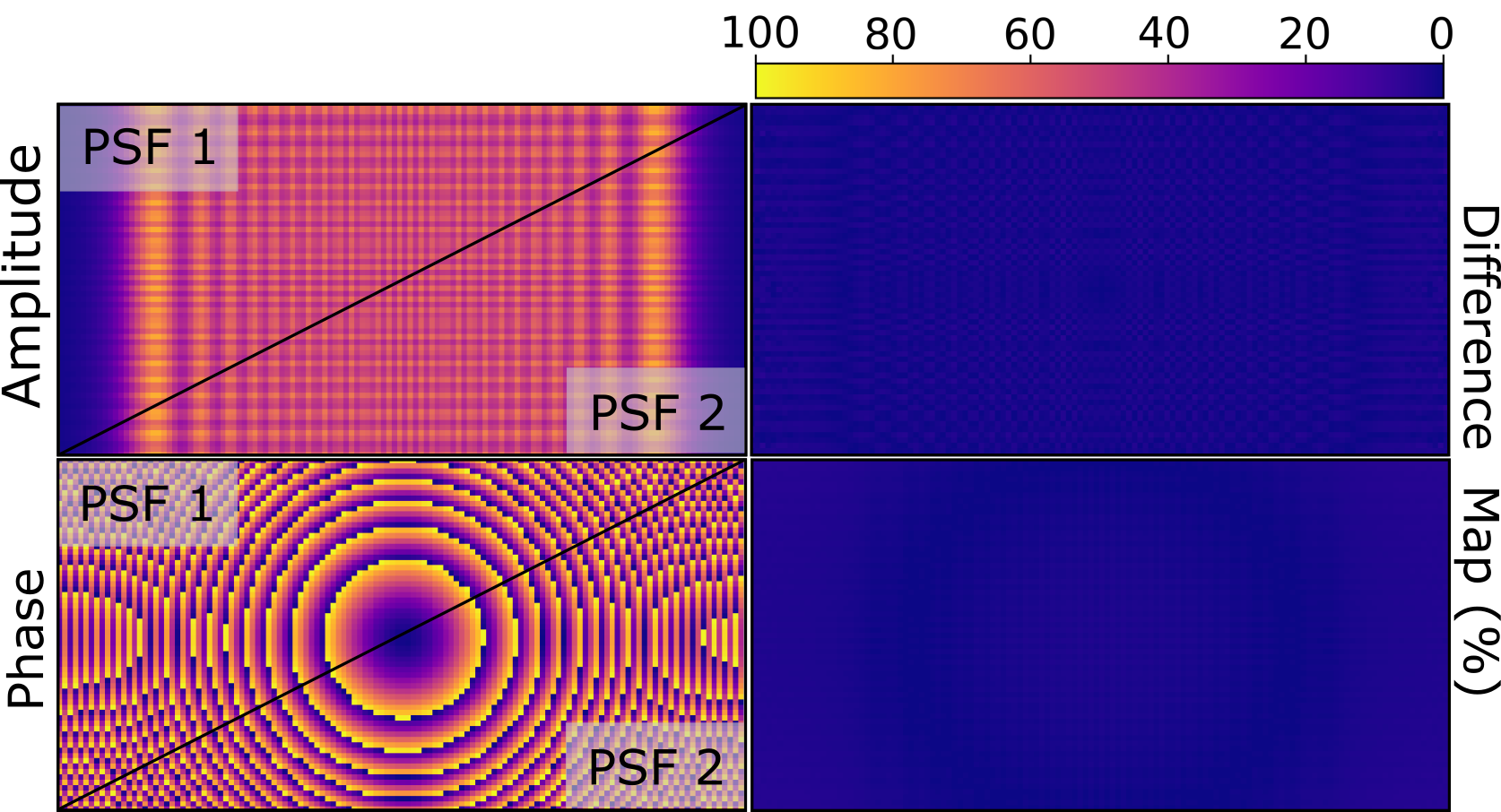}
  \caption{Diffraction scalability. Scaling propagation distance by the wavelength ratio produces similar \PSF patterns.
  \PSF~1: $Z = 3~mm$, $\wavelength = 515~nm$; \PSF~2: $Z = 3.723~mm$, $\wavelength = 416~nm$.
  Both use $d_x = 3.74~\mu m$. The amplitude and phase remain highly consistent,
  while the difference maps on the right stay close to zero, showing that the main diffraction structure is the same.}
  \label{fig:PSF_diff}
\end{figure}

\section{Model}
\label{supplementary:Model}

\begin{table*}[!htbp]
  \centering
  \begin{threeparttable}
    \footnotesize
    \setlength{\tabcolsep}{4pt}
    \resizebox{\textwidth}{!}{
    \begin{tabular}{m{3.7cm} m{1.2cm} m{0.5cm} m{0.7cm} m{0.6cm} m{0.2cm} m{0.3cm} m{1.0cm} m{0.68cm} m{0.55cm} m{1.25cm} m{0.75cm} m{0.8cm} m{0.7cm}}
      \toprule
      \multirow{2}{*}{\textbf{}} &
      \multirow{2}{*}{\textbf{Input}} &
      \multirow{2}{*}{\parbox{0cm}{\centering \textbf{Conf}}} &
      \multirow{2}{*}{\parbox{0cm}{\centering \textbf{Speed}}} &
      \multicolumn{3}{c}{\textbf{Hologram}} &
      \multirow{2}{*}{\parbox{0cm}{\centering \textbf{SLM\\Refresh\\Rate}}} &
      \multirow{2}{*}{\centering\textbf{Stage}} &
      \multirow{2}{*}{\parbox{0cm}{\centering \textbf{3D}}} &
      \multirow{2}{*}{\parbox{0cm}{\centering \textbf{Depth\\Accuracy}}} &
      \multirow{2}{*}{\textbf{Learned}} &
      \multicolumn{1}{l}{\multirow{2}{*}{\parbox{0.88cm}{\centering \textbf{Max Z\\(mm)}}}}&
      \multirow{2}{*}{\parbox{0cm}{\centering \textbf{VD (mm)}}}
      \\ \cmidrule(lr{.75ex}){5-7}
      & & & & \textbf{Type} & \textbf{PD} & \textbf{DP} & & & & & & &\\
      \addlinespace[0.15em]

      \midrule
      \renewcommand{\arraystretch}{1}
      Our Method                                                          & \cellcolor{lightgreen}{RGB-only}   & \cellcolor{lightgreen}Yes     & \cellcolor{lightgreen}Fast           & \cellcolor{lightgreen}M+S      & 8        & \cellcolor{lightgreen}No   & 60\hspace{1mm}Hz  & \cellcolor{lightgreen}Single  & \cellcolor{lightgreen}True    & \cellcolor{lightyellow}Moderate  & \cellcolor{lightgreen}Full & \textasciitilde10.0  & \textasciitilde8.0  \\
      \parbox{3.6cm}{HoloBeam\cite{aksit2023holobeam}}                    & \cellcolor{lightred}\parbox[c][10pt][c]{1cm}{G-only}     & \cellcolor{lightred}No        & \cellcolor{lightgreen}Fast           & \cellcolor{lightred}S          & 8        & \cellcolor{lightred}Yes    & 60\hspace{1mm}Hz  & \cellcolor{lightgreen}Single  & \cellcolor{lightgreen}True    & \cellcolor{lightred}Inaccurate   & \cellcolor{lightgreen}Full & \textasciitilde0.0   & \textasciitilde6.0  \\
      \parbox{3.6cm}{Multi-DNN \cite{Yoshi2023Optics}}                    & \cellcolor{lightgreen}{RGB-only}   & \cellcolor{lightred}No        & \cellcolor{lightred}Slow             & \cellcolor{lightred}S          & 8        & \cellcolor{lightred}Yes    & 60\hspace{1mm}Hz  & \cellcolor{lightred}Three     & \cellcolor{lightgreen}True    & \cellcolor{lightyellow}Moderate  & \cellcolor{lightgreen}Full & \textasciitilde50.0  & \textasciitilde2.0  \\
      \parbox{2.7cm}{NH \cite{peng2020neural}}                            & \cellcolor{lightgreen}{RGB-only}   & \cellcolor{lightred}No        & \cellcolor{lightgreen}Fast           & \cellcolor{lightred}S          & 8        & \cellcolor{lightgreen}No   & 60\hspace{1mm}Hz  & \cellcolor{lightred}Two       & \cellcolor{lightred}False     & \cellcolor{lightgreen}Accurate   & \cellcolor{lightgreen}Full & \textasciitilde100.0 & \textasciitilde0.0 \\
      \parbox{3.6cm}{NH3D \cite{choi2021neural}}                         & \cellcolor{lightred}\parbox[c][10pt][c]{1cm}{RGB-D}        & \cellcolor{lightred}No        & \cellcolor{lightred}Slow             & \cellcolor{lightred}S          & 8        & \cellcolor{lightgreen}No   & 60\hspace{1mm}Hz  & \cellcolor{lightred}Two       & \cellcolor{lightgreen}True    & \cellcolor{lightgreen}Accurate   & \cellcolor{lightred}Semi   & \textasciitilde8.2   & \textasciitilde4.4  \\
      \parbox{3.6cm}{TensorV2 \cite{shi2022end}}                          & \cellcolor{lightred}\parbox[c][10pt][c]{1cm}{RGB-D}        & \cellcolor{lightred}No        & \cellcolor{lightgreen}Fast           & \cellcolor{lightred}S          & 8        & \cellcolor{lightred}Yes    & 60\hspace{1mm}Hz  & \cellcolor{lightred}Two       & \cellcolor{lightgreen}True    & \cellcolor{lightgreen}Accurate   & \cellcolor{lightgreen}Full & \textasciitilde12.0  & \textasciitilde6.0  \\
      \parbox{3.6cm}{DGE-CNN \cite{liu2023dge}}                           & \cellcolor{lightred}\parbox[c][10pt][c]{1cm}{RGB-D}        & \cellcolor{lightred}No        & \cellcolor{lightred}Slow             & \cellcolor{lightred}S          & 8        & \cellcolor{lightgreen}No   & 58\hspace{1mm}Hz  & \cellcolor{lightred}Two       & \cellcolor{lightgreen}True    & \cellcolor{lightyellow}Moderate  & \cellcolor{lightred}Semi   & \textasciitilde10.0  & \textasciitilde30.0 \\
      \parbox{3.7cm}{4K-DMDNet \cite{Kexuan2023}}                         & \cellcolor{lightred}\parbox[c][10pt][c]{1cm}{RGB-D}        & \cellcolor{lightred}No        & \cellcolor{lightred}Slow             & \cellcolor{lightred}S          & 8        & \cellcolor{lightgreen}No   & 58\hspace{1mm}Hz  & \cellcolor{lightred}Two       & \cellcolor{lightred}False     & \cellcolor{lightgreen}Accurate   & \cellcolor{lightgreen}Full & \textasciitilde300.0 & \textasciitilde0.0 \\
      \parbox{3.7cm}{Time-multiplexed \cite{choi2022time}}               & \cellcolor{lightred}\parbox[c][10pt][c]{1cm}{RGB-D}        & \cellcolor{lightred}No        & \cellcolor{lightred}Slow             & \cellcolor{lightred}S          & 4        & \cellcolor{lightgreen}No   & 480\hspace{1mm}Hz & \cellcolor{lightred}Two       & \cellcolor{lightgreen}True    & \cellcolor{lightgreen}Accurate   & \cellcolor{lightred}Semi   & \textasciitilde79.0  & \textasciitilde12.0 \\
      \bottomrule
    \end{tabular}}
  \caption{Comparison of hologram synthesis methods. Our method generates both single-color (S) and multi-color (M) holograms from an RGB-only
  input for a preferred set of display-scene parameters. In \textit{Input}, G-only denotes green-channel-only. In \textit{Speed},
  Fast and Slow denote \(>10\) FPS and \(<10\) FPS, respectively, at 1920 $\times$ 1080 resolution. In \textit{Learned}, \textit{Full}
  denotes a fully learning-based method, while \textit{Semi} denotes a hybrid of learning and optimization. NH \cite{peng2020neural}
  is fully learning-based, whereas NH3D \cite{Choi2021neural} uses learning only for hologram refinement. \textit{Conf}, \textit{PD},
  \textit{DP}, Z, VD, and \textit{SLM} denote Configurable, pixel depth, double phase encoding~\cite{DoublePhase}, maximum propagation distance,
  scene volume depth, and Spatial Light Modulator, respectively.}    \label{tbl:comparison}
  \end{threeparttable}
\end{table*}

\subsection{Model Structure}
\label{supplementary:Model_structure}
\subsubsection{Teacher Model}

\refTbl{arch-details} summarizes the architecture details of both the teacher and student models.

\begin{table}[!htbp]
\centering
\footnotesize
\renewcommand{\arraystretch}{0.82}

\begin{minipage}[t]{0.48\columnwidth}
\centering
\begin{threeparttable}
\resizebox{\linewidth}{!}{%
\begin{tabular}{@{}l c c c@{}}
\toprule
\multicolumn{4}{c}{\textbf{Teacher}} \\
\midrule
\textbf{Module} & \textbf{Layer} & \textbf{Ch.} & \textbf{Match} \\
\midrule
\multirow{5}{*}{\parbox{1.0cm}{\centering \textbf{Enc.}\\ \textit{EffNet-b1}\\ \textit{6.51M}}}
& \#0 & 16$\to$32   & Dec \#0 \\
& \#1 & 32$\to$24   & Dec \#1 \\
& \#2 & 24$\to$40   & Dec \#2 \\
& \#3 & 40$\to$112  & Dec \#3 \\
& \#4 & 112$\to$320 & Dec \#4 \\
\midrule
\multirow{5}{*}{\parbox{1.0cm}{\centering \textbf{Dec.+FPN}\\ \textit{4.17M}}}
& \#4 & 320$\to$272 & Enc \#4 \\
& \#3 & 272$\to$176 & Enc \#3 \\
& \#2 & 176$\to$112 & Enc \#2 \\
& \#1 & 112$\to$88  & Enc \#1 \\
& \#0 & 88$\to$60   & Enc \#0 \\
\midrule
\textbf{Heads} & Phase & 60$\to$3 & -- \\
& Depth & 60$\to$1 & -- \\
& Laser & 60$\to$9 & -- \\
\bottomrule
\end{tabular}%
}
\caption{Teacher model.}
\label{tbl:teacher-details}
\end{threeparttable}
\end{minipage}
\hfill
\begin{minipage}[t]{0.48\columnwidth}
\centering
\begin{threeparttable}
\resizebox{\linewidth}{!}{%
\begin{tabular}{@{}l c c c@{}}
\toprule
\multicolumn{4}{c}{\textbf{Student}} \\
\midrule
\textbf{Module} & \textbf{Layer} & \textbf{Ch.} & \textbf{Match} \\
\midrule
\multirow{5}{*}{\parbox{1.0cm}{\centering \textbf{Enc.}\\ \textit{MBV3-S}\\ \textit{0.93M}}}
& \#0 & 16$\to$16   & Dec \#0 \\
& \#1 & 16$\to$16   & Dec \#1 \\
& \#2 & 16$\to$24   & Dec \#2 \\
& \#3 & 24$\to$48   & Dec \#3 \\
& \#4 & 48$\to$576  & Dec \#4 \\
\midrule
\multirow{5}{*}{\parbox{1.0cm}{\centering \textbf{Dec.}\\ \textit{1.2M}}}
& \#4 & 576$\to$336 & Enc \#4 \\
& \#3 & 336$\to$192 & Enc \#3 \\
& \#2 & 192$\to$112 & Enc \#2 \\
& \#1 & 112$\to$72  & Enc \#1 \\
& \#0 & 72$\to$52   & Enc \#0 \\
\midrule
\textbf{Heads} & Phase & 52$\to$3 & -- \\
& Depth & 52$\to$1 & -- \\
& Laser & 52$\to$9 & -- \\
\bottomrule
\end{tabular}%
}
\caption{Student model.}
\label{tbl:student-details}
\end{threeparttable}
\end{minipage}
\end{table}

\paragraph{\FPN Structure}

Following the U-Net decoder, we aggregate features from multiple decoder stages using an \FPN\cite{lin2017feature}. Each decoder output $D_{i}$ is fed into its corresponding \FPN layers according to their level. Each \FPN layer in our model consists of a convolutional layer, batch normalization, and a nonlinear activation function ReLU followed by a bilinear upsample layer and an additional convolutional layer. Each \FPN layer upsamples feature map by a scale of 2.
Each decoder output is projected to a common channel dimension and progressively upsampled to the full resolution. To upsample every decoder output to the same scale, the $D_{i}$ will be processed iteratively by \FPN layers. For example, consider the first decoder output $Dec 4$ of size $D_4$x$\frac{H}{32}$x$\frac{W}{32}$, the $Dec 4$ will be processed by \FPN layers five times. Each time the resolution is doubled, resulting in an output that is the same size $D_0$x$H$x$W$. The aligned features are then fused (summation after lightweight convolutions) to obtain a single latent code that is shared by all prediction heads.

\begin{figure}[htbp]
  \centering
  \includegraphics[angle=90,width=0.38\columnwidth]{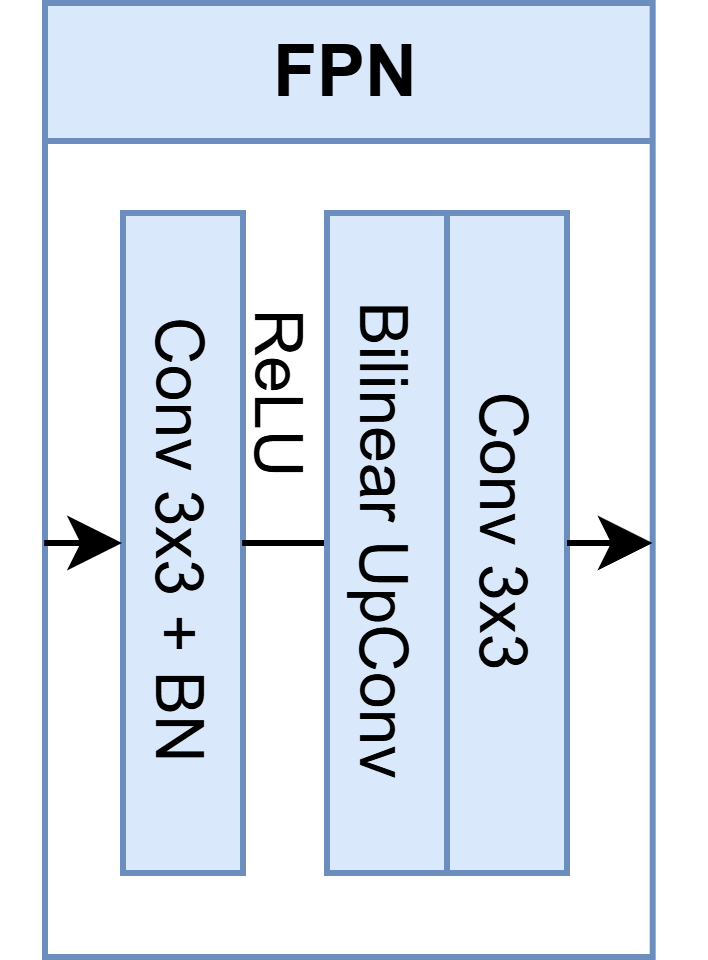}
  \caption{Overview of the \FPN aggregation module.}
  \label{fig:FPN}
\end{figure}

\paragraph{Phase Prediction Layer Structure}
\label{supplementary:Phase_Prediction_Layer_Structure}

The phase prediction head is a Conv2d layer with no activation function, allowing the model to infer phase values without explicit constraints. Following the phase head are bandlimited ASM kernel and ASM CNN. We found that our model will only converge when extra learnable parameters are provided after the bandlimited ASM method during training, which makes ASM CNN essential for long propagation distance estimation. The ASM CNN contains a convolutional layer, followed by three consecutive convolutional layers, batch normalization, and ReLU.
The phase head predicts a complex-valued field and uses band-limited \ASM, with an additional refinement block for long propagation distances. A similar structure is also used in Tensor V2, where the second-stage network is used after free space propagation. This observation highlights that long propagation distance prediction is inherently more challenging and requires extra prior as part of the network to adapt to the long-distance light propagation better. Please also note that our model only supports direct phase encoding.

\begin{figure}[htbp]
  \centering
  \includegraphics[width=0.99\columnwidth]{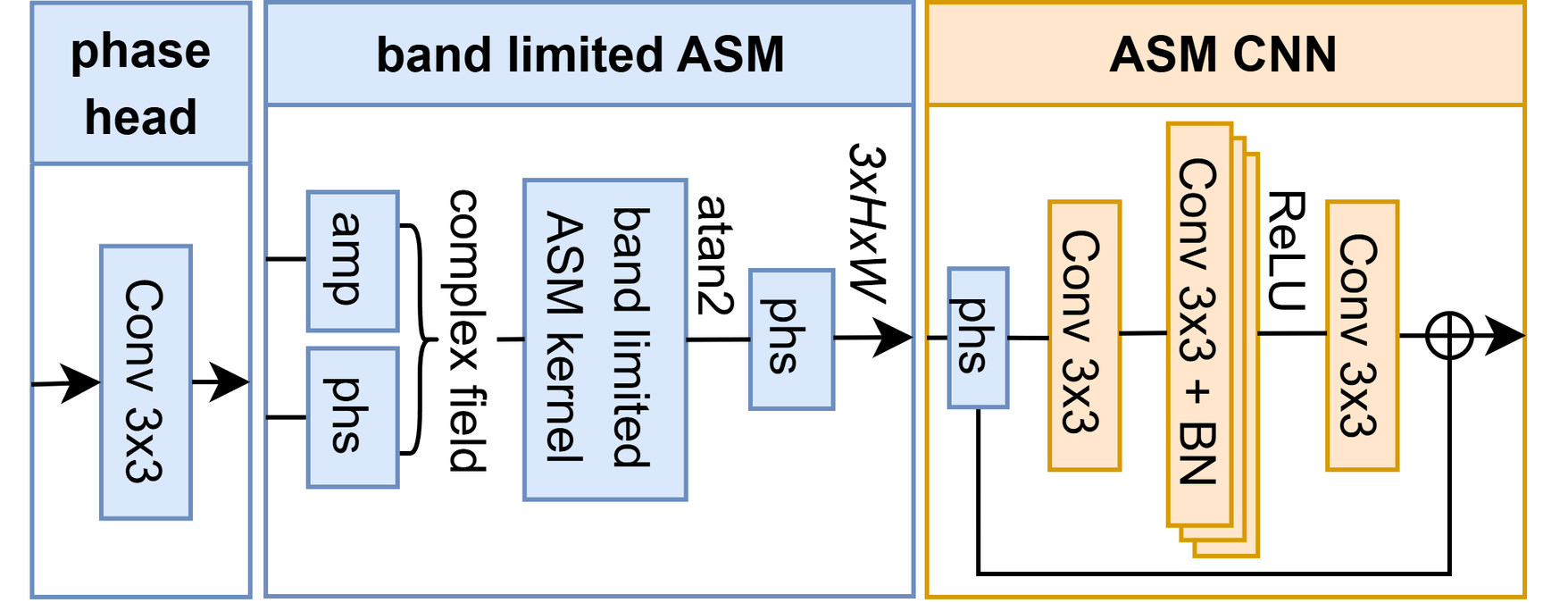}
  \caption{Overview of the phase head module.}
  \label{fig:phaseHead}
\end{figure}

\paragraph{Light Prediction Layer Structure}

The light prediction layer aggregates information in the model prediction $out$ and predicts light intensity. The light head aggregates spatial features and predicts per-(subframe, color primary) light powers constrained to $[0,1]$.
Given the output of the model of size $D_0$x$H$x$W$, first, the convolutional layers will downsample it by a scale of 4. Then, an adaptive pooling operation is applied to reduce the spatial dimensions to $D_0$x$1$x$1$. The pooled data will be fed into two linear layers, which manipulate the channel dimension to yield a value of 9, resulting in a $9$x$1$x$1$ feature map. The final light intensity prediction $laser$ is obtained by reshaping this output to a $1$x$3$x$3$ matrix. This design ensures an efficient and accurate prediction of light intensity from the decoder output. Since the data range of light intensity is between 0 and 1, Sigmoid is used to ensure the model output is appropriately scaled.

\begin{figure}[htbp]
  \centering
  \includegraphics[width=0.65\columnwidth]{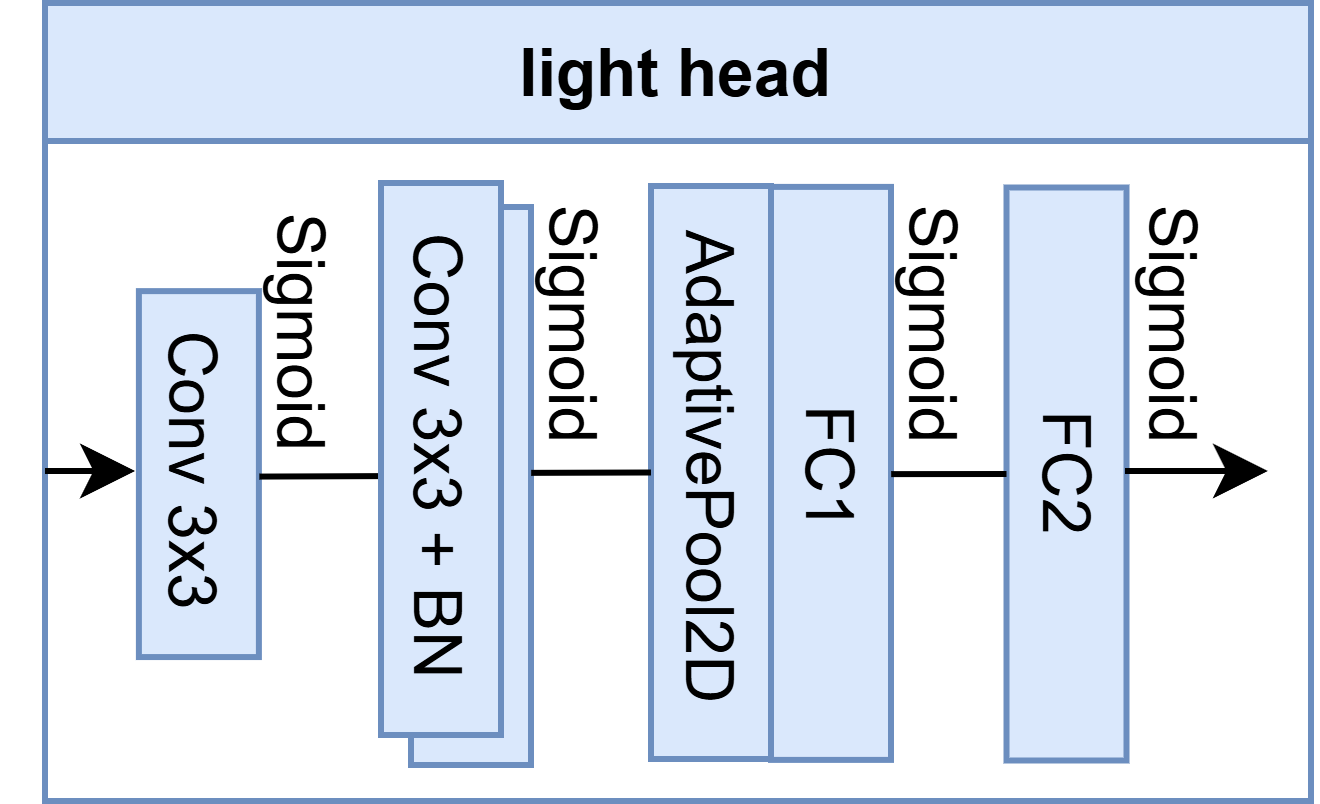}
  \caption{Overview of the light head module.}
  \label{fig:lightHead}
\end{figure}

\paragraph{Depth Prediction Layer Structure}

The depth prediction layer consists of a \BCP and a depth head, following the PixelFormer implementation~\cite{pixelformer2022}.
It adopts a bin-based formulation with \PSP and a bin-center predictor.
\noindent \textit{Bin Center Predictor: }
Given a feature map $E_4$ of size $E_4 \times \frac{H}{32} \times \frac{W}{32}$,
\PSP applies adaptive global pooling at scales $\{1,2,3,6\}$. The pooled feature is then passed to the \BCP,
following \cite{agarwal2022attention}. By predicting bin-center maps, \BCP reformulates
depth estimation from pure regression into a classification-regression task.
\noindent \textit{Depth Prediction Head: }
Given a feature map $D_4$ of size $D_4 \times \frac{H}{32} \times \frac{W}{32}$,
the depth head takes the latent feature and \BCP output, then multiplies them along the channel dimension to
produce the final depth.

\section{Loss Functions}
\label{supplementary:Loss}

Our training objective combines four loss components: flip loss, reconstruction loss, light loss, and depth loss. We provide the mathematical formulations below.

\subsection{Flip Loss}

We employ FLIP loss~\cite{andersson2020flip} to improve color accuracy by penalizing perceptual color differences in IPT space. We simplify the original formulation by setting $\Delta feature = 0$:
\begin{equation}
  \mathcal{L}_{flip} = \sum_{i} \Delta color(I_{p_i}, I_{t_i}),
\end{equation}
where $\Delta color$ computes the redistributed Euclidean distance in IPT color space between predicted $I_p$ and target $I_t$.

\subsection{Reconstruction Loss}

The reconstruction loss $\lossRecon$ is shown below:
\begin{align}
   \lossRecon &= m_0L_2(rec_k, target_k) + m_1L_2(rec_k * mask_k, target_k * mask_k) \nonumber \\
   &+ m_2L_2(rec_k * target_k, target_k * target_k) + L_smooth,
\end{align}
where $m_0$, $m_1$, and $m_2$ represent weights, $rec_k$, $mask_k$ and $target_k$ represent the reconstructed image,
binary mask, and target image at the kth plane. Weighted with $m_0$, $L_2(rec_k, target_k)$ is the L2 norm that evaluates the entire image with respect to a target image, $m_1 * L_2(rec_k * mask_k, target_k * mask_k)$ applies a mask to the reconstructed and target images before computing the L2 norm. Finally, the $m_2 * L_2(rec_k * target_k, target_k * target_k)$ multiplies the reconstructed and target images together before computing the L2 norm, which emphasizes the regions of the target image that have high values.
Additionally, $L_smooth$ refers to phase smoothing loss, we used multi-scale TV loss introduced by~\cite{kavakli2023multicolor} to ensure the smoothness in phase prediction,
and regularization loss introduced by~\cite{shi2022end} to constrain the standard deviation and mean value of phase prediction to be close to 0.

\subsection{Light Loss}

Following~\cite{kavakli2023multicolor}, we constrain laser intensity per frame through four regularization terms. Let $channel_{sum}[i] = \sum_{c} laser[i, c]$ for frame $i \in \{0,1,2\}$:
\begin{align}
  \mathcal{L}_{mean} &= \frac{1}{N} \sum_{i} (channel_{sum}[i] - peak_{amplitude})^{2}, \\
  \mathcal{L}_{abs} &= \frac{1}{N} \sum_{i} \left| channel_{sum}[i] - peak_{amplitude} \right|, \\
  \mathcal{L}_{amax} &= \sum_{j} (\max(recons[j]) - \textstyle\sum_{dim=0}(peak_{amplitude}))^2, \\
  \mathcal{L}_{rgb} &= \sum_{c,i} (\max(peak_{amplitude}[c]) - peak_{amplitude}[c,i]),
\end{align}
where $j$ indexes depth frames, $c$ indexes color primaries, and $i$ indexes subframes.
$\mathcal{L}_{mean}$ and $\mathcal{L}_{abs}$ penalize deviations of per-frame channel sums from the target amplitude; $\mathcal{L}_{amax}$ constrains maximum reconstructed intensity; and $\mathcal{L}_{rgb}$ encourages the three primaries to follow the R-G-B pattern.
The combined light loss is $\lossLight = \gamma (\mathcal{L}_{mean} + \mathcal{L}_{abs} + \mathcal{L}_{rgb} + \mathcal{L}_{amax})$ with $\gamma = 1\times10^5$.

\subsection{Depth Loss}

The depth loss function consists of three sub-loss functions: $\lossScaleInvariant$, $\lossGradingMatching$, and $\lossTV$. $\lossScaleInvariant$ represents the scale-invariant loss invented by eigen et al. \cite{eigen2014depth}, $\lossGradingMatching$ represents the gradient matching loss, which compares edges of estimated depths with ground truth depth maps, and $\lossTV$ represents the total variant loss, which smoothes the edge of objects in the depth map.

$\lossScaleInvariant$: Given predicted depth $\widehat{d_i}$ and ground truth $d_i^*$ at ith pixel, the logarithmic distance between $log(\widehat{d_i})$ and $log(d_i^*)$ is calculated as $D_i = log(\widehat{d_i}) - log(d_i^*)$,
calculated as:
\begin{equation}
\label{eq:sil_loss}
      \mathcal{L}_{silog} =  \alpha \Biggl(\frac{1}{n} \sum_{i}(D_i)^2  - \frac{ \lambda }{n^2}\Bigl(\sum_{i}D_i\Bigl)^2 \Biggl)
\end{equation}

$\lossGradingMatching$: We compute image gradients using the Sobel operator in horizontal (x) and vertical (y) directions. For predicted depth $\widehat{d}$ and ground truth $d^*$, gradient magnitudes are:
\begin{equation}
    \widehat{G} = \sqrt{\widehat{G}_{x}^{2} + \widehat{G}_{y}^{2}}, \quad {G^*} = \sqrt{{G_{x}^*}^{2} + {G_{y}^*}^{2}},
\end{equation}
where $\widehat{G}_x$, $\widehat{G}_y$ are gradients of $\widehat{d}$, and ${G_{x}^*}$, ${G_{y}^*}$ are gradients of $d^*$. The gradient matching loss is:
\begin{equation}
\label{eq:gm_loss}
\lossGradingMatching = \beta \frac{1}{n} \sum_{i}(\widehat{G_i} - G_i^{*})^2,
\end{equation}
where $\beta$ is the weight and $n$ is the total number of pixels.

$\lossTV$: Given the predicted depth $\widehat{d}$, we first compute the adjacent gradients in both the x and y directions $g\_x$ and $g\_y$:
\begin{gather}
    \text{g\_x} = \widehat{d}[:, :, 1:, :] - \widehat{d}[:, :, :-1, :] \\
    \text{g\_y} = \widehat{d}[:, :, :, 1:] - \widehat{d}[:, :, :, :-1]
\end{gather}
The edge-smoothing loss $\lossTV$ is defined as the sum of the mean absolute values of these gradients, where $n$ denotes the total number of pixels.
\begin{equation}
\label{eq:tv_loss}
\lossTV = \frac{1}{n} \sum_{i}|\text{g\_x\_i}| + \frac{1}{n} \sum_{i}|\text{g\_y\_i}|
\end{equation}
The final $\lossDepth$ is defined as the sum of the $\lossScaleInvariant$, $\lossGradingMatching$ and $\lossTV$.
\begin{equation}
  \label{eq:depth_loss}
  \lossDepth = \lossScaleInvariant + \lossGradingMatching + \lossTV
\end{equation}

\section{Non-configurability of Iterative Methods}
\label{supplementary:SGD_configurable}

Iterative hologram optimization solves for a phase-only hologram under a fixed propagation kernel.
To test whether such methods can be extended to support multiple configurations simultaneously,
we optimize either: (i) a hologram with SGD under a single fixed propagation distance, or (ii) a single hologram jointly over four propagation distances $Z \in \{1, 2, 3, 4\}$~mm by summing the reconstruction losses across all four settings.

\refFig{SGD_configurable} shows the resulting reconstructions and holograms.
Panel~(a) presents reconstructions obtained by SGD with a fixed $Z$.
The optimized hologram yields clean results at the target setting, with correct front/back focus behavior.
Panel~(b) presents reconstructions obtained when SGD jointly optimizes a single hologram over four different $Z$ values.
In this case, the reconstructions at all distances exhibit severe noise and loss of correct focus, indicating strong interference among the inconsistent optimization objectives.
Panel~(c) compares the corresponding holograms from (a) and (b):
the hologram optimized for multiple $Z$ values is visibly corrupted with fringes, which explains the degraded reconstructions across all focal settings.

These results show that iterative methods cannot produce a single shared hologram that remains valid across different propagation configurations.
The phase updates required by one propagation kernel conflict with those required by others, so changing the configuration necessitates re-optimizing the hologram from scratch.
This makes iterative optimization fundamentally non-configurable.

\begin{figure*}[t!]
  \centering
  \includegraphics[width=1\textwidth]{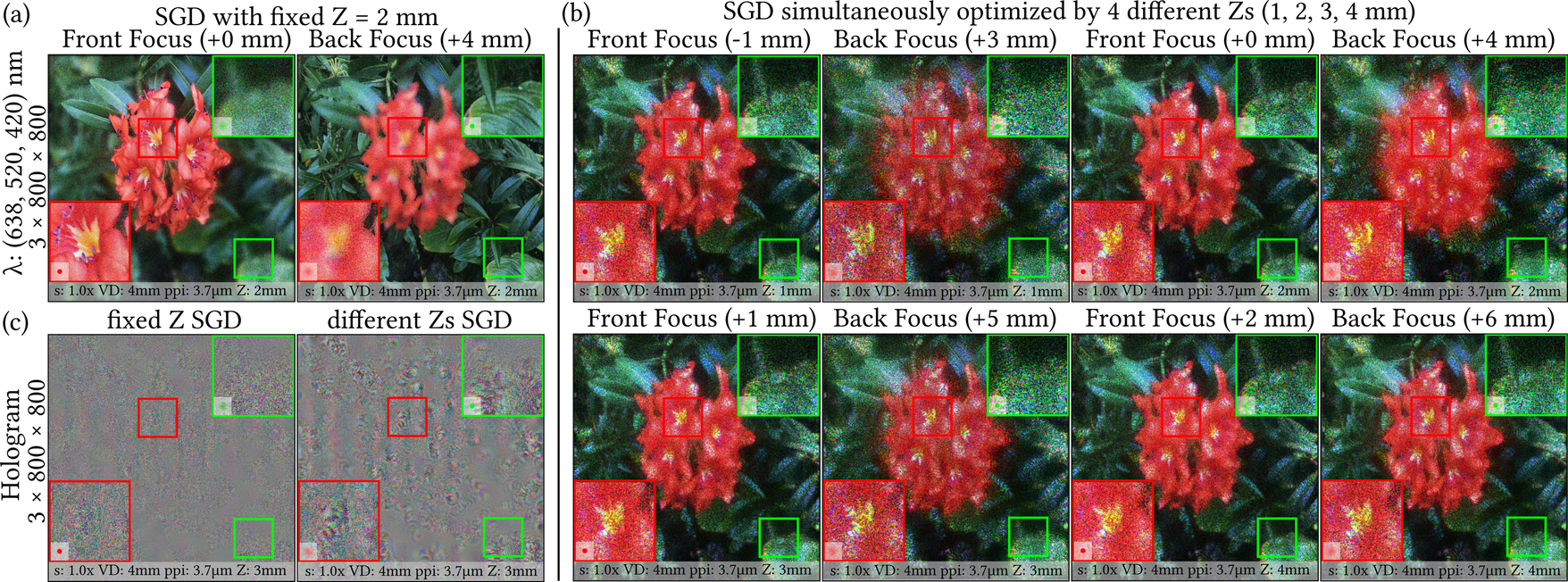}
  \caption{Empirical evidence that SGD-based hologram optimization is non-configurable.
  (a)~Reconstructions from SGD with a fixed $Z$.
  (b)~Reconstructions from jointly optimizing one hologram over four propagation distances ($Z=1,2,3,4$~mm), causing severe artifacts at all settings.
  (c)~Corresponding holograms from (a) and (b) (Source Image: \cite{mildenhall2019llff}).}
  \label{fig:SGD_configurable}
\end{figure*}

\section{\KD and \CGH Analysis}
\label{supplementary:KD_analysis}

\subsection{Failure Cases Of Individually Trained Student Model}
\refFig{student_fail_more} compares the independently trained student with the distilled student.
Although both use the same training setting, the independently trained student produces less accurate depth maps, leading to weaker focus/defocus effects and lower reconstruction quality.
By contrast, the distilled student achieves better depth estimation, color preservation, and overall image quality.
\begin{figure*}[t!]
  \centering
  \includegraphics[width=1\textwidth]{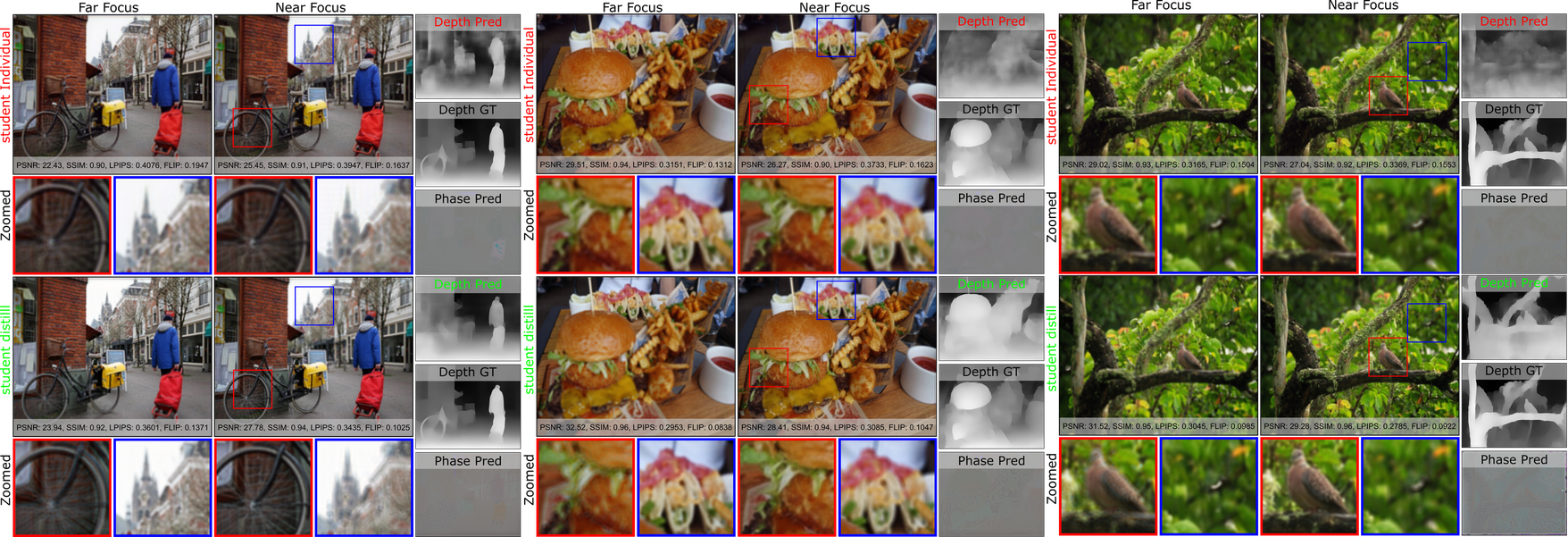}
  \caption{Comparison of reconstructions, phase, and depth between the independently trained student model (in red) and the distilled student model (in green).
  From left to right, top to bottom: (Source Image: \cite{cake2009}, \cite{Street2014}, and \cite{Burger2014})}
  \label{fig:student_fail_more}
\end{figure*}

\section{Training Data Strategy}
\label{supplementary:training_data_strategy}

Unlike depth estimation, which requires extensive training data to learn robust geometric priors across diverse scenes, \CGH is fundamentally a mapping task that reflects diffraction physics.
Consequently, \CGH networks exhibit significantly lower data requirements for convergence and generalization.
Prior work~\cite{shi2022end} has demonstrated that typical learned \CGH models can achieve robust performance with only around 1,000 training images, as the primary learning objective is to capture the wavelength- and distance-dependent light transport kernel rather than high-level scene understanding.
Given this asymmetry, our joint learning framework faces a design trade-off: while \CGH alone requires less data, incorporating \MDE as an auxiliary task demands much larger and more diverse training data.
To address this, we select the Segment Anything dataset SA-1B~\cite{kirillov2023segment}, which provides two advantages.
First, SA-1B contains tens of millions of high-resolution images with diverse content, satisfying the data scale and variability required for effective depth estimation.
Second, unlike common depth datasets that are limited to lower resolutions (\eg NYU Depth V2~\cite{SilbermanECCV12} at $640\times480$), SA-1B provides high-resolution images suitable for hologram synthesis training.
For ground-truth depth supervision, we employ MiDaS~\cite{MiDaS} to generate pseudo-labels.
We acknowledge that MiDaS is not the \SOTA in \MDE and exhibits known limitations, including scale ambiguity and reduced accuracy in complex scenes.
However, as our goal is to demonstrate the \emph{feasibility} of depth-assisted RGB-only \3D hologram synthesis rather than advancing \MDE itself, MiDaS provides sufficient supervision quality for this proof of concept.
Our framework is independent of the specific depth estimator used for pseudo-labeling and naturally inherits improvements from future depth models.

\section{Extended Evaluation Table}
\label{supplementary:Extended_Evaluation_Table}
Tbl.~\ref{supplmentary:table_holography_evaluation} provides the complete quantitative evaluation at different holography settings, including all conditions supported by our teacher and student models.

\begin{table*}[!htbp]
  \centering
  \begin{threeparttable}
    \footnotesize
    \setlength{\tabcolsep}{2.5pt}
    \renewcommand{\arraystretch}{0.95}
    \begin{tabular}{
      m{2.1cm}
      m{1.0cm}
      l
      c  c
      c  c
      m{0.8cm}
      m{0.8cm}
      m{1.0cm}
      m{1.1cm}
      m{0.8cm}
    }
    \hline
    \textbf{Method} & \textbf{Input} & \parbox{5cm}{\centering \textbf{Display-scene Parameters}}
    & \multicolumn{2}{c}{\textbf{PSNR↑ (dB)}} & \multicolumn{2}{c}{\textbf{SSIM↑}}
    & \textbf{LPIPS↓} & \textbf{FLIP↓} & \textbf{FVVDP↑} & \parbox{1.1cm}{\textbf{Parameters}} & \textbf{Speed}\\
    \addlinespace[0.15em]
    \cline{4-10}
    & & & Mean & Std & Mean & Std & Mean & Mean & Mean & & \\
    \hline
    \multirow{24}{*}{\parbox{2cm}{Our Method \\ (teacher)}}
    & \multirow{24}{*}{RGB-only}
      & \parbox{5cm}{\centering $\pixelPitch$: 3.74 $\mu m$, Z: 2 mm, VD: 4 mm, 1.0x}
      & \cellcolor{lightgreen}\textbf{29.33} & 2.73 & 0.95 & 0.03 & 0.35 & \cellcolor{lightgreen}\textbf{0.10} & 8.30 & \multirow{24}{*}{10.74 M} & \multirow{24}{*}{651 ms} \\

    & & \parbox{5cm}{\centering $\pixelPitch$: 3.74 $\mu m$, Z: 2 mm, VD: 4 mm, 1.4x}
      & 28.78 & 2.84 & 0.94 & 0.03 & 0.36 & 0.11 & 8.23 & &\\

    & & \parbox{5cm}{\centering $\pixelPitch$: 3.74 $\mu m$, Z: 2 mm, VD: 4 mm, 1.8x}
      & 27.65 & 2.80 & 0.94 & 0.04 & 0.36 & 0.11 & 8.20 & &\\

    & & \parbox{5cm}{\centering $\pixelPitch$: 3.74 $\mu m$, Z: 4 mm, VD: 4 mm, 1.0x}
      & 28.16 & 3.07 & 0.94 & 0.03 & 0.38 & 0.12 & 8.18 & &\\

    & & \parbox{5cm}{\centering $\pixelPitch$: 3.74 $\mu m$, Z: 4 mm, VD: 4 mm, 1.4x}
      & 28.15 & 2.99 & 0.94 & 0.03 & 0.38 & 0.11 & 8.19 & &\\

    & & \parbox{5cm}{\centering $\pixelPitch$: 3.74 $\mu m$, Z: 4 mm, VD: 4 mm, 1.8x}
      & 27.86 & 2.97 & 0.93 & 0.04 & 0.39 & 0.11 & 8.18 & &\\

    & & \parbox{5cm}{\centering $\pixelPitch$: 3.74 $\mu m$, Z: 7 mm, VD: 4 mm, 1.0x}
      & 27.80 & 3.06 & 0.93 & 0.03 & 0.40 & 0.12 & 8.11 & &\\

    & & \parbox{5cm}{\centering $\pixelPitch$: 3.74 $\mu m$, Z: 7 mm, VD: 4 mm, 1.4x}
      & 27.80 & 2.95 & 0.93 & 0.03 & 0.40 & 0.12 & 8.13 & &\\

    & & \parbox{5cm}{\centering $\pixelPitch$: 3.74 $\mu m$, Z: 7 mm, VD: 4 mm, 1.8x}
      & 27.52 & 2.92 & 0.92 & 0.04 & 0.41 & 0.13 & 8.11 & &\\

    & & \parbox{5cm}{\centering $\pixelPitch$: 3.74 $\mu m$, Z: 10 mm, VD: 4 mm, 1.0x}
      & 26.92 & 2.71 & 0.91 & 0.04 & 0.46 & 0.14 & 7.89 & &\\

    & & \parbox{5cm}{\centering $\pixelPitch$: 3.74 $\mu m$, Z: 10 mm, VD: 4 mm, 1.4x}
      & 26.97 & 2.66 & 0.91 & 0.04 & 0.47 & 0.14 & 7.93 & &\\

    & & \parbox{5cm}{\centering $\pixelPitch$: 3.74 $\mu m$, Z: 10 mm, VD: 4 mm, 1.8x}
      & 26.54 & 2.56 & 0.90 & 0.04 & 0.48 & 0.15 & 7.83 & &\\

    & & \parbox{5cm}{\centering $\pixelPitch$: 3.74 $\mu m$, Z: 2 mm, VD: 8 mm, 1.0x}
      & 28.23 & 2.87 & 0.93 & 0.05 & 0.39 & 0.12 & 8.00 & &\\

    & & \parbox{5cm}{\centering $\pixelPitch$: 3.74 $\mu m$, Z: 2 mm, VD: 8 mm, 1.4x}
      & 27.72 & 2.94 & 0.93 & 0.05 & 0.40 & 0.12 & 7.91 & &\\

    & & \parbox{5cm}{\centering $\pixelPitch$: 3.74 $\mu m$, Z: 2 mm, VD: 8 mm, 1.8x}
      & 26.89 & 2.91 & 0.90 & 0.06 & 0.40 & 0.12 & 7.85 & &\\

    & & \parbox{5cm}{\centering $\pixelPitch$: 3.74 $\mu m$, Z: 4 mm, VD: 8 mm, 1.0x}
      & 27.24 & 2.91 & 0.91 & 0.05 & 0.42 & 0.13 & 7.89 & &\\

    & & \parbox{5cm}{\centering $\pixelPitch$: 3.74 $\mu m$, Z: 4 mm, VD: 8 mm, 1.4x}
      & 27.21 & 2.94 & 0.91 & 0.06 & 0.42 & 0.12 & 7.88 & &\\

    & & \parbox{5cm}{\centering $\pixelPitch$: 3.74 $\mu m$, Z: 4 mm, VD: 8 mm, 1.8x}
      & 26.94 & 2.91 & 0.90 & 0.06 & 0.42 & 0.12 & 7.87 & &\\

    & & \parbox{5cm}{\centering $\pixelPitch$: 3.74 $\mu m$, Z: 7 mm, VD: 8 mm, 1.0x}
      & 27.00 & 3.01 & 0.90 & 0.05 & 0.44 & 0.14 & 7.89 & &\\

    & & \parbox{5cm}{\centering $\pixelPitch$: 3.74 $\mu m$, Z: 7 mm, VD: 8 mm, 1.4x}
      & 26.96 & 2.92 & 0.90 & 0.05 & 0.44 & 0.13 & 7.90 & &\\

    & & \parbox{5cm}{\centering $\pixelPitch$: 3.74 $\mu m$, Z: 7 mm, VD: 8 mm, 1.8x}
      & 26.73 & 2.88 & 0.89 & 0.06 & 0.44 & 0.13 & 7.89 & &\\

    & & \parbox{5cm}{\centering $\pixelPitch$: 3.74 $\mu m$, Z: 10 mm, VD: 8 mm, 1.0x}
      & 26.31 & 2.69 & 0.88 & 0.05 & 0.49 & 0.15 & 7.70 & &\\

    & & \parbox{5cm}{\centering $\pixelPitch$: 3.74 $\mu m$, Z: 10 mm, VD: 8 mm, 1.4x}
      & 26.33 & 2.65 & 0.88 & 0.05 & 0.49 & 0.15 & 7.72 & &\\

    & & \parbox{5cm}{\centering $\pixelPitch$: 3.74 $\mu m$, Z: 10 mm, VD: 8 mm, 1.8x}
      & 25.97 & 2.60 & 0.87 & 0.05 & 0.50 & 0.15 & 7.61 & &\\
    \hline
    \multirow{24}{*}{\parbox{2cm}{Our Method \\ (student)}}
    & \multirow{24}{*}{RGB-only}
      & \parbox{5cm}{\centering $\pixelPitch$: 3.74 $\mu m$, Z: 2 mm, VD: 4 mm, 1.0x}
      & 28.55 & 2.88 & 0.95 & 0.03 & 0.35 & \cellcolor{lightgreen}\textbf{0.10} & \cellcolor{lightgreen}\textbf{8.48} & \multirow{24}{*}{2.19 M} & \multirow{24}{*}{39 ms}\\

    & & \parbox{5cm}{\centering $\pixelPitch$: 3.74 $\mu m$, Z: 2 mm, VD: 4 mm, 1.4x}
      & 28.32 & 2.79 & 0.94 & 0.03 & 0.35 & 0.10 & 8.33 & &\\

    & & \parbox{5cm}{\centering $\pixelPitch$: 3.74 $\mu m$, Z: 2 mm, VD: 4 mm, 1.8x}
      & 27.69 & 2.78 & 0.94 & 0.04 & 0.36 & 0.11 & 8.18 & &\\

    & & \parbox{5cm}{\centering $\pixelPitch$: 3.74 $\mu m$, Z: 4 mm, VD: 4 mm, 1.0x}
      & 28.29 & 3.01 & 0.94 & 0.03 & 0.36 & 0.11 & 8.22 & &\\

    & & \parbox{5cm}{\centering $\pixelPitch$: 3.74 $\mu m$, Z: 4 mm, VD: 4 mm, 1.4x}
      & 28.21 & 2.98 & 0.94 & 0.03 & 0.37 & 0.11 & 8.13 & &\\

    & & \parbox{5cm}{\centering $\pixelPitch$: 3.74 $\mu m$, Z: 4 mm, VD: 4 mm, 1.8x}
      & 28.06 & 2.97 & 0.93 & 0.04 & 0.37 & 0.12 & 8.11 & &\\

    & & \parbox{5cm}{\centering $\pixelPitch$: 3.74 $\mu m$, Z: 7 mm, VD: 4 mm, 1.0x}
      & 28.28 & 3.15 & 0.93 & 0.03 & 0.39 & 0.12 & 8.15 & &\\

    & & \parbox{5cm}{\centering $\pixelPitch$: 3.74 $\mu m$, Z: 7 mm, VD: 4 mm, 1.4x}
      & 28.03 & 3.05 & 0.92 & 0.03 & 0.39 & 0.11 & 8.09 & &\\

    & & \parbox{5cm}{\centering $\pixelPitch$: 3.74 $\mu m$, Z: 7 mm, VD: 4 mm, 1.8x}
      & 27.93 & 2.85 & 0.92 & 0.04 & 0.40 & 0.12 & 8.03 & &\\

    & & \parbox{5cm}{\centering $\pixelPitch$: 3.74 $\mu m$, Z: 10 mm, VD: 4 mm, 1.0x}
      & 27.15 & 2.81 & 0.91 & 0.03 & 0.44 & 0.15 & 7.94 & &\\

    & & \parbox{5cm}{\centering $\pixelPitch$: 3.74 $\mu m$, Z: 10 mm, VD: 4 mm, 1.4x}
      & 27.07 & 2.72 & 0.91 & 0.04 & 0.44 & 0.15 & 7.90 & &\\

    & & \parbox{5cm}{\centering $\pixelPitch$: 3.74 $\mu m$, Z: 10 mm, VD: 4 mm, 1.8x}
      & 26.92 & 2.43 & 0.90 & 0.04 & 0.45 & 0.16 & 7.64 & &\\

    & & \parbox{5cm}{\centering $\pixelPitch$: 3.74 $\mu m$, Z: 2 mm, VD: 8 mm, 1.0x}
      & 27.89 & 2.98 & 0.93 & 0.04 & 0.39 & 0.12 & 8.02 & &\\

    & & \parbox{5cm}{\centering $\pixelPitch$: 3.74 $\mu m$, Z: 2 mm, VD: 8 mm, 1.4x}
      & 27.13 & 3.00 & 0.92 & 0.05 & 0.39 & 0.12 & 7.85 & &\\

    & & \parbox{5cm}{\centering $\pixelPitch$: 3.74 $\mu m$, Z: 2 mm, VD: 8 mm, 1.8x}
      & 26.76 & 2.86 & 0.91 & 0.06 & 0.40 & 0.13 & 7.71 & &\\

    & & \parbox{5cm}{\centering $\pixelPitch$: 3.74 $\mu m$, Z: 4 mm, VD: 8 mm, 1.0x}
      & 27.46 & 2.91 & 0.91 & 0.04 & 0.41 & 0.13 & 7.96 & &\\

    & & \parbox{5cm}{\centering $\pixelPitch$: 3.74 $\mu m$, Z: 4 mm, VD: 8 mm, 1.4x}
      & 27.34 & 2.88 & 0.90 & 0.05 & 0.41 & 0.13 & 7.84 & &\\

    & & \parbox{5cm}{\centering $\pixelPitch$: 3.74 $\mu m$, Z: 4 mm, VD: 8 mm, 1.8x}
      & 26.57 & 2.75 & 0.89 & 0.06 & 0.42 & 0.14 & 7.66 & &\\

    & & \parbox{5cm}{\centering $\pixelPitch$: 3.74 $\mu m$, Z: 7 mm, VD: 8 mm, 1.0x}
      & 27.24 & 3.07 & 0.90 & 0.05 & 0.43 & 0.13 & 7.91 & &\\

    & & \parbox{5cm}{\centering $\pixelPitch$: 3.74 $\mu m$, Z: 7 mm, VD: 8 mm, 1.4x}
      & 27.03 & 2.88 & 0.90 & 0.05 & 0.43 & 0.13 & 7.74 & &\\

    & & \parbox{5cm}{\centering $\pixelPitch$: 3.74 $\mu m$, Z: 7 mm, VD: 8 mm, 1.8x}
      & 26.32 & 2.80 & 0.88 & 0.06 & 0.45 & 0.14 & 7.58 & &\\

    & & \parbox{5cm}{\centering $\pixelPitch$: 3.74 $\mu m$, Z: 10 mm, VD: 8 mm, 1.0x}
      & 26.56 & 2.71 & 0.89 & 0.05 & 0.46 & 0.15 & 7.78 & &\\

    & & \parbox{5cm}{\centering $\pixelPitch$: 3.74 $\mu m$, Z: 10 mm, VD: 8 mm, 1.4x}
      & 26.23 & 2.76 & 0.88 & 0.05 & 0.46 & 0.16 & 7.64 & &\\

    & & \parbox{5cm}{\centering $\pixelPitch$: 3.74 $\mu m$, Z: 10 mm, VD: 8 mm, 1.8x}
      & 26.14 & 2.44 & 0.88 & 0.06 & 0.47 & 0.16 & 7.40 & &\\
    \hline
    \multirow{3}{*}{\parbox{2cm}{Our Method \\ (RGB-D condition)}}
    & \multirow{3}{*}{RGB-D}
      & \parbox{5cm}{\centering $\pixelPitch$: 6.4 $\mu m$, Z: \textit{4.88} mm, VD: \textit{6.75} mm, 1.0x\textsuperscript{1}}
      & 27.73 & 1.98 & 0.93 & 0.02 & 0.37 & 0.11 & 8.64 & \multirow{3}{*}{6.84 M} & \multirow{3}{*}{566 ms}\\

    & & \parbox{5cm}{\centering $\pixelPitch$: \textit{7.19} $\mu m$, Z: 2 mm, VD: \textit{4.23} mm, 1.0x\textsuperscript{1}}
      & 29.83 & 2.81 & 0.96 & 0.02 & 0.31 & 0.09 & 8.79 & &\\

    & & \parbox{5cm}{\centering $\pixelPitch$: \textit{4.57} $\mu m$, Z: 10 mm, VD: \textit{7.61} mm, 1.0x\textsuperscript{1}}
      & 26.93 & 2.19 & 0.91 & 0.03 & 0.42 & 0.13 & 8.07 & &\\
    \hline
    \multirow{4}{*}{HoloBeam}
    & \multirow{4}{*}{RGB-only\textsuperscript{2}}
      & \parbox{5cm}{\centering $\pixelPitch$: 3.74 $\mu m$, Z: 2 mm, VD: 4 mm, 1.0x}
      & 29.12 & 2.70 & 0.93 & 0.04 & 0.37 & \cellcolor{lightgreen}\textbf{0.10} & 8.37 & \multirow{4}{*}{1.94 M} & \multirow{4}{*}{\textbf{27 ms}}\\

    & & \parbox{5cm}{\centering $\pixelPitch$: 3.74 $\mu m$, Z: 4 mm, VD: 4 mm, 1.0x}
      & 27.15 & 2.62 & 0.90 & 0.03 & 0.43 & 0.11 & 8.24 & &\\

    & & \parbox{5cm}{\centering $\pixelPitch$: 3.74 $\mu m$, Z: 7 mm, VD: 4 mm, 1.0x}
      & 25.31 & 2.24 & 0.87 & 0.03 & 0.47 & 0.13 & 8.11 & &\\

    & & \parbox{5cm}{\centering $\pixelPitch$: 3.74 $\mu m$, Z: 10 mm, VD: 4 mm, 1.0x}
      & 20.62 & 2.51 & 0.82 & 0.03 & 0.49 & 0.14 & 7.99 & &\\
    \hline
    \multirow{4}{*}{\parbox{2cm}{Tensor\\V2}}
    & \multirow{4}{*}{RGB-D}
      & \parbox{5cm}{\centering $\pixelPitch$: 3.74 $\mu m$, Z: 2 mm, VD: 4 mm, 1.0x}
      & 29.23 & 2.26 & \cellcolor{lightgreen}\textbf{0.97} & 0.03 & \cellcolor{lightgreen}\textbf{0.34} & \cellcolor{lightgreen}\textbf{0.10} & 8.47 & \multirow{4}{*}{0.04 M} & \multirow{4}{*}{75 ms}\\

    & & \parbox{5cm}{\centering $\pixelPitch$: 3.74 $\mu m$, Z: 4 mm, VD: 4 mm, 1.0x}
      & 28.01 & 2.01 & 0.95 & 0.03 & 0.37 & 0.10 & 8.32 & &\\

    & & \parbox{5cm}{\centering $\pixelPitch$: 3.74 $\mu m$, Z: 7 mm, VD: 4 mm, 1.0x}
      & 25.89 & 2.09 & 0.93 & 0.03 & 0.39 & 0.11 & 8.29 & &\\

    & & \parbox{5cm}{\centering $\pixelPitch$: 3.74 $\mu m$, Z: 10 mm, VD: 4 mm, 1.0x}
      & 23.04 & 1.98 & 0.91 & 0.03 & 0.41 & 0.11 & 8.12 & &\\
    \hline
    \multirow{4}{*}{\parbox{2cm}{modified\\3D NH\textsuperscript{3}}}
    & \multirow{4}{*}{RGB-D}
      & \parbox{5cm}{\centering $\pixelPitch$: 3.74 $\mu m$, Z: 2 mm, VD: 4 mm, 1.0x}
      & 29.01 & 2.45 & 0.94 & 0.03 & 0.40 & \cellcolor{lightgreen}\textbf{0.10} & 8.41 & \multirow{4}{*}{3.87 M} & \multirow{4}{*}{49 ms}\\

    & & \parbox{5cm}{\centering $\pixelPitch$: 3.74 $\mu m$, Z: 4 mm, VD: 4 mm, 1.0x}
      & 29.03 & 2.31 & 0.92 & 0.04 & 0.41 & 0.11 & 8.25 & &\\

    & & \parbox{5cm}{\centering $\pixelPitch$: 3.74 $\mu m$, Z: 7 mm, VD: 4 mm, 1.0x\textsuperscript{4}}
      & 28.92 & 2.18 & 0.90 & 0.04 & 0.43 & 0.13 & 8.17 & &\\

    & & \parbox{5cm}{\centering $\pixelPitch$: 3.74 $\mu m$, Z: 10 mm, VD: 4 mm, 1.0x\textsuperscript{4}}
      & 28.98 & 2.39 & 0.89 & 0.04 & 0.45 & 0.13 & 8.05 & &\\

    \hline

    \multirow{4}{*}{\parbox{2cm}{Two-stage\\DepthAny V2}}
    & \multirow{4}{*}{RGB-D}
      & \parbox{4.9cm}{\centering $\pixelPitch$: 3.74 $\mu m$, Z: 2 mm, VD: 4 mm, 1.0x}
      & 29.21 & 3.15 & 0.95 & 0.03 & 0.35 & \cellcolor{lightgreen}\textbf{0.10} & 8.39 & \multirow{4}{*}{29.2 M} & \multirow{4}{*}{126 ms}\\

    & & \parbox{4.9cm}{\centering $\pixelPitch$: 3.74 $\mu m$, Z: 4 mm, VD: 4 mm, 1.0x}
      & 28.65 & 3.54 & 0.94 & 0.04 & 0.37 & 0.11 & 8.14 & &\\

    & & \parbox{4.9cm}{\centering $\pixelPitch$: 3.74 $\mu m$, Z: 7 mm, VD: 4 mm, 1.0x\textsuperscript{4}}
      & 27.92 & 3.28 & 0.92 & 0.04 & 0.40 & 0.12 & 8.07 & &\\

    & & \parbox{4.9cm}{\centering $\pixelPitch$: 3.74 $\mu m$, Z: 10 mm, VD: 4 mm, 1.0x\textsuperscript{4}}
      & 27.37 & 3.28 & 0.90 & 0.05 & 0.45 & 0.13 & 8.00 & &\\
    \hline
    \end{tabular}
  \caption{The extended evaluation table at different holography settings. All the other CGH models in the table, excluding ours, are separately trained with fixed configurations and have no configurability at all. \textit{Parameters} refers to the size of the model. \textit{Speed} refers to the inference time of the model under \fp32. Note that \textit{RGB teacher}, \textit{RGB student}, \textit{HoloBeam}, \textit{Tensor V2} and \textit{modified 3D NH} were trained with 800 $\times$ 800, 1024 $\times$ 1024, 1024 $\times$ 1024, 384 $\times$ 384 and 1024 $\times$ 1024 data, respectively. The test resolution is 1920 $\times$ 1080. \textsuperscript{1}Novel cases are in italic and \textit{not} included in metrics comparison. \textsuperscript{2}We improve the HoloBeam from G-only to RGB-only input. \textsuperscript{3}Since NH only supports 2D holograms, we modified NH`s HoloNet to generate 3D holograms. \textsuperscript{4}A notable disadvantage of the \textit{modified 3D NH} is its sensitivity to resolution under long propagation distances.}
  \label{supplmentary:table_holography_evaluation}
  \end{threeparttable}
\end{table*}

\subsection{Quantitative Analysis}
\label{supplementary:Extended_Evaluation_Analysis}
We compute PSNR, SSIM, LPIPS~\shortcite{zhang2019LPIPS}, FLIP~\shortcite{andersson2020flip}, and FVVDP~\shortcite{Mantiuk2021FVVDP} using 100 test images from DIV2K~\shortcite{Agustsson_2017_CVPR_Workshops_DIV2K} not used during training.
The student model maintains comparable performance to the teacher ($\triangle$PSNR$<0.1\%$), supporting the practical value of distillation.
Across configurations, PSNR and SSIM vary by $7.0\%$ and $6.3\%$ between best and worst settings, while remaining competitive with fixed-configuration baselines ($+4.9\%$ PSNR, $+2.9\%$ SSIM relative to per-configuration models).
Consistent with prior iterative pipelines~\cite{kavakli2023multicolor, kavakli2023realistic}, increasing $s$ or $Z$ reduces quality; for example, $s=1.8\times$ induces decreases of $1.9\%$ (PSNR), $1.1\%$ (SSIM), $2.6\%$ (LPIPS), $8.3\%$ (FLIP), and $3.6\%$ (FVVDP).

\section{Hardware Captured Results}
\label{supplementary:Hardware_Results}

\subsection{Captured Results}
\refFig{different_laser_focus_short} shows the hardware captured result of student model for short propagation distance.
\refFig{different_laser_focus_long} shows the hardware captured result of student model for long propagation distance.

\subsection{Comparisons under varying peak brightness}
\label{supplementary:PeakBrightness_SimCapture}
To facilitate direct comparison, this subsection aggregates the hardware-captured results from the previous two subsections into two large figures, allowing side-by-side evaluation with simulations.
\refFig{eval_short_prop_1_4} presents simulated and captured results of the student model at a short propagation distance ($Z = 2~mm$). Post-processing is performed using our in-house homography pipeline implemented in Python with OpenCV. Our method preserves fine details, color accuracy, and texture across different $s$ values.
\refFig{eval_long_prop_1_4} shows the results at a longer propagation distance ($Z = 10~mm$). Compared with the short-distance case, moderate color deviations appear in the hardware captures. We attribute this to the larger diffraction cone in long-distance propagation, which increases the required spatial bandwidth and amplifies the impact of limited training resolution, leading to chromatic artifacts. Training at higher resolution can alleviate this effect but significantly increases computational cost. To further investigate this sensitivity, we train an RGB-D conditioned model using full-HD data for $Z = 10~mm$ and compare it with modified 3D NH, Tensor V2, and Two-stage DepthAny V2.

\subsection{Hardware-captured Result On Other Holographic Displays}
\refFig{result_another_display} shows the hardware captured result of RGB-D condition on Holoeye Pluto-VIS holographic display
(resolution at 1080 $\times$ 1920 and pixel pitch at 8.0 $\mu m$).
\refFig{result_another_display_leto} shows the hardware captured result of RGB-D condition on Holoeye LETO-3 holographic display
(resolution at 1920 $\times$ 1080 and pixel pitch at 6.4 $\mu m$).
Both models are trained under~\refSupSec{RGBD_condition}'s condition.

\section{RGB-D Condition Model}
\label{supplementary:RGBD_condition}

\subsection{Model Architecture and Training Configuration}

To demonstrate that our model structure can condition various pixel pitches, wavelengths, and other display-scene parameters, we derive an RGB-D version of our RGB-only model.
\refFig{RGBD_condition_network} shows the RGB-D version of our model. Similar to the RGB-only model, this model also contains two consecutive \MLP layers for conditioning.

We choose the RGB-D approach for extensive parameter variation studies for two key reasons: First, RGB-D input directly provides depth information as prior knowledge,
allowing the model to focus on phase-only holography prediction, which greatly reduces data requirements compared to RGB-only models. Second, as the variable set grows larger,
the number of permutations increases exponentially, making RGB-D conditioning more computationally tractable.

\begin{figure*}[t!]
  \centering
  \includegraphics[width=1.0\textwidth]{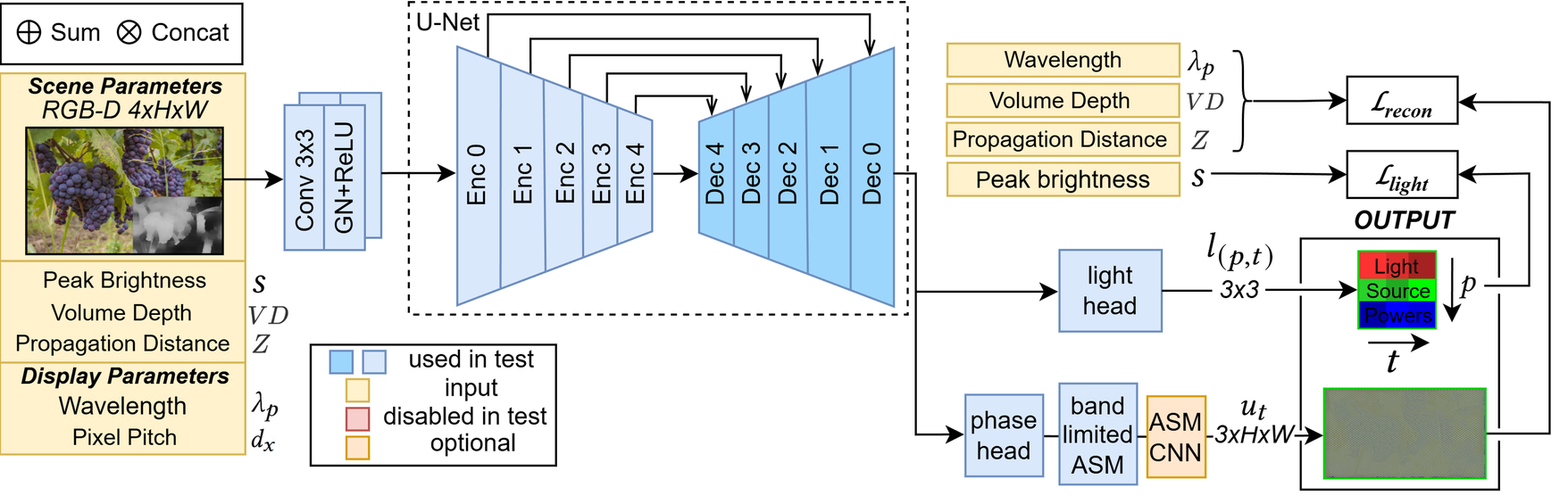}
  \caption{The overview of the RGB-D condition model. (RGB input: \cite{grapes2012})}
  \label{fig:RGBD_condition_network}
\end{figure*}

\subsection{Pair-wise \ANOVA}
\label{supplementary:Pairwise_ANOVA}

Pair-wise \ANOVA~\cite{ross2017anova} was employed to evaluate the impact of six different condition settings within the RGB-D condition model on image quality metrics.
Across PSNR, SSIM, LPIPS, and FLIP, conditions (1)--(4) (pixel pitches of $10.8~\mu m$ and $8.0~\mu m$) show no statistically significant differences (p-values ranging from 0.5 to 1.0), indicating that image quality is stable at larger pixel pitches.
In contrast, all comparisons involving conditions (5) and (6) (pixel pitch $3.74~\mu m$) yield p-values below $2\times10^{-16}$, confirming that smaller pixel pitch has a statistically significant impact on reconstruction quality.
\refTbl{anova_compare_various_model} presents the result of a pair-wise \ANOVA test conducted between our student model and other models.
For PSNR, comparisons with Holobeam and modified 3D NH yield p-values of 0.49 and 0.79 respectively, suggesting statistical equivalence in PSNR stability.
For Tensor V2, lower PSNR p-values (0.12) are observed, likely because Tensor V2 was evaluated using its short propagation weight (Z: 0mm, VD: 6mm) rather than a matched configuration, as reimplementing the MIT\_CGH\_4K dataset~\cite{shi2022end} for retraining was not feasible.
For SSIM and FLIP, p-values below $1\times10^{-16}$ against modified 3D NH and HoloBeam indicate that our student model achieves statistically better perceptual quality (student: SSIM $0.96\pm0.03$, FLIP $0.08$ vs.\ modified 3D NH: $0.94\pm0.03$, $0.11$ and HoloBeam: $0.93\pm0.04$, $0.10$).
Tensor V2 comparisons on SSIM (p=0.02) and FLIP (p=0.01) suggest the two methods are close but not statistically equivalent.
For LPIPS, the student model, HoloBeam (p=0.21), and Tensor V2 (p=0.09) are weakly related, with mean LPIPS of 0.35, 0.37, and 0.29 respectively.

\begin{table}[ht]
  \footnotesize
  \renewcommand{\arraystretch}{0.85}
  \setlength{\tabcolsep}{4pt}
  \centering
  \begin{tabular}{c c c c c}
  \hline
  & PSNR & SSIM & LPIPS & FLIP \\
  \hline
  student model \& modified 3D NH  & 0.79 & <2e-16 & 1e-15 & 1e-9 \\
  student model \& HoloBeam  & 0.49 & <2e-16 & 0.21 & <2e-16 \\
  student model \& Tensor V2  & 0.12 & 0.02 & 0.09 & 0.01 \\
  \hline
  \end{tabular}
  \caption{The p-value results of pair-wise \ANOVA test between student model and various models. The optic setting of the student model, modified 3D NH and HoloBeam is $\pixelPitch$: 3.74 $\mu m$, Z: 2 mm, VD: 4 mm, x1.0. The optic setting of Tensor V2 is $\pixelPitch$: 8.0 $\mu m$, Z: 0 mm, VD: 6 mm, x1.0.}
 \label{tbl:anova_compare_various_model}
\end{table}

\section{Additional Analyses}
\label{supplementary:additional_analyses}

\subsection{Comparison Between HoloBeam and Student Model}
\label{supplementary:teacher_student_compare}

\refFig{eval_holobeam_teacher_student} shows the reconstruction comparisons between HoloBeam and student model. When the input data is RGB-only, HoloBeam's results exhibited a deficiency in accurately representing the 3D phase, with incorrect defocus relationships. Conversely, our student model do not have such focusing issues.

\begin{figure*}[t!]
  \centering
  \includegraphics[width=0.99\textwidth]{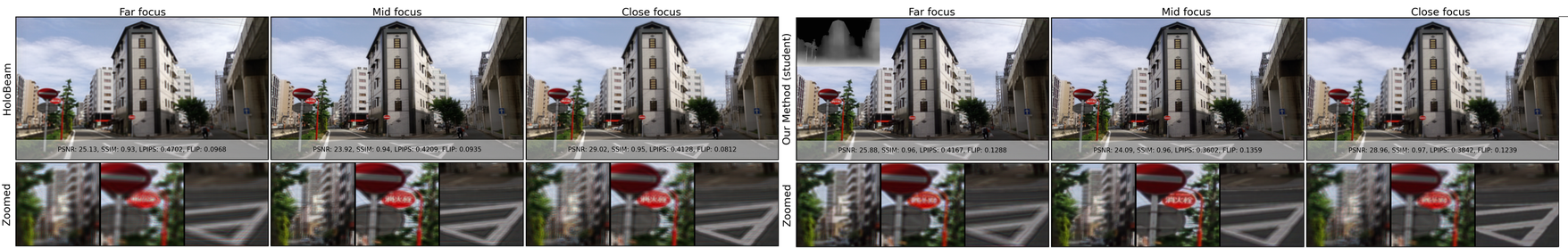}
  \caption{The simulated reconstructions comparison between HoloBeam and student model. The volume depth of the results is 4mm and the propagation distance is 2mm (Source Image: \cite{city2023}).}
\label{fig:eval_holobeam_teacher_student}
\end{figure*}

\subsection{Impact of Depth Estimation Inaccuracies}
\label{supplementary:MDEinaccuracy}

\refFig{depth_error_case} shows that depth estimation errors can cause unintended defocus in
regions meant to be sharp at specific depth planes, especially in high-resolution images.

\begin{figure*}[t!]
  \centering
  \includegraphics[width=0.99\textwidth]{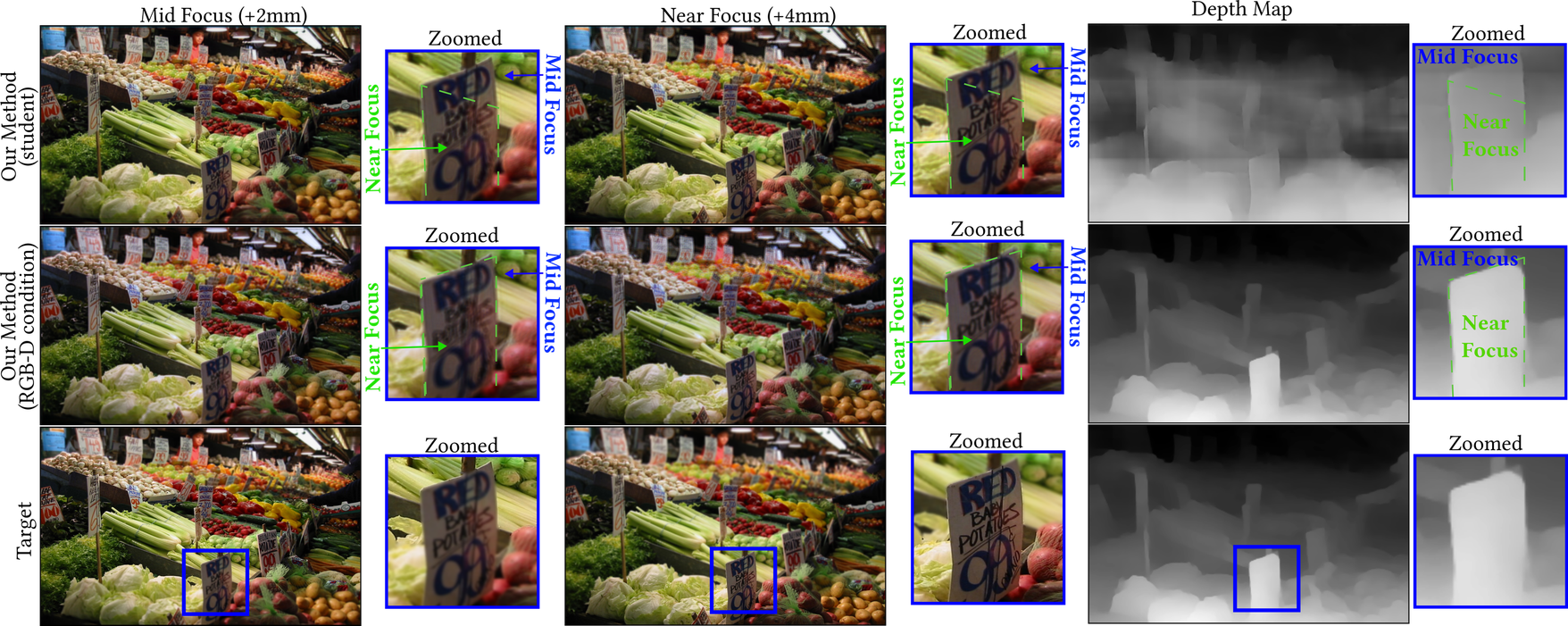}
\caption{Depth estimation errors causing incorrect defocus in high-resolution images. The student model predicts the top of the price board at the mid-focus plane, leading to erroneous reconstructions in the mid- and near-focus planes. Blue indicates the mid-focus region and green the near-focus region. (Source image: \cite{Fruitmarket2003}).}   \label{fig:depth_error_case}
\end{figure*}

\subsection{Ablation Study: Parameter Embedding}
\label{supplementary:PSF_ablation}

To evaluate the contribution of each component in our parameter embedding layer ,
we conduct an inference-time ablation study using a trained RGB-D condition model checkpoint.
We compare three variants: (a)~the full embedding combining sinusoidal scalar encoding and 1D PSF,
(b)~sinusoidal scalars only,
and (c)~1D PSF only.
We evaluate on 100 DIV2K test images
across four representative configurations spanning short/long propagation distances and two pixel pitches.
Results are reported in Tbl.~S\ref{tbl:PSF_ablation}.
\begin{table}[!htbp]
  \centering
  \footnotesize
  \setlength{\tabcolsep}{3pt}
  \renewcommand{\arraystretch}{1.0}
  \begin{tabular}{l c c c c}
    \toprule
    \textbf{Embedding Variant}
    & \textbf{PSNR$\uparrow$}
    & \textbf{SSIM$\uparrow$}
    & \textbf{LPIPS$\downarrow$}
    & \textbf{FLIP$\downarrow$} \\
    \midrule
    \multicolumn{5}{l}{\textit{$d_x$=3.74\,$\mu$m, Z=2\,mm, VD=4\,mm, $\times$1.0}} \\
    (a) Full (sinusoidal + 1D PSF) & \textbf{28.85} & \textbf{0.96} & \textbf{0.36} & \textbf{0.13} \\
    (b) Sinusoidal scalars only    & 27.52 & 0.94 & 0.41 & 0.15 \\
    (c) 1D PSF only                & 25.13 & 0.91 & 0.48 & 0.19 \\
    \midrule
    \multicolumn{5}{l}{\textit{$d_x$=3.74\,$\mu$m, Z=10\,mm, VD=4\,mm, $\times$1.0}} \\
    (a) Full (sinusoidal + 1D PSF) & \textbf{26.41} & \textbf{0.94} & \textbf{0.40} & \textbf{0.15} \\
    (b) Sinusoidal scalars only    & 25.18 & 0.92 & 0.45 & 0.18 \\
    (c) 1D PSF only                & 23.06 & 0.88 & 0.53 & 0.22 \\
    \midrule
    \multicolumn{5}{l}{\textit{$d_x$=8.0\,$\mu$m, Z=2\,mm, VD=6\,mm, $\times$1.0}} \\
    (a) Full (sinusoidal + 1D PSF) & \textbf{31.22} & \textbf{0.97} & \textbf{0.30} & \textbf{0.07} \\
    (b) Sinusoidal scalars only    & 30.05 & 0.96 & 0.34 & 0.09 \\
    (c) 1D PSF only                & 27.84 & 0.93 & 0.42 & 0.12 \\
    \midrule
    \multicolumn{5}{l}{\textit{$d_x$=8.0\,$\mu$m, Z=10\,mm, VD=6\,mm, $\times$1.0}} \\
    (a) Full (sinusoidal + 1D PSF) & \textbf{28.76} & \textbf{0.95} & \textbf{0.35} & \textbf{0.10} \\
    (b) Sinusoidal scalars only    & 27.61 & 0.93 & 0.39 & 0.12 \\
    (c) 1D PSF only                & 25.49 & 0.90 & 0.47 & 0.16 \\
    \bottomrule
  \end{tabular}
  \caption{Ablation of the parameter embedding layer. }
  \label{tbl:PSF_ablation}
\end{table}

The full embedding consistently achieves the best performance across all four configurations.
Removing the 1D PSF branch (variant~b) causes an average PSNR drop of 1.3\,dB.
Removing the sinusoidal scalars (variant~c) leads to a larger average drop of 3.5\,dB, as the 1D PSF alone cannot disambiguate configurations that share similar PSF shapes but differ in wavelength or volume depth.
These results confirm that both branches provide complementary information. The sinusoidal scalars supply exact parameter values while the 1D PSF injects physics-aware spatial structure.

\section{Varying Display-Scene Parameters}
\label{supplementary:ParametersRange}
\subsection{Generalizing Novel Cases Outside Of Training (Pixel Pitch)}

Our model can generalize novel pixel pitch values when the distance between the training conditions is small. Empirically, we found that the model has the best performance when the step size between cases is around 0.05 $\mu m$. We trained the RGB-D condition model with the following conditions:

\begin{equation}
  \begin{split}
    \wavelength_{\pIndex} \subseteq \{(640,515,470)\} ~nm, \\
    \scale \subseteq \{1.0 \}, VD \subseteq \{4.0, 8.0\} ~mm,\\
    Z \subseteq \{2.0, 10.0 \} ~mm, \pixelPitch \subseteq \{3.7 - 8.0\} ~\mu m.
    \label{eq:variable_set8}
    \end{split}
\end{equation}

In this experiment, we include the pixel pitch covering a span of 4.3 $\mu m$ range, 2 propagation distance, and a volume depth as large as 4mm. The pixel pitch conditions have a 0.05 $\mu m$ step size, which results in 54 conditions in total. The entire permutation set contains 344 conditions in total and the model is trained at the resolution of 896 $\times$ 896.

To conduct a comprehensive evaluation of the model, we employ an uniform sampling approach across the entire range of pixel pitch. For each pixel pitch interval (e.g. 4-5~$\mu m$), we generate a set of 20 novel pixel pitches outside of the training set. These randomly selected pixel pitches are uniformly distributed within the interval, cases are approximately 0.05~$\mu m$ between each other. We use the following novel cases as test cases:

\begin{equation}
  \begin{split}
    \wavelength_{\pIndex} \subseteq \{(640,515,470)\} ~nm, \\
    \scale \subseteq \{1.0 \}, VD \subseteq \{\text{randomized between 4.0 - 8.0} \} ~mm,\\
    Z \subseteq \{ 2.0, 10.0\} ~mm, \\
    \pixelPitch \subseteq \{ \text{randomized pixel pitch distribution} \} ~\mu m.
    \label{eq:variable_set9}
    \end{split}
\end{equation}

The test permutation set contains 344 novel cases at a resolution of $1280 \times 1280$. Due to the large number of conditions, tabular reporting is
impractical; instead, we present the results using figures. \refFig{DiffDx_continue} reports PSNR, SSIM, and FLIP
distributions, where each data point is computed from the same 100 images from the DIV2K dataset
~\cite{Agustsson_2017_CVPR_Workshops_DIV2K}. Overall, the model maintains stable image quality with low standard
deviation across randomly sampled pixel pitch conditions over a continuous range. About 10\% of cases show
noticeable quality variations, likely due to the limited model capacity when generalizing $\pixelPitch$ across a
wide continuous range (4.3~$\mu m$), which is more challenging than other display–scene parameters and makes adapting to hundreds of conditions difficult.

\begin{figure}[t!]
  \centering
  \includegraphics[width=0.99\columnwidth]{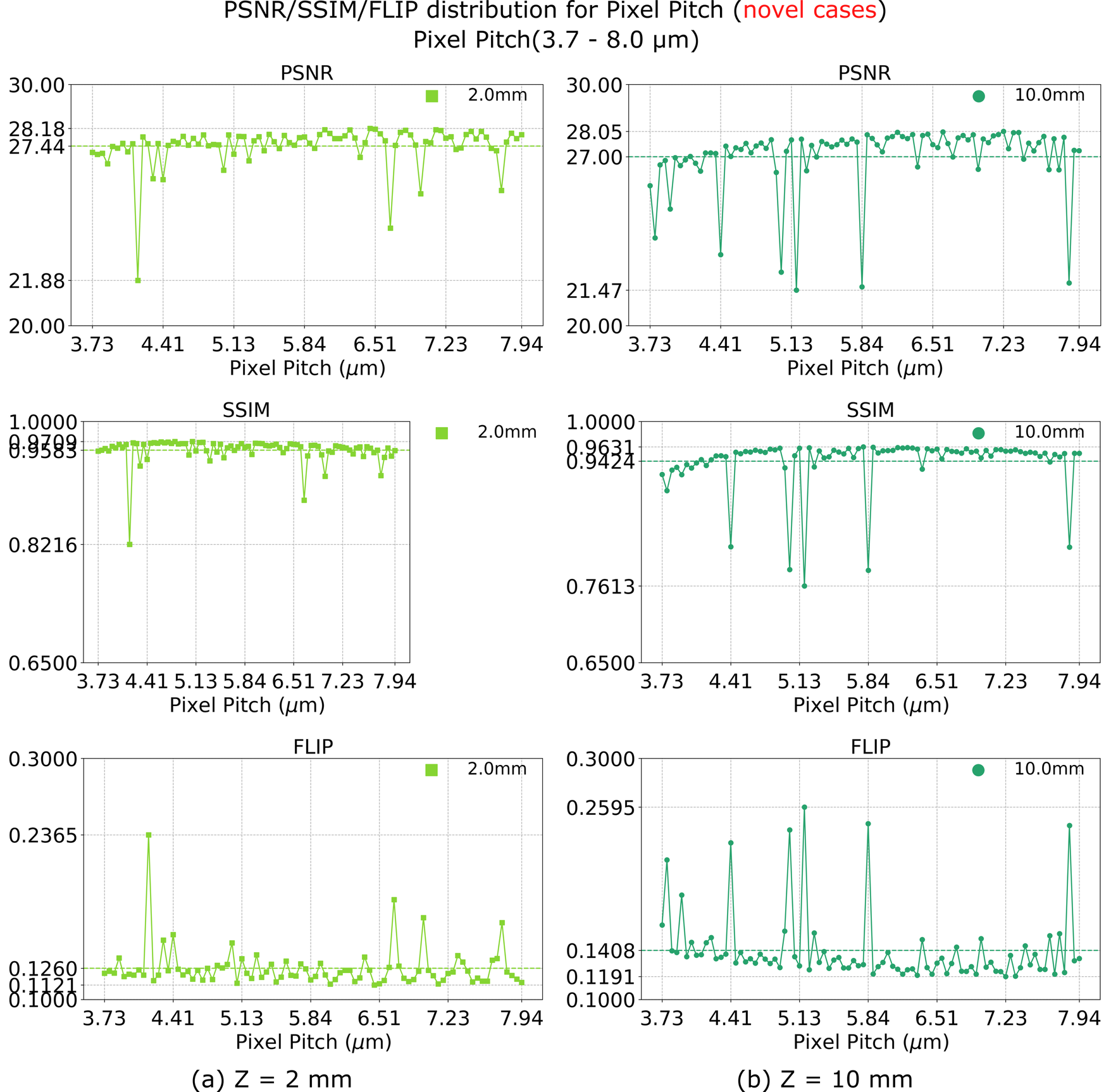}
  \caption{The PSNR, SSIM, and FLIP distribution of our RGB-D condition model when conditioned on a 4.3 $\mu m$ total pixel pitch range across two propagation distances.}
  \label{fig:DiffDx_continue}
  \vspace{-5mm}
\end{figure}

\subsection{Generalizing Novel Cases Outside Of Training (Propagation Distance)}
Similar to pixel pitch, our model can generalize novel propagation distance values when the distance between the training conditions is small.
Empirically, we found that the model has the best performance when the step size between cases is around 0.005 mm. We trained the RGB-D condition model with the following conditions:

\begin{equation}
  \begin{split}
    \wavelength_{\pIndex} \subseteq \{(640,515,470)\} ~nm, \\
    \scale \subseteq \{1.0 \}, VD \subseteq \{4.0, 8.0\} ~mm,\\
    Z \subseteq \{2.0 - 3.0; 4.0 - 5.0; 6.0 - 7.0; \\
    8.0 - 9.0; 10.0 - 11.0 \} ~mm, \\
    \pixelPitch \subseteq \{3.74, 6.4, 8.0\} ~\mu m.
    \label{eq:variable_set11}
    \end{split}
\end{equation}

In this experiment, we include the propagation distance covering a span of 5mm range, 3 common pixel pitches, and a volume depth as large as 4mm. The training propagation distance conditions are 0.005 mm between each others, which results in 1000 conditions in total. The entire permutation set contains 6000 conditions in total and the model is trained at the resolution of 512 $\times$ 512. We choose a discrete, rather than a continuous, 5mm range because we want to maximize the Z range coverage while avoiding excessive computational demands. The conditioning of a continuous 5mm range of Z will also work under our training setting.
To conduct a comprehensive evaluation of the model, we employ an uniform sampling approach across the entire range of propagation distances. For each propagation distance interval (e.g. 2 - 3 mm), we generate a set of 100 novel propagation distances outside of the training set. These randomly selected distances are uniformly distributed within the interval, cases are approximately 0.01mm between each other. We use the following novel cases as test cases:

\begin{equation}
  \begin{split}
    \wavelength_{\pIndex} \subseteq \{(640,515,470)\} ~nm, \\
    \scale \subseteq \{1.0 \}, VD \subseteq \{\text{randomized between 4.0 - 8.0} \} ~mm,\\
    Z \subseteq \{ \text{randomized distance distribution}\} ~mm, \\
    \pixelPitch \subseteq \{3.74, 6.4, 8.0\} ~\mu m.
    \label{eq:variable_set12}
    \end{split}
\end{equation}

The test permutation set contains 3000 novel cases at a resolution of $1024 \times 1024$. Due to the large number of conditions, tabular evaluation is impractical; therefore, results are presented in figure form. \refFig{PSNR5mm_continue_Z} shows the PSNR distribution, while SSIM and FLIP follow the same trend and are omitted for brevity. Each data point is computed from the same 100 images from the DIV2K dataset~\cite{Agustsson_2017_CVPR_Workshops_DIV2K}.
Overall, the model maintains stable image quality with low standard deviation across randomly generated propagation distances over a continuous range. About 20\% of cases show noticeable quality variations, likely due to the limited model capacity when generalizing continuous $Z$ over a wide range (5~mm), which is more challenging than other display–scene parameters and makes adapting to thousands of conditions difficult.

\subsection{Generalizing Novel Cases Outside Of Training (Wavelength)}
As demonstrated in~\refSupSec{RGBD_condition}, our model structure effectively adapts to a wide range of wavelengths.
In this section, we focus on showcasing the model's capability for continuous wavelength generalization under the RGB-D condition.
Our model can generalize novel wavelength values continously. Empirically, we found that the model has the best performance when the step size between cases is around 5~$nm$. We trained the RGB-D condition model with the following conditions:

\begin{equation}
  \begin{split}
    \wavelength_{\pIndex} \subseteq \{(625 - 680, 510 - 565, 425 - 480)\} ~nm, \\
    \scale \subseteq \{1.0 \}, VD \subseteq \{4.0, 8.0\} ~mm,\\
    Z \subseteq \{2.0, 10.0\} ~mm, \pixelPitch \subseteq \{3.74, 8.0\} ~\mu m.
    \label{eq:variable_set13}
    \end{split}
\end{equation}

In this experiment, we consider wavelengths spanning a 55~$nm$ range, two propagation distances, two common pixel pitches, and a volume depth up to 4~mm. Wavelength conditions are sampled at 5~$nm$ intervals, and the model is trained at a resolution of $640 \times 640$. Given the large number of possible wavelength combinations (a 55~$nm$ range yields $55^3 = 166375$ combinations), we evaluate the model on twenty randomly sampled novel cases at a test resolution of $1024 \times 1024$:

  Condition Set (1) \begin{equation}
    \begin{split}
      \wavelength_{\pIndex} \subseteq \{(641, 546, 478), (668, 547, 461), \\
      (676, 563, 441), (632, 538, 473), (659, 519, 432)\} ~nm, \\
      \scale \subseteq \{1.0 \}, VD \subseteq \{\text{randomized between 4.0 - 8.0}\} ~mm,\\
      Z \subseteq \{2.0\} ~mm, \pixelPitch \subseteq \{3.74\} ~\mu m.
      \label{eq:variable_set14}
      \end{split}
  \end{equation}

  Condition Set (2) \begin{equation}
    \begin{split}
      \wavelength_{\pIndex} \subseteq \{(661, 558, 433), (628, 554, 480), \\
      (631, 525, 474), (653, 562, 471), (679, 521, 462)\} ~nm, \\
      \scale \subseteq \{1.0 \}, VD \subseteq \{\text{randomized between 4.0 - 8.0}\} ~mm,\\
      Z \subseteq \{10.0\} ~mm, \pixelPitch \subseteq \{3.74\} ~\mu m.
      \label{eq:variable_set15}
      \end{split}
  \end{equation}

  Condition Set (3) \begin{equation}
    \begin{split}
      \wavelength_{\pIndex} \subseteq \{(668, 518, 426), (628, 510, 444), \\
       (678, 557, 467), (639, 541, 437), (655, 531, 434)\} ~nm, \\
      \scale \subseteq \{1.0 \}, VD \subseteq \{\text{randomized between 4.0 - 8.0}\} ~mm,\\
      Z \subseteq \{2.0\} ~mm, \pixelPitch \subseteq \{8.0\} ~\mu m.
      \label{eq:variable_set16}
      \end{split}
  \end{equation}

  Condition Set (4) \begin{equation}
    \begin{split}
      \wavelength_{\pIndex} \subseteq \{(659, 523, 454), (663, 562, 466), \\
      (677, 515, 443), (628, 542, 429), (642, 557, 457)\} ~nm, \\
      \scale \subseteq \{1.0 \}, VD \subseteq \{\text{randomized between 4.0 - 8.0}\} ~mm,\\
      Z \subseteq \{10.0\} ~mm, \pixelPitch \subseteq \{8.0\} ~\mu m.
      \label{eq:variable_set17}
      \end{split}
  \end{equation}

\refTbl{continueWave} shows our model's performance across twenty test conditions with different randomized RGB wavelengths under various combinations of propagation distance and pixel pitch. The model can take arbitrary wavelengths as the input and maintain consistent image quality over different propagation distances and pixel pitches. Each data in the table is contributed by the same 100 images from the DIV2K dataset~\cite{Agustsson_2017_CVPR_Workshops_DIV2K}.

\begin{table*}[!htbp]
  \centering
  \begin{threeparttable}
    \footnotesize
    \renewcommand{\arraystretch}{0.85}
    \setlength{\tabcolsep}{4pt}
    \begin{tabular}{
      m{1.9cm}
      m{0.8cm}
      l
      c  c
      c  c
      m{0.8cm}
      m{0.8cm}
      m{1.0cm}
    }
    \hline
    \textbf{Method}
    & \textbf{Input}
    & \parbox{6cm}{\centering \textbf{Display-scene Parameters}}
    & \multicolumn{2}{c}{\textbf{PSNR↑ (dB)}}
    & \multicolumn{2}{c}{\textbf{SSIM↑}}
    & \textbf{LPIPS↓}
    & \textbf{FLIP↓}
    & \textbf{FVVDP↑} \\
    \addlinespace[0.15em]
    \cline{4-10}
    & & & Mean & Std & Mean & Std & Mean & Mean & Mean \\
    \hline

    \multirow{20}{*}{\parbox{2cm}{Our Method \\ (RGB-D condition)}}
    & \multirow{20}{*}{RGB-D}
      & \parbox{6.3cm}{\centering $\lambda$: (\novel{641},\novel{546},\novel{478}) nm, Z: 10 mm, VD: \novel{4.2} mm, $\pixelPitch$: 3.74 $\mu m$, x1.0}
      & 25.10 & 2.52 & 0.90 & 0.03 & 0.43 & 0.19 & 7.41 \\
    & &
      \parbox{6.3cm}{\centering $\lambda$: (\novel{668}, \novel{547}, \novel{461}) nm, Z: 10 mm, VD: \novel{5.3} mm, $\pixelPitch$: 3.74 $\mu m$, x1.0}
      & 24.90 & 2.51 & 0.89 & 0.03 & 0.44 & 0.19 & 7.36 \\
    & &
      \parbox{6.3cm}{\centering $\lambda$: (675, \novel{563}, \novel{441}) nm, Z: 10 mm, VD: \novel{7.1} mm, $\pixelPitch$: 3.74 $\mu m$, x1.0}
      & 24.58 & 2.47 & 0.87 & 0.04 & 0.46 & 0.20 & 7.24 \\
    & &
      \parbox{6.3cm}{\centering $\lambda$: (\novel{632}, \novel{538}, \novel{473}) nm, Z: 10 mm, VD: \novel{5.6} mm, $\pixelPitch$: 3.74 $\mu m$, x1.0}
      & 24.94 & 2.50 & 0.89 & 0.03 & 0.44 & 0.19 & 7.37 \\
    & &
      \parbox{6.3cm}{\centering $\lambda$: (\novel{659}, \novel{519}, \novel{432}) nm, Z: 10 mm, VD: \novel{6.4} mm, $\pixelPitch$: 3.74 $\mu m$, x1.0}
      & 24.76 & 2.51 & 0.88 & 0.04 & 0.45 & 0.19 & 7.35 \\
    & &
      \parbox{6.3cm}{\centering $\lambda$: (\novel{661}, \novel{558}, \novel{433}) nm, Z: 2 mm, VD: \novel{6.9} mm, $\pixelPitch$: 3.74 $\mu m$, x1.0}
      & 26.11 & 2.93 & 0.92 & 0.04 & 0.37 & 0.14 & 7.87 \\
    & &
      \parbox{6.3cm}{\centering $\lambda$: (\novel{628}, \novel{554}, 480) nm, Z: 2 mm, VD: \novel{4.5} mm, $\pixelPitch$: 3.74 $\mu m$, x1.0}
      & 27.06 & 2.99 & 0.94 & 0.03 & 0.34 & 0.12 & 8.17 \\
    & &
      \parbox{6.3cm}{\centering $\lambda$: (\novel{631}, \novel{521}, 475) nm, Z: 2 mm, VD: \novel{5.8} mm, $\pixelPitch$: 3.74 $\mu m$, x1.0}
      & 26.53 & 2.94 & 0.93 & 0.03 & 0.35 & 0.13 & 8.04 \\
    & &
      \parbox{6.3cm}{\centering $\lambda$: (\novel{653}, \novel{562}, \novel{471}) nm, Z: 2 mm, VD: \novel{6.1} mm, $\pixelPitch$: 3.74 $\mu m$, x1.0}
      & 26.35 & 2.93 & 0.93 & 0.03 & 0.36 & 0.13 & 7.96 \\
    & &
      \parbox{6.3cm}{\centering $\lambda$: (\novel{679}, \novel{521}, \novel{462}) nm, Z: 2 mm, VD: \novel{7.8} mm, $\pixelPitch$: 3.74 $\mu m$, x1.0}
      & 25.76 & 2.94 & 0.91 & 0.05 & 0.39 & 0.14 & 7.80 \\

    & &
      \parbox{6.3cm}{\centering $\lambda$: (\novel{668}, \novel{518}, \novel{426}) nm, Z: 10 mm, VD: \novel{4.1} mm, $\pixelPitch$: 8.0 $\mu m$, x1.0}
      & 28.15 & 3.12 & 0.95 & 0.03 & 0.30 & 0.12 & 8.19 \\
    & &
      \parbox{6.3cm}{\centering $\lambda$: (\novel{628}, 510, \novel{444}) nm, Z: 10 mm, VD: \novel{5.7} mm, $\pixelPitch$: 8.0 $\mu m$, x1.0}
      & 28.27 & 3.14 & 0.95 & 0.02 & 0.29 & 0.12 & 8.23 \\
    & &
      \parbox{6.3cm}{\centering $\lambda$: (\novel{678}, \novel{557}, \novel{467}) nm, Z: 10 mm, VD: \novel{4.7} mm, $\pixelPitch$: 8.0 $\mu m$, x1.0}
      & 28.22 & 3.14 & 0.95 & 0.02 & 0.30 & 0.12 & 8.21 \\
    & &
      \parbox{6.3cm}{\centering $\lambda$: (\novel{639}, \novel{541}, \novel{437}) nm, Z: 10 mm, VD: \novel{6.5} mm, $\pixelPitch$: 8.0 $\mu m$, x1.0}
      & 28.32 & 3.16 & 0.95 & 0.02 & 0.30 & 0.12 & 8.25 \\
    & &
      \parbox{6.3cm}{\centering $\lambda$: (655, \novel{531}, \novel{434}) nm, Z: 10 mm, VD: \novel{7.3} mm, $\pixelPitch$: 8.0 $\mu m$, x1.0}
      & 28.35 & 3.18 & 0.95 & 0.02 & 0.30 & 0.12 & 8.27 \\
    & &
      \parbox{6.3cm}{\centering $\lambda$: (\novel{659}, \novel{523}, \novel{454}) nm, Z: 2 mm, VD: \novel{4.9} mm, $\pixelPitch$: 8.0 $\mu m$, x1.0}
      & 28.45 & 3.11 & 0.96 & 0.02 & 0.26 & 0.11 & 8.40 \\
    & &
      \parbox{6.3cm}{\centering $\lambda$: (\novel{663}, \novel{562}, \novel{466}) nm, Z: 2 mm, VD: \novel{5.6} mm, $\pixelPitch$: 8.0 $\mu m$, x1.0}
      & 28.40 & 3.13 & 0.96 & 0.02 & 0.26 & 0.11 & 8.41 \\
    & &
      \parbox{6.3cm}{\centering $\lambda$: (\novel{677}, 515, \novel{443}) nm, Z: 2 mm, VD: \novel{6.9} mm, $\pixelPitch$: 8.0 $\mu m$, x1.0}
      & 28.29 & 3.18 & 0.96 & 0.02 & 0.27 & 0.11 & 8.40 \\
    & &
      \parbox{6.3cm}{\centering $\lambda$: (\novel{628}, \novel{542}, \novel{429}) nm, Z: 2 mm, VD: \novel{7.1} mm, $\pixelPitch$: 8.0 $\mu m$, x1.0}
      & 28.29 & 3.19 & 0.96 & 0.02 & 0.27 & 0.11 & 8.40 \\
    & &
      \parbox{6.3cm}{\centering $\lambda$: (\novel{642}, \novel{557}, \novel{457}) nm, Z: 2 mm, VD: \novel{4.6} mm, $\pixelPitch$: 8.0 $\mu m$, x1.0}
      & 28.47 & 3.11 & 0.96 & 0.02 & 0.26 & 0.11 & 8.40 \\
    \hline
    \end{tabular}
  \caption{Evaluation of RGB-D condition model with random wavelengths and VD values at different pixel pitches, and propagation distances settings. The test resolution is at 1024 $\times$ 1024. The novel cases` (outside of training set) metrics are marked in red.}
  \label{tbl:continueWave}
  \label{supplementary:continueWave}
  \end{threeparttable}
\end{table*}

\subsection{Effect of Conditioning Range on Quality Stability}
\label{supplementary:narrower_Z_range}

In the main paper and \refSupSec{ParametersRange}, we observe quality fluctuations when the model conditions on
propagation distances across a wide 5\,mm total range (five disjoint 1\,mm intervals).
To investigate whether reducing the total conditioning range mitigates these fluctuations,
we train an RGB-D condition model on a single continuous 1\,mm interval from 6--7\,mm
with all other parameters fixed:
\begin{equation}
  \begin{split}
    \wavelength_{\pIndex} \subseteq \{(644,519,468)\} ~nm, \\
    \scale \subseteq \{1.0 \}, VD \subseteq \{4.0, 8.0\} ~mm,\\
    Z \subseteq \{6.0 - 7.0\} ~mm, \pixelPitch \subseteq \{3.74\} ~\mu m.
    \label{eq:variable_set_narrow_Z}
  \end{split}
\end{equation}
The training propagation distance conditions are sampled at 0.005\,mm intervals within the 6--7\,mm range,
yielding 200 $Z$ conditions. Combined with two $VD$ values, the total training permutation set contains 400 conditions.
The model is trained at a resolution of 512 $\times$ 512.
For evaluation, we generate 100 novel propagation distances randomly sampled within 6--7\,mm (approximately 0.01\,mm apart),
each paired with a randomly sampled $VD$ in $[4.0, 8.0]$\,mm.
The test resolution is 1024 $\times$ 1024, and each data point is the mean PSNR over the same 100 DIV2K images~\cite{Agustsson_2017_CVPR_Workshops_DIV2K}.
\refFig{narrow_vs_wide_Z} compares the PSNR distribution of the constrained-range model (1\,mm, 6--7\,mm)
against the corresponding 6--7\,mm interval from the wide-range model (5\,mm total, five disjoint 1\,mm intervals).
\begin{figure}[t!]
  \centering
  \includegraphics[width=0.99\columnwidth]{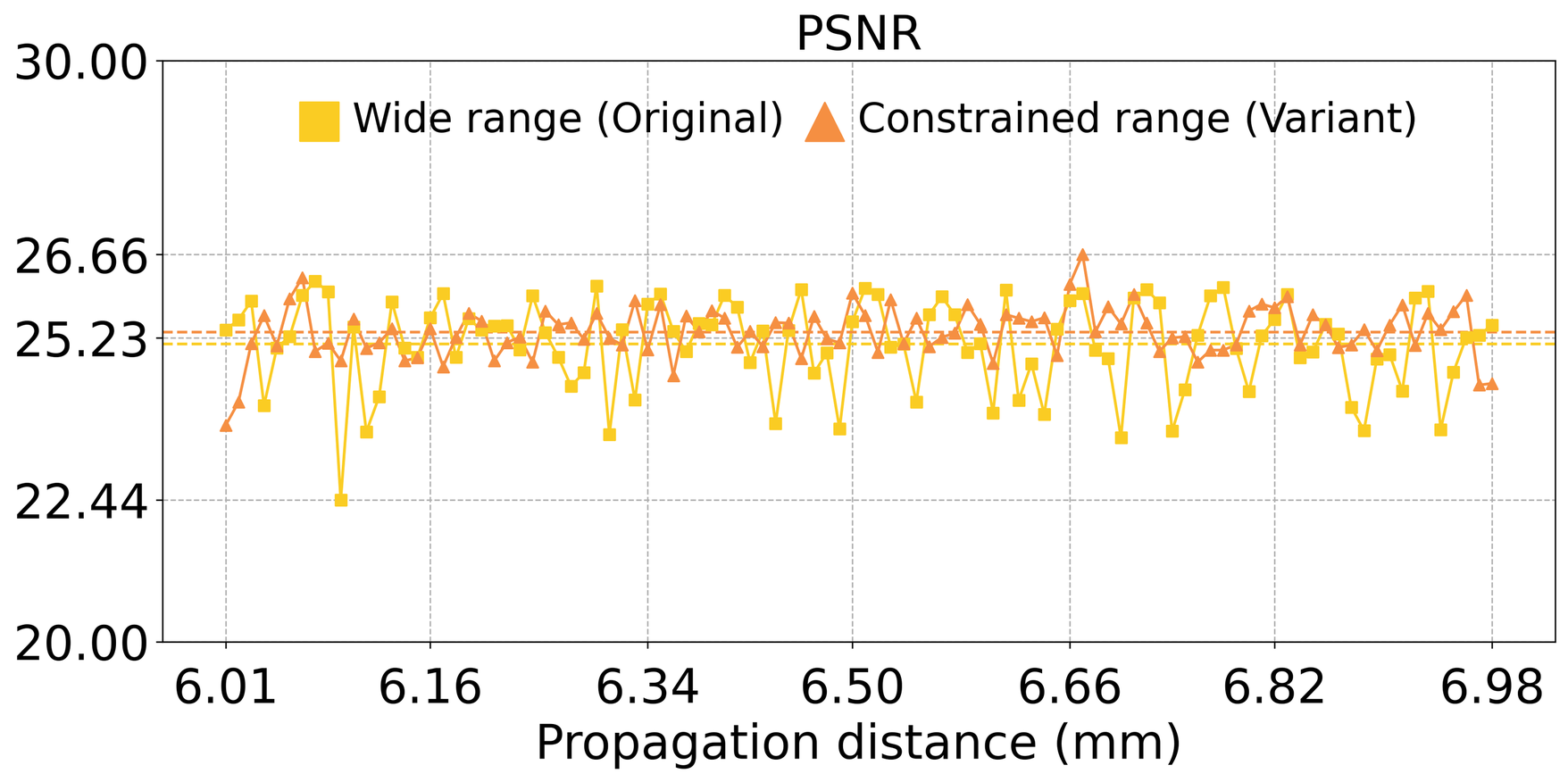}
  \caption{PSNR distribution over 100 novel propagation distances at $\pixelPitch=3.74\,\mu$m.
  Each point is the mean PSNR over 100 DIV2K test images at a randomly sampled $(VD)$.}
  \label{fig:narrow_vs_wide_Z}
\end{figure}
The constrained-range model achieves a mean PSNR of 25.33\,dB with a standard deviation of 0.44\,dB,
compared to 25.13\,dB mean and 0.78\,dB standard deviation for the same $Z$ interval extracted from the
wide-range model.
Moreover, the minimum per-condition PSNR rises from 22.44\,dB to 23.72\,dB,
indicating that reducing the total conditioning range can suppress the
quality drops observed in the wide-range setting.

\begin{figure*}[t!]
  \centering
  \includegraphics[width=1.0\textwidth]{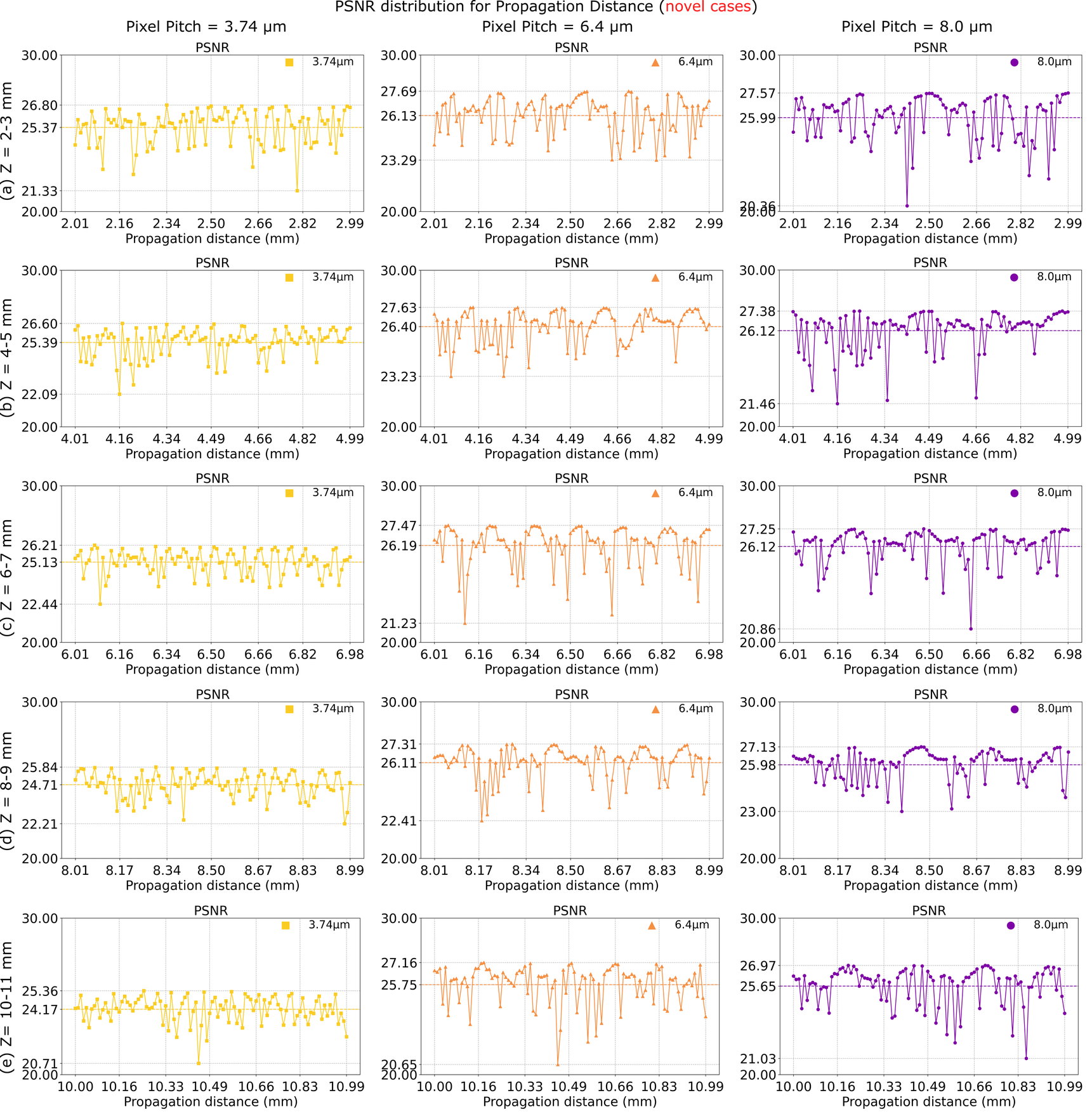}
  \caption{PSNR distribution of our RGB-D condition model across novel propagation distances for three pixel pitches ($3.74$, $6.4$, $8.0~\mu m$). Each panel covers a continuous 1~mm $Z$ interval; each data point is the mean PSNR over 100 DIV2K test images at a randomly sampled novel $(Z, VD)$ pair. The model maintains consistent quality (average PSNR $\approx 26$~dB, average std $\approx 1.1$~dB) with isolated drops in roughly 20\% of cases, which we attribute to limited model capacity over the wide conditioning range. SSIM and FLIP distributions follow the same trend and are omitted for brevity.}
  \label{fig:PSNR5mm_continue_Z}
  \vspace{-5mm}
\end{figure*}

\begin{figure*}[t!]
  \centering
  \includegraphics[width=1.0\textwidth]{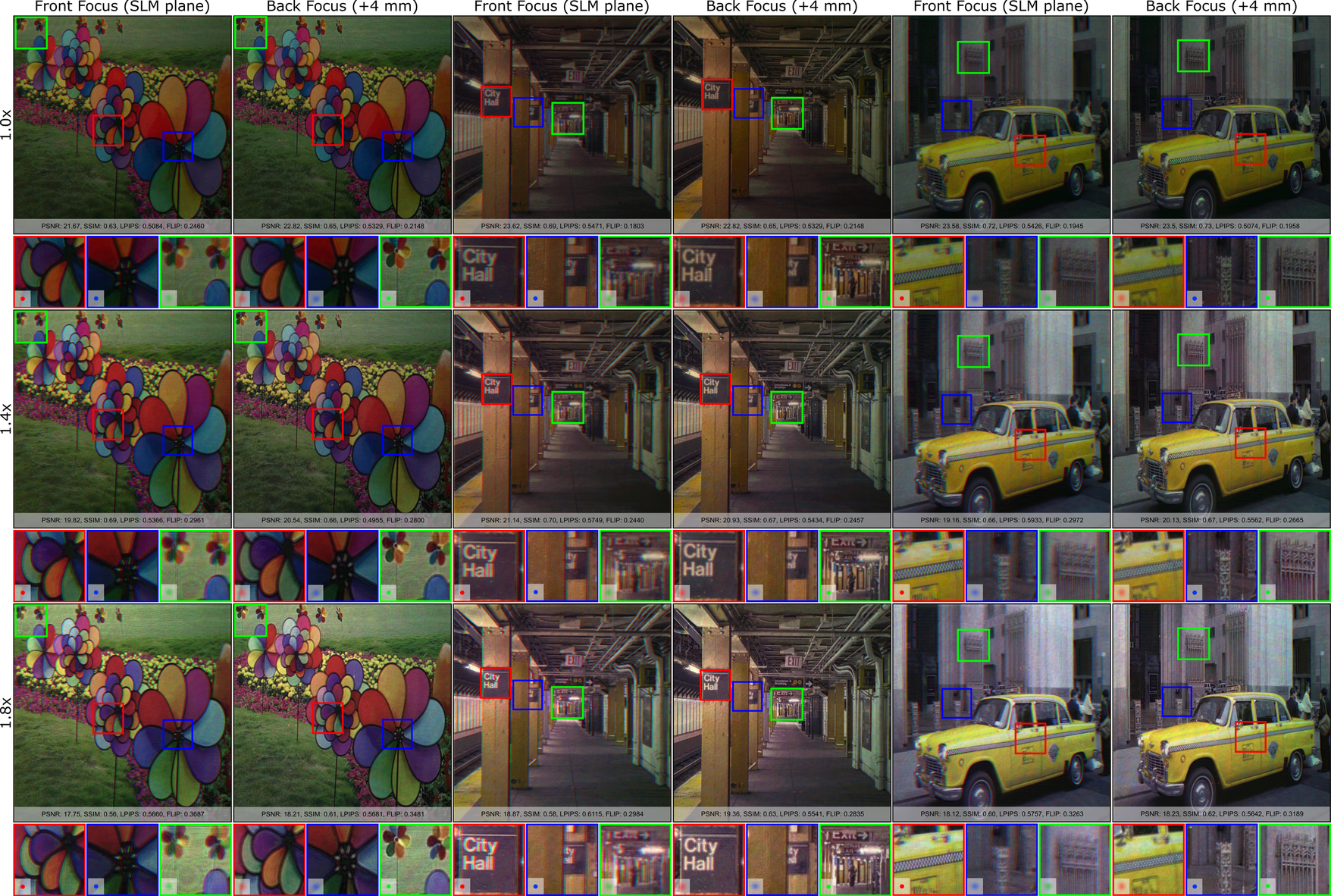}
  \caption{The hardware-captured short propagation result of student model when peak brightness set to 1.0, 1.4 and 1.8.
  From left to right: (Image Source of the first example: [Bernard Spragg 2009]) (Image Source of the second example: [LogicalRailfan 2023]) (Image Source of the third example: [Pilettes 2011])}
  \label{fig:different_laser_focus_short}
  \vspace{-5mm}
\end{figure*}

\begin{figure}[t!]
  \centering
  \includegraphics[width=0.99\columnwidth]{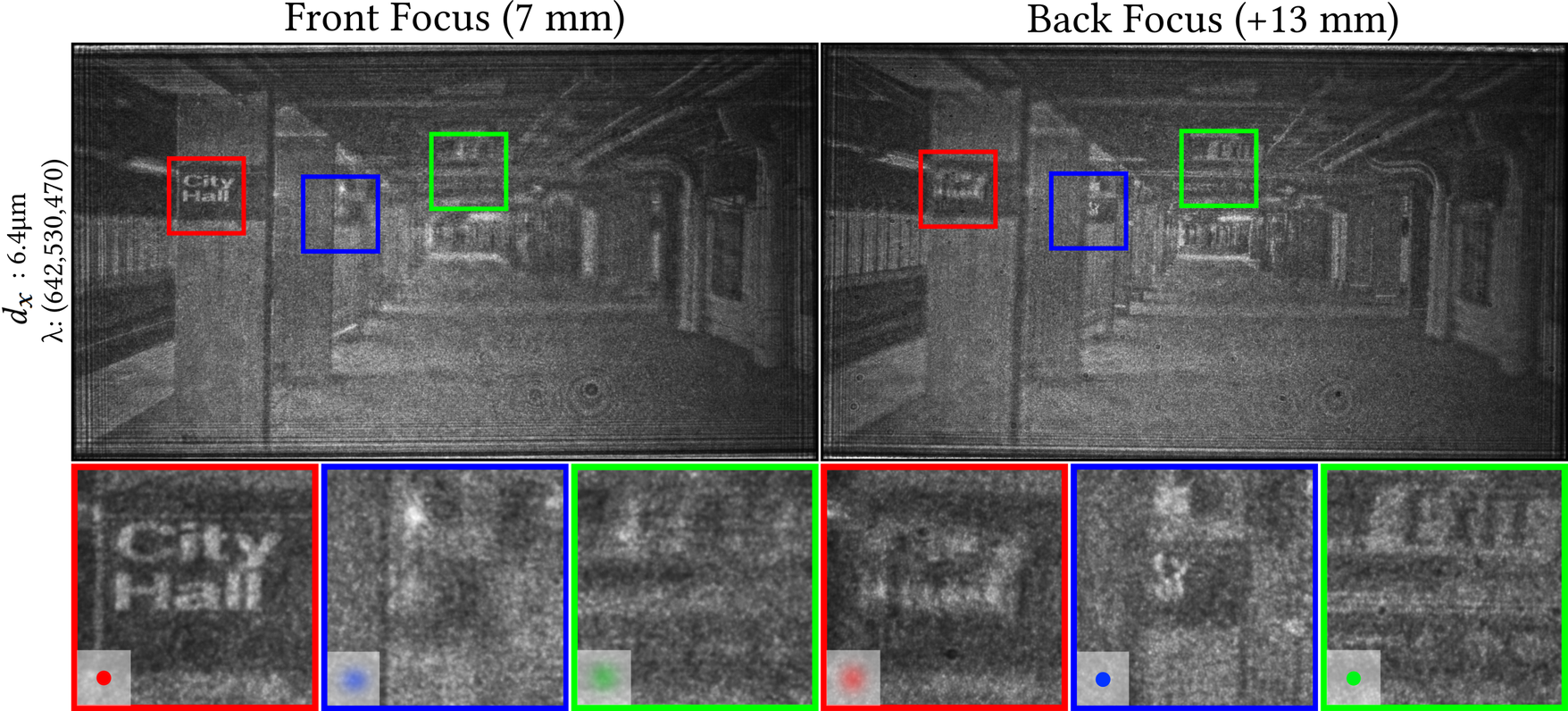}
  \caption{The hardware-captured result of RGB-D condition model on Holoeye LETO-3 (pixel pitch = $6.4 ~\mu m$) with volume depth 6 mm.
  Due to maintenance issue, the result was captured using a green laser only. (Source Image: \cite{subway2023}).}
  \label{fig:result_another_display_leto}
  \vspace{-5mm}
\end{figure}

\begin{figure*}[t!]
  \centering
  \includegraphics[width=1.0\textwidth]{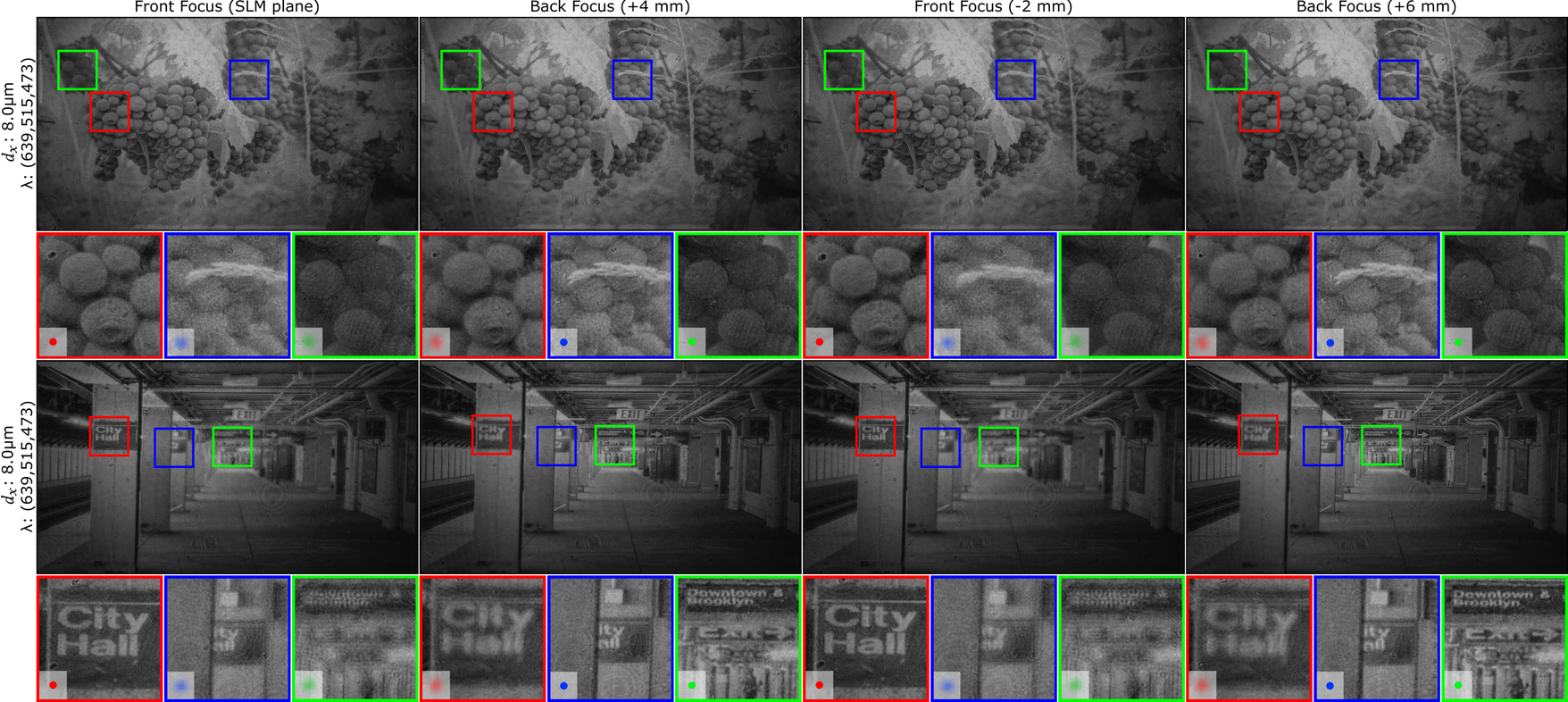}
  \caption{The hardware-captured result of RGB-D condition model on Holoeye Pluto-VIS (pixel pitch = $8.0 ~\mu m$) when volume depth set to 4.0 and 8.0 mm.
  Due to maintenance issue, the result was captured using a red laser only.
  From top to bottom: (Image Source of the top example: \cite{grapes2012}) (Image Source of the bottom example: \cite{city2023})}
  \label{fig:result_another_display}
\end{figure*}

\begin{figure*}[t!]
  \centering
  \includegraphics[width=1.0\textwidth]{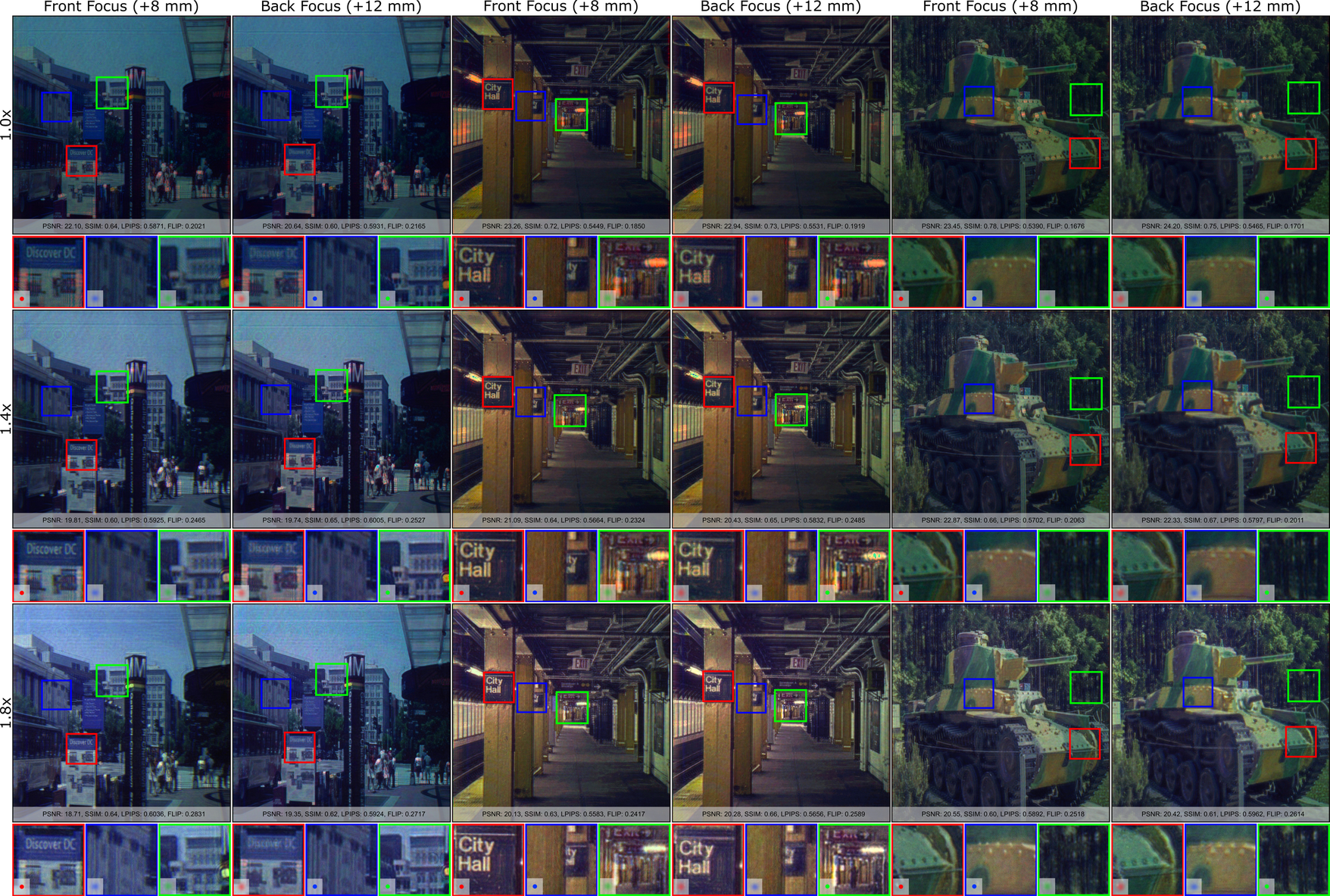}
  \caption{The hardware-captured long propagation result of student model when peak brightness set to 1.0, 1.4 and 1.8. From left to right: (Image Source of the first example: [Steve Tatum 2010]) (Image Source of the second example: [LogicalRailfan 2023]) (Image Source of the third example: [Mike1979 Russia 2014])}
  \label{fig:different_laser_focus_long}
\end{figure*}

\begin{figure*}[t!]
  \centering
  \includegraphics[width=0.86\textwidth]{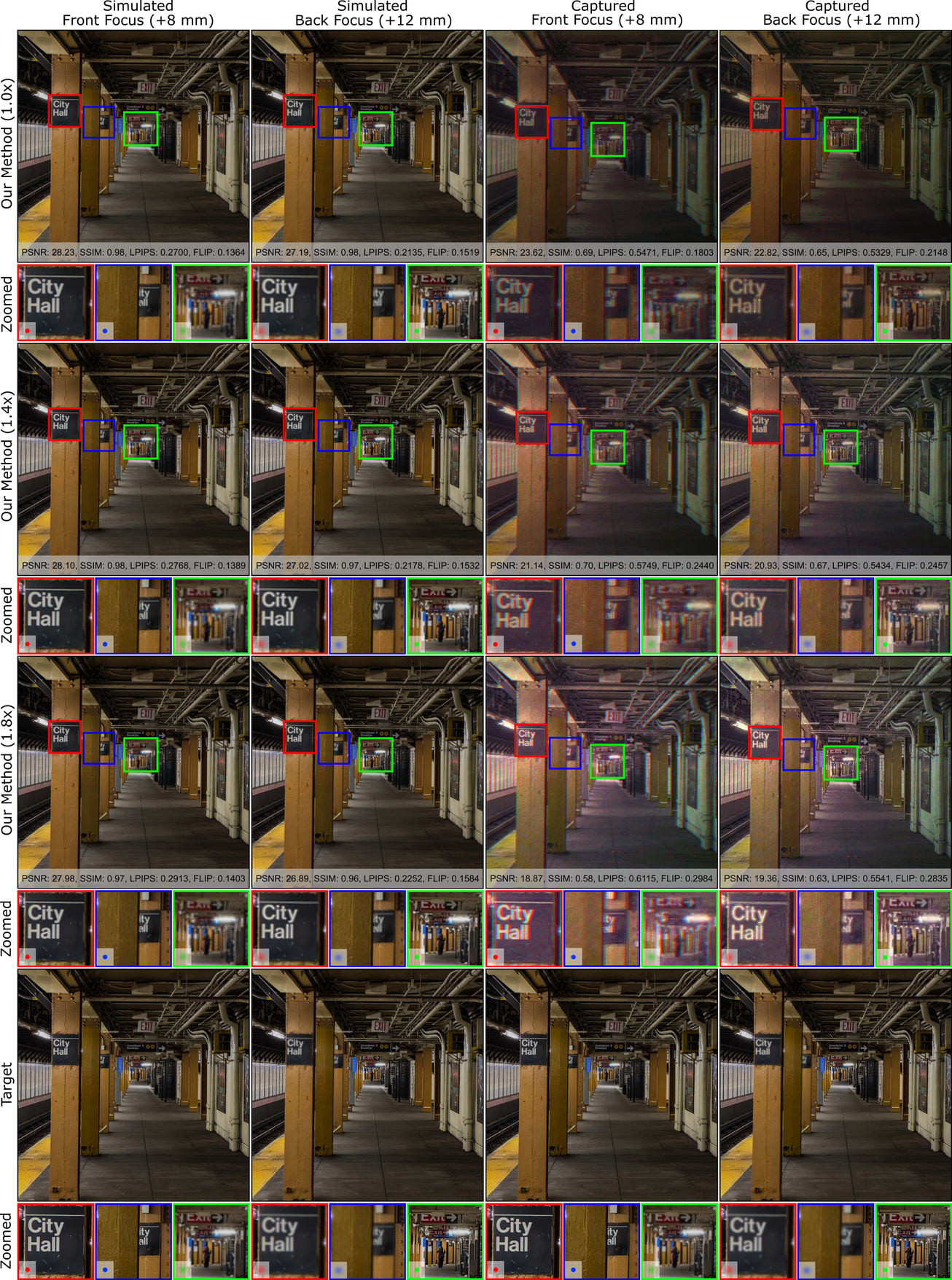}
  \caption{The simulated and captured short $Z$ distance reconstructions comparison of student model (RGB-only) when peak brightnesses are 1.0, 1.4, and 1.8. The volume depth of the results is 4~$mm$ and the propagation distance is 2~$mm$. The resolution of the tested hologram is 2816 $\times$ 2816. All the photographs are captured at 16.6 $ms$ exposure time (Source Image: \cite{subway2023}).}
  \label{fig:eval_short_prop_1_4}
\end{figure*}

\begin{figure*}[t!]
  \centering
  \includegraphics[width=0.86\textwidth]{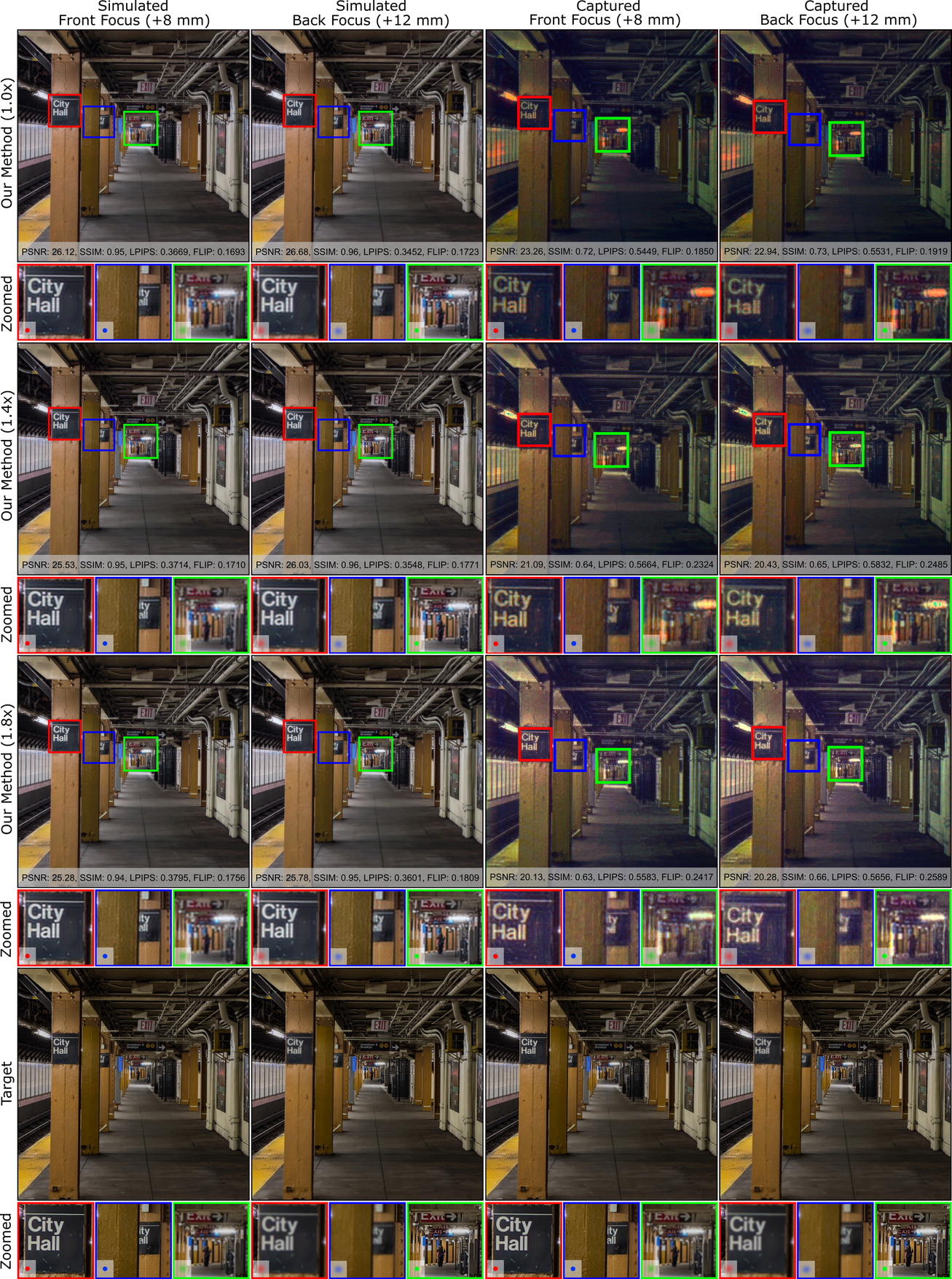}
  \caption{The simulated and captured long $Z$ distance reconstructions comparison of student model (RGB-only) when peak brightnesses are 1.0, 1.4, and 1.8. The volume depth of the results is 4~$mm$ and the propagation distance is 10~$mm$ The resolution of the tested hologram is 2816 $\times$ 2816. All the photographs are captured at 16.6 $ms$ exposure time (Source Image: \cite{subway2023}).}
  \label{fig:eval_long_prop_1_4}
\end{figure*}



\end{document}